%% file: arxiv.tex
\pdfoutput=1
\documentclass[leqno,onefignum,onetabnum]{siamltex}
\input macros.tex
\usepackage[left=2.5cm,right=2.5cm, top=2cm, bottom=2cm]{geometry}

\newcommand\atsign{@}

\makeatletter
\newcommand\footnoteref[1]{\protected@xdef\@thefnmark{\ref{#1}}\@footnotemark}
\makeatother

\title{Inefficiency of K-FAC for Large Batch Size Training}

\author{
Linjian Ma$^{1}$\footnote{\label{ecc}Equal contribution. Authors ordered alphabetically.}
, 
Gabe Montague$^{1}$\footnoteref{ecc}
, 
Jiayu Ye$^{1}$\footnoteref{ecc}
,\\ 
Zhewei Yao$^1$, 
Amir Gholami$^1$, 
Kurt Keutzer$^1$, AND 
Michael W. Mahoney$^1$
\\
$^1$University of California at Berkeley, 
\\
\{linjian, gabe\_montague, yejiayu, zheweiy, amirgh, keutzer, mahoneymw\}\atsign berkeley.edu
}

\usepackage{fancyhdr}
\begin{document}
\maketitle
\thispagestyle{empty}

\input abstract.tex
\input s1_intro.tex

\input s2_related_work.tex

\input s3_experimental_setup.tex

\input s4_results.tex

\input s5_conclusion.tex

\section*{Acknowledgments}
This work was supported by a gracious fund from Intel corporation. We would like
to thank the Intel VLAB team for providing us with access to their computing cluster.
We also gratefully acknowledge the support of NVIDIA Corporation for their donation of the Titan Xp GPU used for this research.
MWM would also like to acknowledge ARO, DARPA, NSF, ONR, and Intel for providing partial support of this work.

\clearpage
{\small
\bibliography{ref.bib}
 
\bibliographystyle{ieeetr}
}

\clearpage
\input figure_dump.tex

\end{document}

%% file: macros.tex
\usepackage{subfig}

\usepackage[utf8]{inputenc} %
\usepackage[T1]{fontenc}    %
\usepackage{hyperref}       %
\usepackage{url}            %
\usepackage{booktabs}       %
\usepackage{amsfonts}       %
\usepackage{nicefrac}       %
\usepackage{microtype}      %
\usepackage{color}
\usepackage{amsmath}
\usepackage{amsthm}
\usepackage{graphicx}
\usepackage{tikz}
\usepackage{caption}
\usepackage{multirow}
\usepackage{wrapfig}
\usepackage{algorithm, algpseudocode}%
\usepackage{amsfonts, amsmath}
\usepackage{enumitem}
\usepackage{soul}
\usepackage{bm}
\usepackage{wrapfig}
\usepackage{listings}
\usepackage[title]{appendix}
\usepackage[bottom]{footmisc}
\usepackage[makeroom]{cancel}

\usepackage{siunitx}

\definecolor{light-gray}{gray}{0.80}

\renewcommand\paragraph{\subsubsection*}

\newcommand\fref{Fig.~\ref}

\def\0{{\bf 0}}

\def\R{{\mathbb R}}

\usepackage{amsthm}

\newtheorem{mytheorem}{Theorem}
\newtheorem{assumption}[mytheorem]{Assumption}
\newtheorem{lemma}[mytheorem]{Lemma}
\newtheorem{remk}[mytheorem]{Remark}

%% file: abstract.tex
\begin{abstract}
In stochastic optimization, using large batch sizes during training can leverage parallel resources to produce faster wall-clock training times per training epoch.
However, for both training loss and testing error, recent results analyzing large batch Stochastic Gradient Descent (SGD) have found sharp diminishing returns, beyond a certain critical batch size.
In the hopes of addressing this, it has been suggested that the Kronecker-Factored Approximate Curvature (\mbox{K-FAC}) method allows for greater scalability to large batch sizes, for non-convex machine learning problems such as neural network optimization, as well as greater robustness to variation in model hyperparameters. 
Here, we perform a detailed empirical analysis 
of large batch size training 
for both \mbox{K-FAC} and SGD,
evaluating performance in terms of both wall-clock time and aggregate computational cost.
Our main results are twofold: 
first, we find that 
both \mbox{K-FAC} and SGD doesn't have ideal scalability behavior beyond a certain batch size, 
and that
\mbox{K-FAC} does not exhibit improved large-batch scalability behavior, as compared to SGD; and
second, we find that \mbox{K-FAC}, in addition to requiring more hyperparameters to tune, suffers from similar hyperparameter sensitivity behavior as does SGD. 
We discuss extensive results using 
ResNet and AlexNet on \mbox{CIFAR-10} and SVHN, respectively, as well as more general implications of our findings.
\end{abstract}

%% file: s1_intro.tex
\section{Introduction}
\label{sec:intro}
As the boundaries of parallelism are pushed by modern hardware and distributed systems, researchers are increasingly turning their attention toward leveraging these advances for faster training of deep neural networks (DNNs).
When using the prevailing Stochastic Gradient Descent (SGD) method, a batch of training data is split across computational processing units, which together compute a stochastic gradient used to update the parameters of the DNN.

To allow for efficient parallel scalability to a large number of processors, one would like to use a large batch of training data to compute the stochastic gradient estimate~\cite{gholami2017integrated}.
However, using a large batch size changes the dynamics of the training. 
It has been demonstrated both theoretically~\cite{ma2017power,MM18_TR,MM19_HTSR_ICML} and empirically~\cite{yao2018hessian,Noah-EMP-CRIT-BS,OpenAI-EMP-LBS,GOOG-LB-KFAC-HYPO,keskar2016large} that, in many cases, training with large-batch SGD comes with significant drawbacks.
This includes degraded testing performance, worse implicit regularization, and diminishing returns in terms of training loss reduction. 
Among other things,
there exists a critical batch size beyond which these effects are most acute.
For practitioners operating on a data-driven computational budget, large batch size comes with the additional inconvenience of increased sensitivity to hyperparameters and thus increased tuning time and cost~\cite{GOOG-LB-KFAC-HYPO}. 
In attempts to mitigate these shortcomings, a number of solutions have been proposed and demonstrated to exhibit varying degrees of effectiveness~\cite{FACEBOOK-IMAGENET-1H,smith2017don,yao2018large,mu2018parameter,osawa2018second,ginsburg2018large,GOOG-2M-IMAGENET,jia2018highly}.

An important practical consideration in methods such as~\cite{FACEBOOK-IMAGENET-1H,smith2017don,mu2018parameter,osawa2018second,ginsburg2018large,jia2018highly} is the sensitivity to hyperparameters. 
Generally, this sensitivity is quite 
strong, and 
the required tuning process is expensive in terms of both analyst time and in-search training time.
If performing large batch training requires significant hyperparameter tuning for each batch-size, then one would not achieve any effective speed up in total training time (i.e., hyperparameter tuning time plus final training time).
With the exception of the recent work of~\cite{yao2018large}, these discussions are largely ignored in the proposed solutions.

Recently, it has been suggested~\cite{GOOG-LB-KFAC-HYPO,OpenAI-EMP-LBS,ba2016distributed} that the (very) approximate second-order optimization method known as Kronecker-Factored Approximate Curvature (K-FAC)~\cite{KFAC-G15} may help to alleviate the issues of large-batch data inefficiency and generalization degradation exhibited by the first-order SGD method. 
K-FAC views the parameter space as a manifold of distribution space, in which distance between parameter vectors is measured by a variant of the Kullback–Leibler divergence between their corresponding distributions. 
In certain circumstances, K-FAC has been demonstrated to attain comparable effectiveness with large batch size as SGD~\cite{ba2016distributed,osawa2018second}, but the effects of batch size on training under K-FAC remain largely unstudied. 

In this work, we investigate these issues, evaluating the hypothesis that K-FAC is capable of alleviating the difficulties of large-batch SGD. 
In particular, we focus on the following two questions regarding K-FAC and large batch training:

\begin{itemize}
[noitemsep,topsep=0pt,parsep=0pt,partopsep=0pt,leftmargin=*]
    \item What is the scalability behavior of K-FAC, 
    and how does it 
    compare with that of SGD? 
    \item How does increasing batch size affect the hyperparameter sensitivity of K-FAC?
\end{itemize}
To answer these questions, we conduct a comprehensive investigation in the context of image classification on \mbox{CIFAR-10}~\cite{krizhevsky2009learning} and SVHN~\cite{netzer2011reading}. We investigate the performance of CIFAR-10 with a Residual Network (ResNet20 and ResNet32) classifier~\cite{he2016deep}, and we investigate SVHN with a AlexNet classifier~\cite{krizhevsky2012imagenet}. 
We investigate the problem of large-batch diminishing returns by measuring iteration speedup 
and comparing it to an ideal scaling scenario. 
Our key observations are as follows:
\begin{itemize}
[noitemsep,topsep=0.5pt,parsep=0.5pt,partopsep=0.5pt,leftmargin=*]
    \item 
    \textbf{Performance.} 
    Even with extensive hyperparameter tuning, K-FAC has comparable, but not superior, train/test performance to SGD
    (\fref{fig:combined_batch_v_acc}). 
    \item 
    \textbf{Speedup.}
    Both \mbox{K-FAC} and SGD have diminishing returns with the increase of batch size. Increasing batch size for K-FAC yields lower, i.e., less prominent, speedup, as compared with SGD, when measured in terms of iterations (\fref{fig:combined_batch_loss_reduce}). 
    \item 
    \textbf{Hyperparameter Sensitivity.}
    \mbox{K-FAC} hyperparameter sensitivity depends on both batch size and epochs/iterations. For fixed epochs, i.e., running the same number of epochs, larger batch sizes result in greater hyperparameter sensitivity and smaller regions of hyperparameter space which result in ``good convergence.'' For fixed iterations, i.e., running the same number of iterations, larger batch sizes result in less sensitivity and larger regions of hyperparameter space which result in ``good convergence'' (\fref{fig:heatmap}, \fref{fig:batch_v_acc_bp}). 
     
\end{itemize}

We start with mathematical background and related work in Section~\ref{sec:related_work}, followed by a description of our experimental setup in Section~\ref{sec:setup}. 
Our empirical results demonstrating the inefficiencies of both \mbox{K-FAC} and SGD with large batch sizes appear in Section~\ref{sec:result}. Our conclusions are in Section~\ref{sec:conclusion}.
Additional material is presented in the Appendix.

%% file: s2_related_work.tex
\section{Background and Related Work}\label{sec:related_work}

For a supervised learning framework, the goal is to minimize a loss function expressed as 
\begin{equation}
\label{eqn:basic_problem}
L(\theta) = \frac{1}{N} \sum_{i=1}^{N} l(x_i, y_i, \theta),
\end{equation}
where $\theta\in\R^d$ is the vector of model parameters,
and $l(x, y, \theta)$ is the loss for a datum $(x, y) \in (X, Y)$. 
Here, $X$ is the input, $Y$ is the corresponding label, and $N=|X|$ is the cardinality of the training set.
SGD is typically used to optimize the loss 
by taking steps of the form:
\small
\begin{equation}
\label{eqn:sgd}
\theta_{t+1} = \theta_t - \eta_t \frac{1}{|B|} \sum_{(x,y) \in B} \nabla_\theta l(x, y, \theta_t),
\end{equation}
\normalsize
where $B$ is a mini-batch of examples drawn randomly from $X\times Y$, and $\eta_t$ is the learning rate at iteration $t$. 

\subsection{Kronecker-Factored Approximate Curvature}
As opposed to SGD, which treats model parameter space as Euclidean, natural gradient descent methods~\cite{amari1998natural} for DNN optimization operate in the space of distributions defined by the model, in which the parameter distance between two vectors is defined using the KL-divergence between the two corresponding distributions. Denoting $D_{KL}$ as our vector norm in this space, it can be shown that $D_{KL} (\Delta \theta) \approx \frac{1}{2} \Delta\theta^\top F \Delta \theta$, where $F$ is the Fisher Information Matrix (FIM) defined~as:
\small
\begin{equation}
    F = \mathbb{E} [\nabla_\theta \log p(y |x, \theta ) \nabla_\theta \log p(y |x, \theta )^\top ],
\end{equation}
\normalsize
where the expectation is taken over both the model's training data space and target variable space~\cite{KFAC-G15}. The update rule for natural gradient descent then becomes 
$\theta_{t+1} = \theta_t - \eta_t F^{-1} \nabla_\theta \ell(\theta_t).$
As noted by~\cite{osawa2018second} and others, the FIM is often poorly-conditioned for DNNs, leading to unstable training. To counter this effect, a damping term is often added (i.e., the FIM is \textit{preconditioned}). Using the preconditioned FIM, the update rule then~becomes:
\small
\begin{equation}
\label{equation:damping}
\theta_{t+1} = \theta_t - \eta_t (F + \lambda I )^{-1} \nabla_\theta \ell(\theta_t),
\end{equation}
\normalsize
with $\lambda$ denoting a positive damping parameter.
Due to the computational intractability of the true Fisher matrix, natural gradient methods typically rely on approximations to $F$. For example, \cite{KFAC-G15} proposes an approximation, the K-FAC methog, exploiting the assumption that (i) $F$ is largely block-diagonal\footnote{Although the K-FAC authors propose an alternative tridiagonal approximation that eases the strength of this assumption, we consider their block diagonal approximation of the Fisher, due to its demonstrated performance.} and (ii) across the training distribution, the products of unit activations and products of unit output derivatives are statistically independent. While these assumptions are inexact, their accuracy has been empirically verified by the authors in several cases. The approximation can be written as:
\small
\begin{equation} \label{kfac-approx}
    F_i = \mathbb{E}[ A_{i-1}A_{i-1}^\top \otimes G_iG_i^\top] \approx \mathbb{E}[A_{i-1}A_{i-1}^\top ]\otimes \mathbb{E}[G_iG_i^\top],
\end{equation}
\normalsize
where $F_i$ represents the Fisher matrix of $i$-th layer, $G_i$ is the gradient of the loss with respect to the $i$-th layer output before non-linear activation function, and $A_{i-1}$ is the activation output of the previous layer. Note that this is the approximation form we use in our implementation.

\subsection{Difficulties of Large Batch Training}

The problems of large batch training under SGD have been studied in detail through both analytical and empirical studies. \cite{ma2017power} proves that for convex cases, increasing batch size by a factor $f$ yields a no worse than a factor $f$ speedup in the number of SGD iterations, so long as batch size is below a critical point.
The batch sizes falling below this critical point are referred to collectively as the \textit{linear scaling regime}.
\cite{Noah-EMP-CRIT-BS} empirically investigates this in the context of non-convex training of DNNs for a variety of training workloads; and it finds evidence of a similar critical batch size for the non-convex case, before which $f$-fold increases of batch size yield $f$-fold reductions in total iterations needed to converge, and after which diminishing returns are observed, eventually leading to stagnation and no further benefit.

Subsequent to \cite{Noah-EMP-CRIT-BS}, \cite{GOOG-LB-KFAC-HYPO,OpenAI-EMP-LBS} obtain broadly similar conclusions with more detailed studies.
In particular,
\cite{OpenAI-EMP-LBS} goes further to predict the critical batch size to the nearest order of magnitude, demonstrating that critical batch size can be predicted from the \textit{gradient noise scale}, representing the noise-to-signal ratio of the stochastic estimation of the gradient. The authors further find that gradient noise scale increases during the course of training. This principle motivates the success of techniques as in~\cite{smith2017don,devarakonda2017adabatch,yao2018large}, in which batch size is adaptively increased during training.

Apart from increasing batch size during training, effort has been undertaken to increase critical batch size and linear scaling throughout the entire training process. \cite{FACEBOOK-IMAGENET-1H} attempts to improve SGD scalability by tuning hyperparameters more carefully using a linear batch-size to learning-rate relationship. While this proves effective for the authors' training setup, \cite{Noah-EMP-CRIT-BS} demonstrates that for a wide variety of other training workloads a linear scaling rule is ineffective to counter inefficiencies of large batch.

\cite{GOOG-LB-KFAC-HYPO} proposes that \mbox{K-FAC} may help to extend the linear scaling regime and increase the critical batch size further than exhibited with SGD, allowing for greater scalability with large batches. 
Recent work has applied K-FAC to large-batch training settings, as in training ResNet50 on ImageNet within 35 epochs~\cite{osawa2018second}, along with the development  of a large-batch parallelized \mbox{K-FAC} implementation~\cite{ba2016distributed}. 
Both provide discussion of this, and both demonstrate ``near-perfect'' scaling behavior for large-batch \mbox{K-FAC}.
\cite{ba2016distributed} goes further in depth to suggest that K-FAC scalability is superior to SGD for high training losses and low batch sizes, for a single training workload.
Our work more formally investigates the scalability of K-FAC versus SGD, and it finds evidence to the contrary; that is, for the workload and batch sizes we consider, K-FAC scalability is no better than that of SGD.

%% file: s3_experimental_setup.tex
\section{Experimental Setup}\label{sec:setup}

We investigate the performances of both K-FAC and SGD on \mbox{CIFAR-10} with ResNet20 and ResNet32, and on SVHN with AlexNet. To be comparable with state-of-art results for K-FAC~\cite{wd-kfac}, we apply batch normalization to the 
models along with standard data augmentation during the training process. We further regularize with a weight decay parameter of $5 \times 10^{-4}$. We perform extensive hyperparameter tuning individually for each batch size ranging from 128 to 16,384.

\subsection{Training Budget and Learning Rate Schedule}
In many scenarios, training within a hyperparameter search is stopped after some fixed amount of time. When this time is specified in terms of number of epochs, we call this a \textit{(normal) epoch budget}. For our training of K-FAC and SGD, we use a modified version of this stopping rule which we refer to as an \textit{adjusted epoch budget}, in which the epoch limit of training is extended proportionally to the log of the batch size. Specifically, we use the rule: for CIFAR-10, number of training epochs equals $(\log_2(\text{batch size} / 128) + 1) \times 100$; for SVHN, it equals $(\log_2(\text{batch size} / 128) + 1) \times 20$. This adjusted schedule allows larger-batch training runs more of a chance to converge by affording them a greater number of iterations than would normally be allowed under a traditional epoch budget.

For experiments on CIFAR-10, we decay the learning rate twice by a factor of ten for each run over the course of training. These two learning rate decays separate the training process into three \textit{stages}. Because training extends to a greater number of epochs for large batches under the adjusted epoch budget, for large batch runs we allow a proportionally greater number of epochs to pass before learning rate decay. For each run we therefore decay the learning rate at $40\%$ and $80\%$ of the total epochs\footnote{This schedule can loosely be regarded as a mixture of an epoch-driven schedule, as in~\cite{hoffer2017train}, and an iteration-based schedule, as in~\cite{he2016deep}.}. We refer to this decay scheme as a \textit{scaled learning rate schedule}. In Appendix~\ref{sec:comparison_fix_scale}, we empirically validate our reasoning that our scaled learning rate schedule
(as opposed to a \textit{fixed learning rate schedule}\footnote{For our investigated fixed learning rate schedule, the two learning rate decays happen at epochs 40 and 80, regardless of the total number of epochs. A similar schedule is used in~\cite{hoffer2017train}.}) helps large-batch performance. Similarly, for SVHN, we also choose the \textit{scaled learning rate schedule} and decay the learning rate by a factor of 5 at 50\% of the total epochs.

\subsection{Hyperparameter Tuning}

For both K-FAC and SGD, in order to perform training, we must deal with hyperparameters, and we describe this here.

\paragraph{\textbf{K-FAC}} For K-FAC, we use the various techniques discussed in~\cite{ba2016distributed,osawa2018second}. We precondition the Fisher matrix based on Eqn.~(\ref{equation:damping}) according to the methodology presented in~\cite[Appendix A.2]{grosse2016kronecker}.
Although~\cite{KFAC-G15} argues in favor of an alternative damping mechanism that approximates the damping of Eqn.~(\ref{equation:damping}), we observed comparable performance using the normal approach. The details of these two methods and our comparison between them are laid out in the Appendix. 

For hyperparameter tuning of our CIFAR-10 experiments, we conduct a log-space grid search over 64 configurations with learning rates ranging from $10^{-3}$ to $2.187$, and with damping ranging from $10^{-4}$ to $0.2187$. 
Similarly, for our SVHN experiments, we conduct a log-space grid search over 64 configurations with learning rates ranging from $10^{-5}$ to $0.02187$, and with damping ranging from $10^{-4}$ to $0.2187$. 
The decay rate for second-order statistics is held constant at 0.9 throughout training. We use update clipping as in~\cite{ba2016distributed}, with a constant parameter of $0.1$.

\paragraph{\textbf{SGD}}
To ensure a fair comparison between methods, we employ a similarly extensive hyperparameter tuning process for SGD. In particular, we conduct a similar log-space grid search over 64 hyperparameter configurations. For CIFAR-10 experiments, learning rates range from 0.05 to 9.62, and momentum range from 0.9 to 0.999. For SVHN experiments, learning rates range from 0.005 to 0.962, and momentum range from 0.9 to 0.999. 

\subsection{Speedup Ratio}

We use \textit{speedup ratio}~\cite{Noah-EMP-CRIT-BS} to measure the efficiency of large batch training based on iterations.
We define the convergence rate $k_c(m)$ as the fewest number of iterations to reach a certain criteria $c$ under the batch size $m$, where $c$ is defined as attaining a target accuracy or loss threshold. Here, $k_{c}(m)$ is a minimum, as it is picked across all configurations of hyperparameters. We then define the \textit{speedup ratio} $s_c(m;m_0)$ as $k_c(m_0) / k_c(m)$, in which we rely on some small batch size $m_0$ as our reference for convergence rate when comparing to larger batch sizes $m > m_0$. In an ideal scenario, the batch size has no effect on the performance increase per training observation, so in such cases $s_c(m;m_0) = \frac{m}{m_0}$.

It should be noted that for K-FAC speedup we solely measure
the number of iterations and ignore the cost of computing the inversion of
the Fisher matrix. The latter can become very expensive, and multiple
approaches such as stochastic low-rank approximation and/or inexact iterative solves
can be used. However, as we will show, K-FAC speedup is far from ideal, even when ignoring this cost of performing more exact computations.

%% file: s4_results.tex
\section{Experimental Results}\label{sec:result}

We perform extensive experiments on both \mbox{CIFAR-10} and SVHN datasets with both K-FAC and SGD.
Section~\ref{sec:comparison_best_performance} compares the training and test performances of K-FAC and SGD resulting from extensive hyperparameter tuning for each batch size.
Section~\ref{sec:comparison_scalability} then discusses the large-batch scaling behaviors of K-FAC and SGD and compares them to the ideal scaling scenario~\cite{Noah-EMP-CRIT-BS,GOOG-LB-KFAC-HYPO}.
Finally, Section~\ref{sec:robustness_kfac} investigates the hyperparameter 
sensitivity
of the K-FAC method.

\begin{figure}[]
\begin{center}
  \includegraphics[width=.35\textwidth]{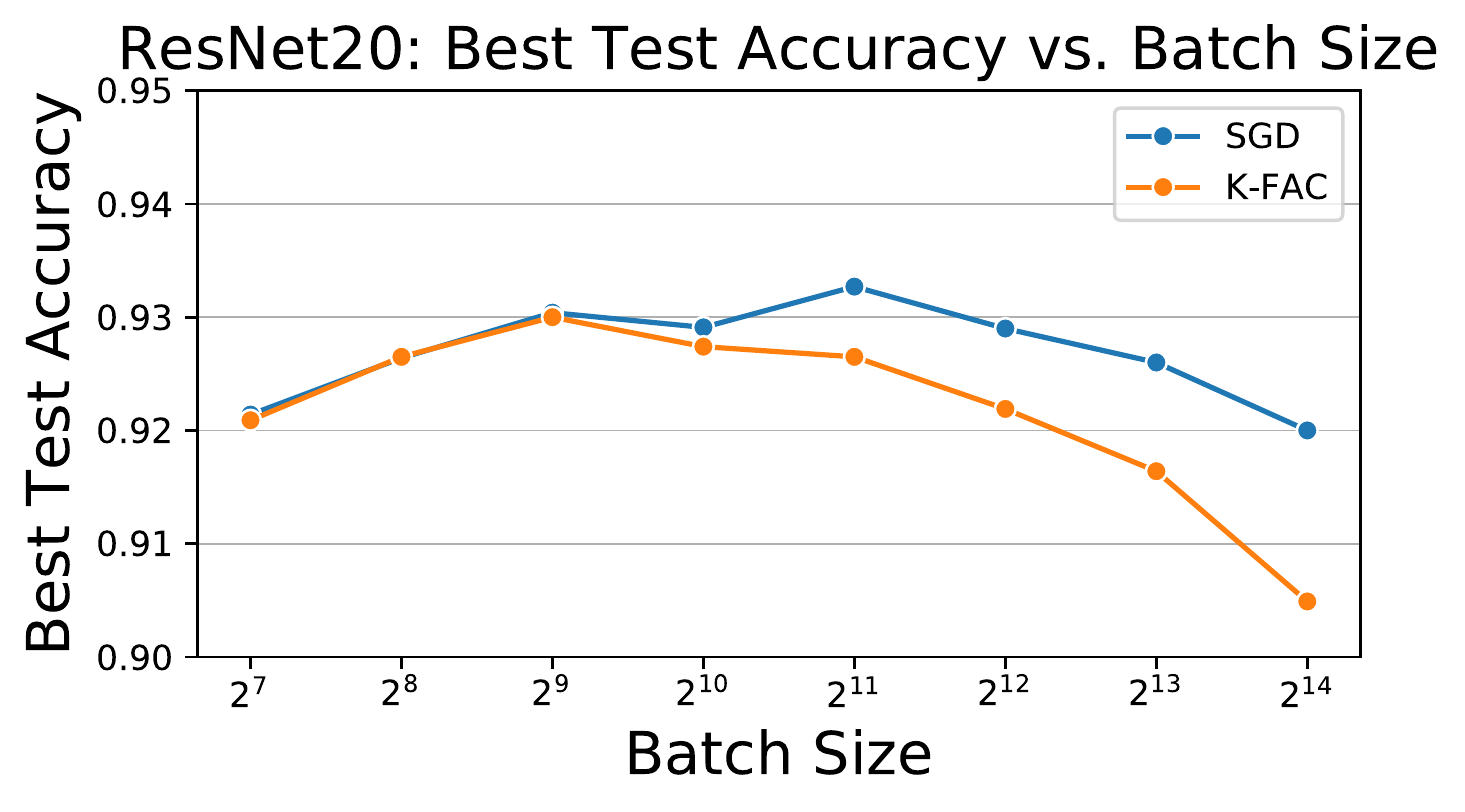}
  \includegraphics[width=.35\textwidth]{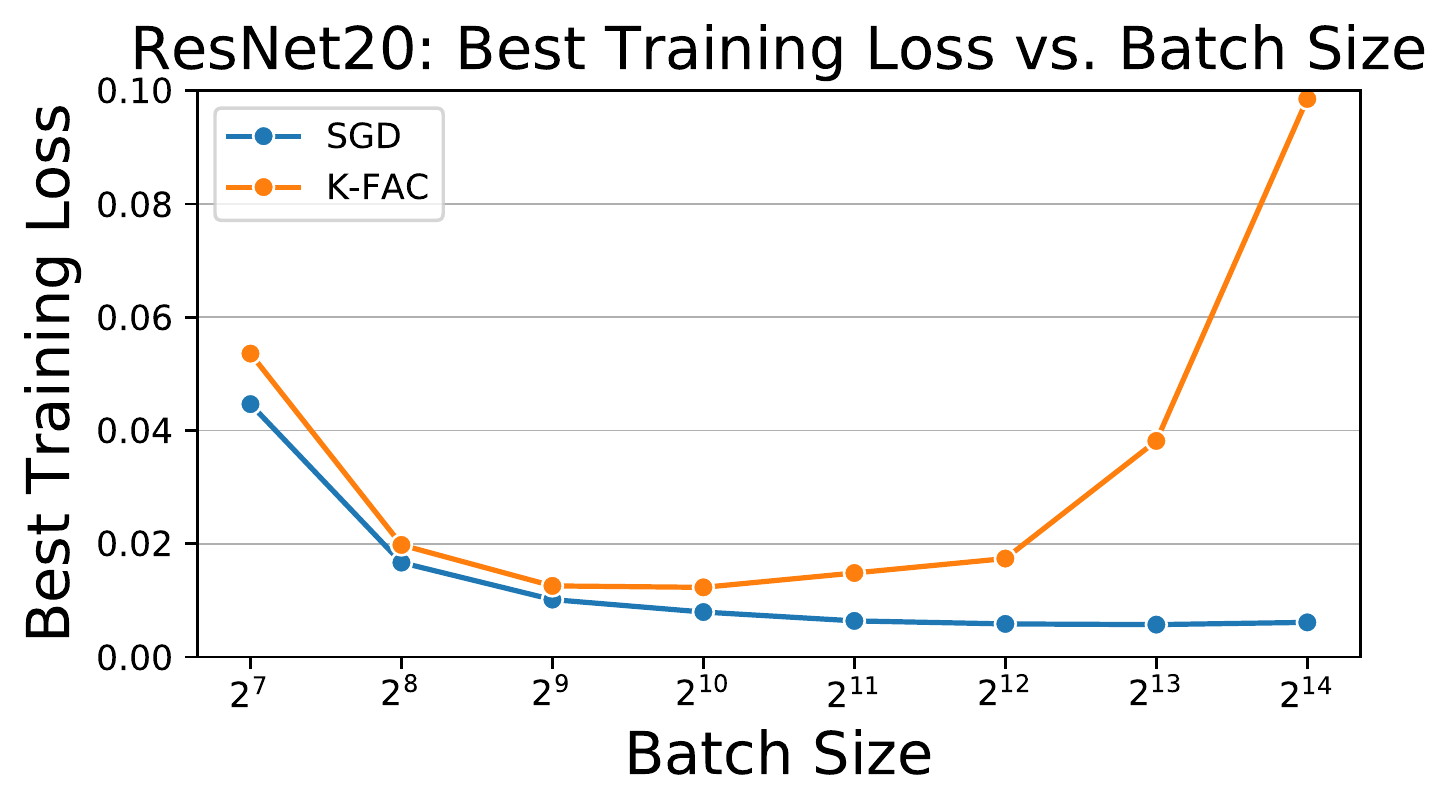}
  
  \includegraphics[width=.35\textwidth]{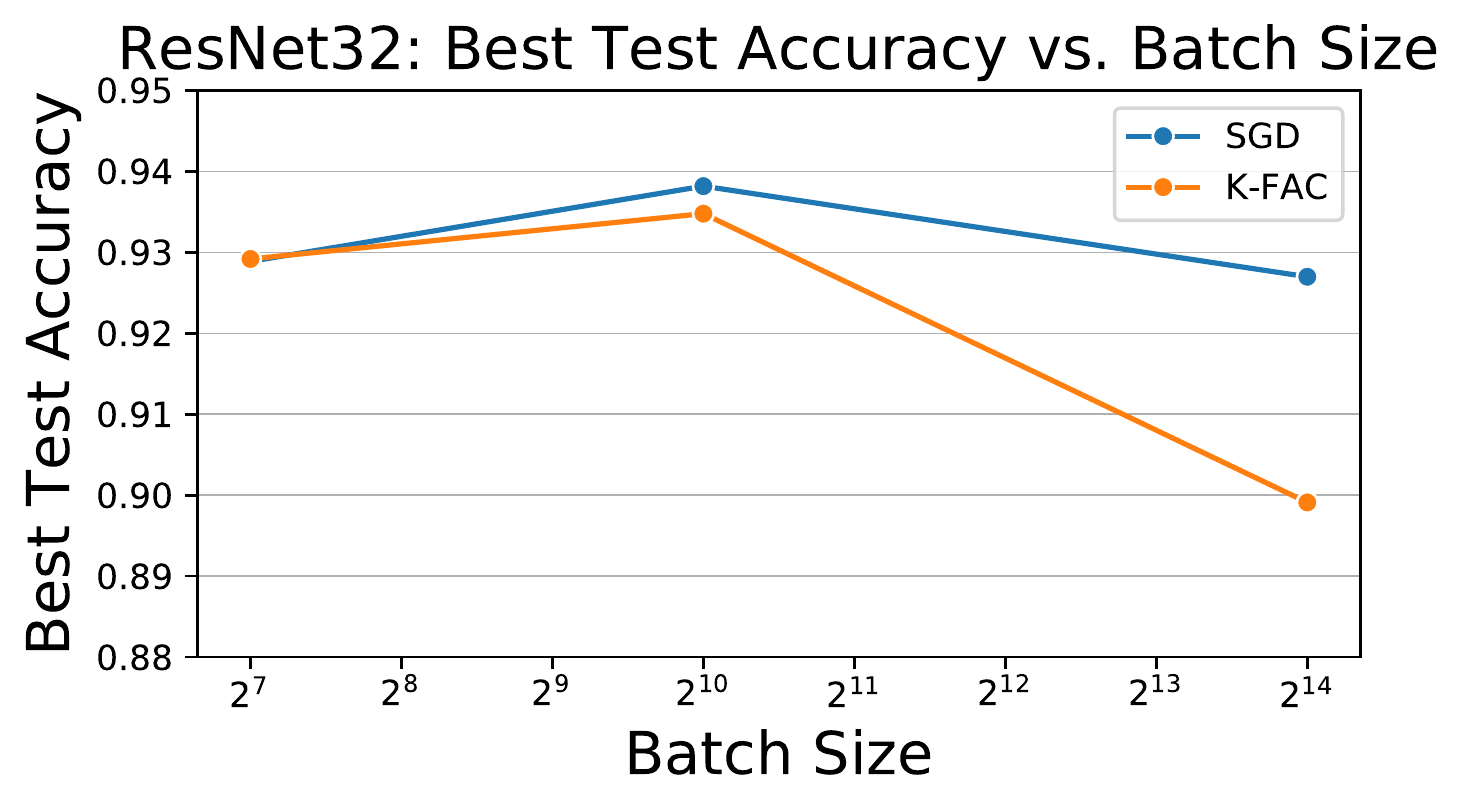}
  \includegraphics[width=.35\textwidth]{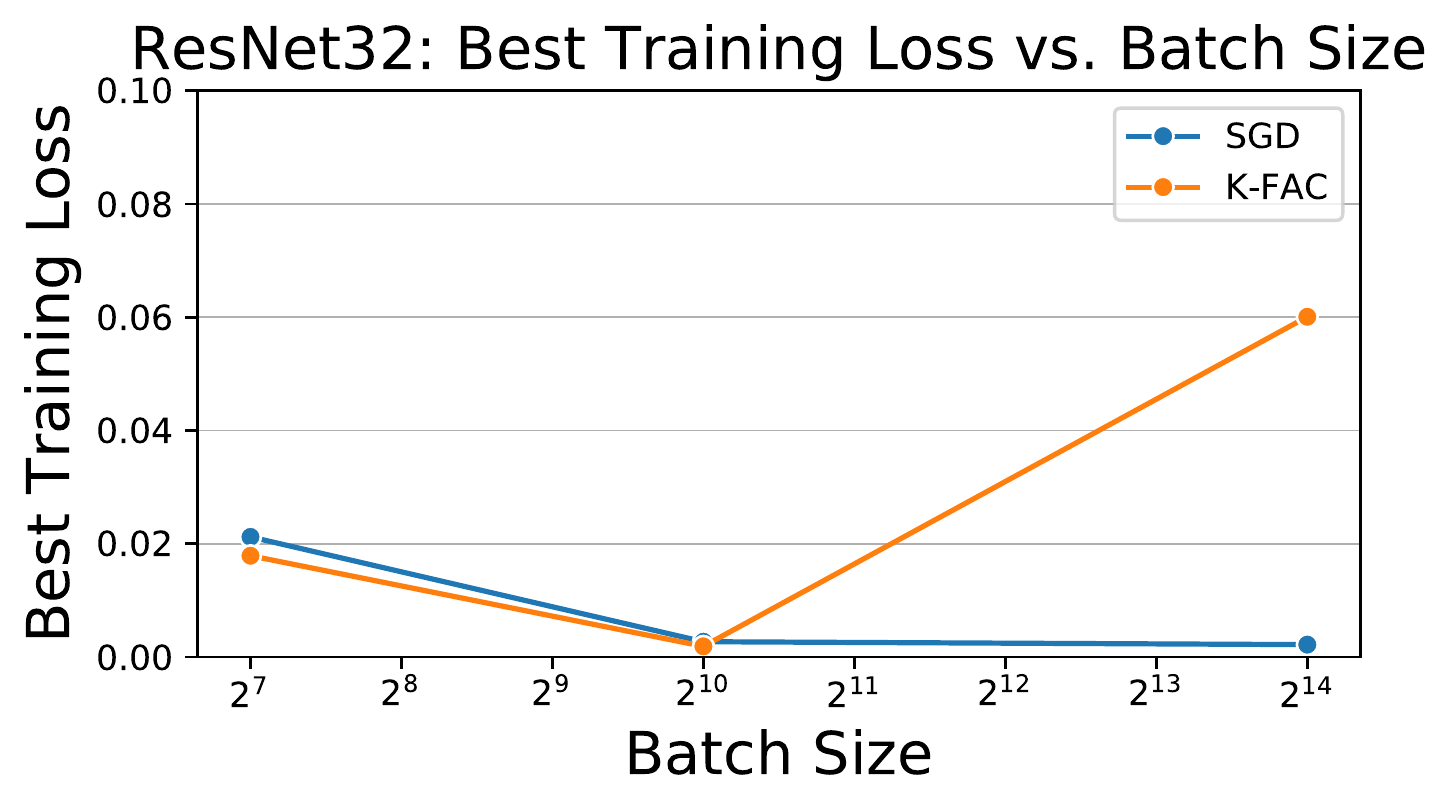}

  \includegraphics[width=.35\textwidth]{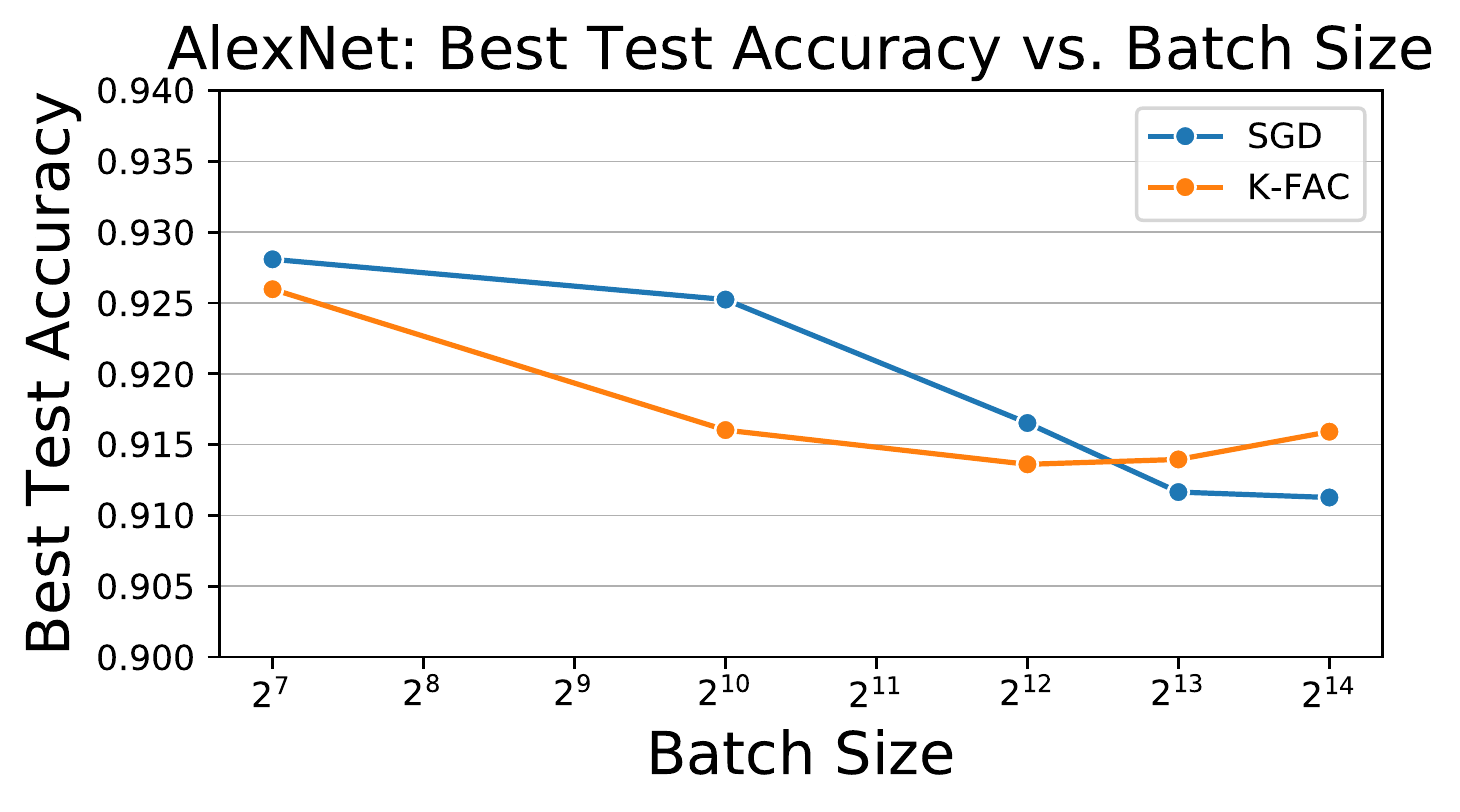}
  \includegraphics[width=.35\textwidth]{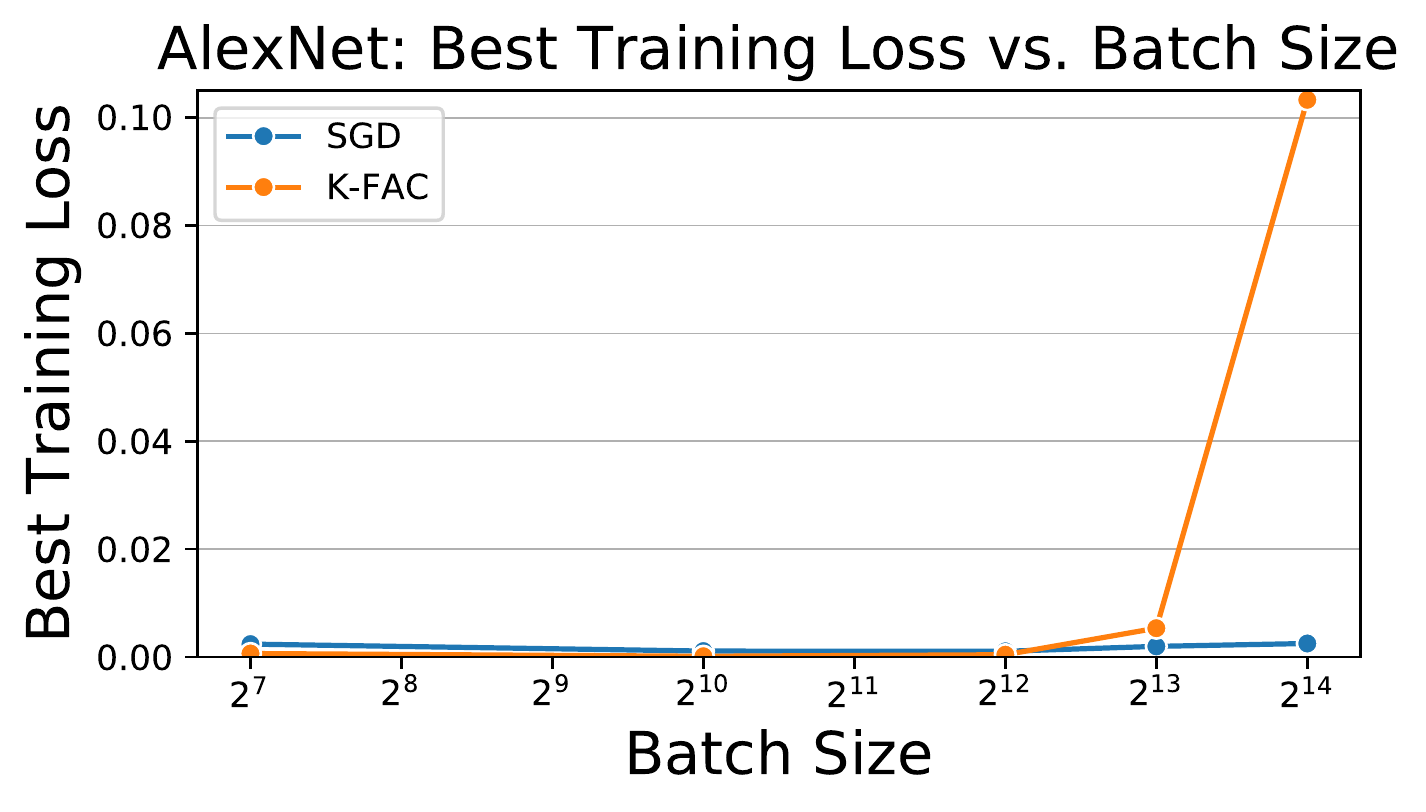}
\end{center}
\caption{From top to bottom: Best test accuracy / training loss versus batch size for SGD and K-FAC with ResNet20 on CIFAR-10, ResNet32 on CIFAR-10, and AlexNet on SVHN, respectively.
Large-batch K-FAC does not achieve higher accuracy or lower losses than large-batch SGD, given the same number of training~epochs.
}
\label{fig:combined_batch_v_acc}
\end{figure}



\subsection{Comparing Best Performance of K-FAC and SGD}\label{sec:comparison_best_performance}

We run K-FAC and SGD for multiple batch sizes, with stopping conditions determined according to the 
adjusted epoch budget and the scaled learning rate schedule. 
The highest test accuracy and the lowest training loss achieved for each batch size are plotted in~\fref{fig:combined_batch_v_acc}.
We include the detailed training trajectories over time for both SGD and K-FAC in Appendix~\ref{subsec:appendix-train-test-curve}. 

In this training context, K-FAC minimizes training loss most effectively for medium-sized batches 
(around $2^{10}$).
Inspecting the training trajectories, we found that both the smallest ($2^{7}$) and largest batch sizes ($2^{14}$) needed more epochs/iterations to converge. 
For SGD however, training loss is minimized prominently at larger batch sizes. When comparing training trajectories with K-FAC, we found that SGD made much more progress per-iteration in reducing loss, allowing it to minimize the objective with a smaller number of updates, as shown in~\fref{fig:combined_batch_v_acc}.  A more detailed comparison of the per-iteration progress of SGD versus K-FAC can be found in the following section.
For CIFAR-10 experiments, the gap between SGD and K-FAC in large-batch training loss is also present in their generalization performance. 
SGD's greater efficiency in maximizing per-iteration accuracy allows it to attain a higher level of test performance with the same number of training epochs.

\subsection{Large-Batch Scalability of K-FAC and SGD}\label{sec:comparison_scalability}

\begin{figure}[]
\begin{center}
  \includegraphics[width=.35\textwidth]{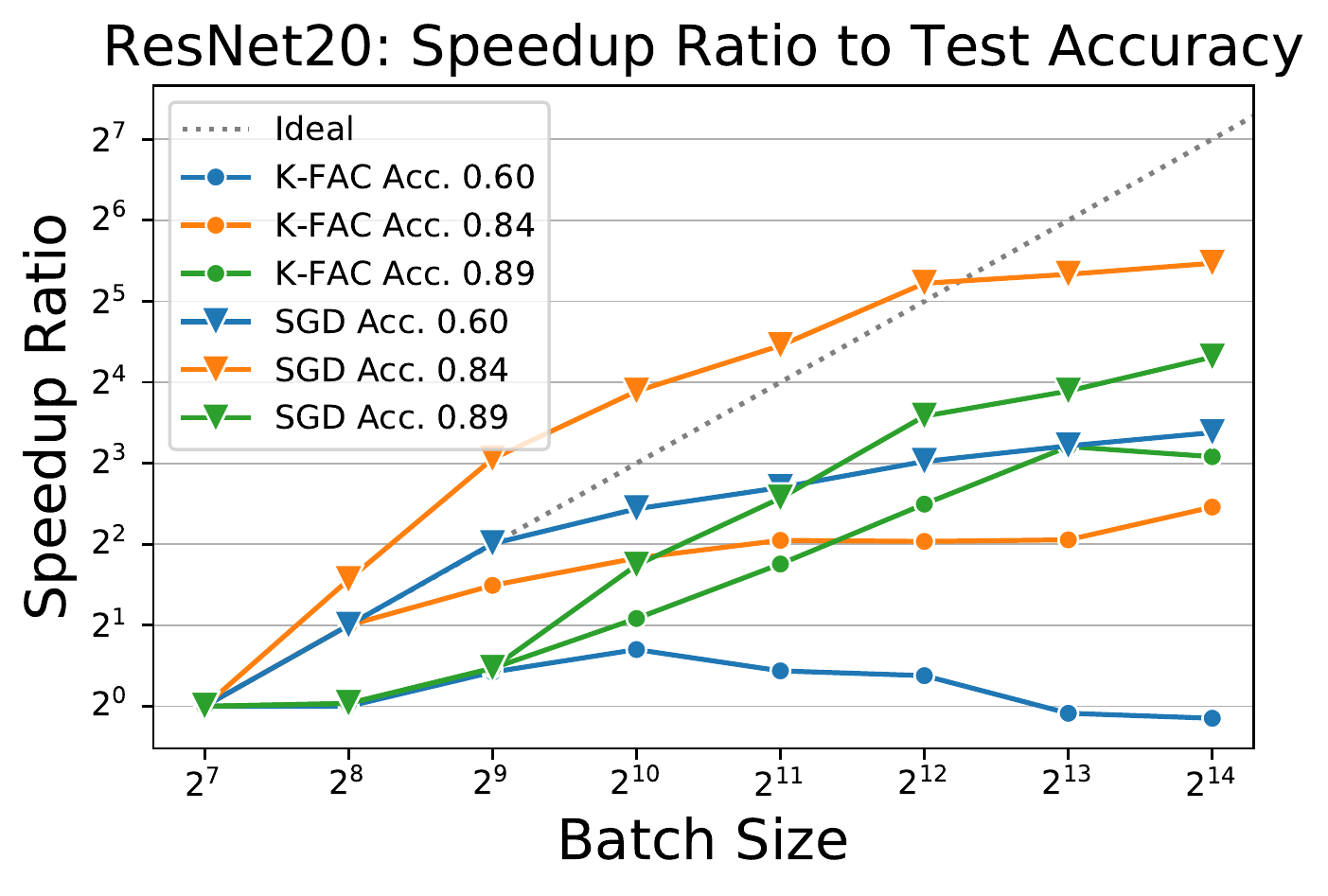}
    \includegraphics[width=.35\textwidth]{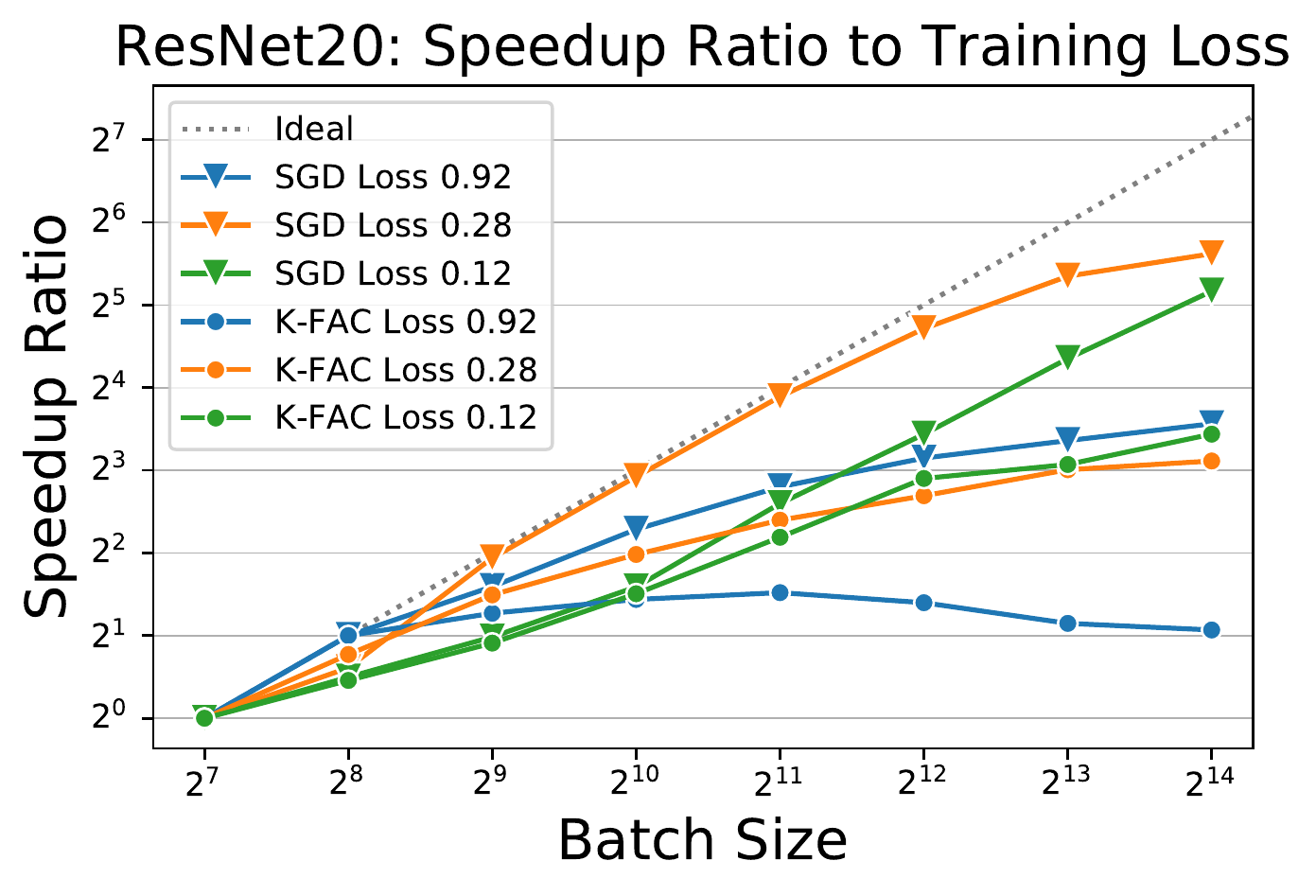}

  \includegraphics[width=.35\textwidth]{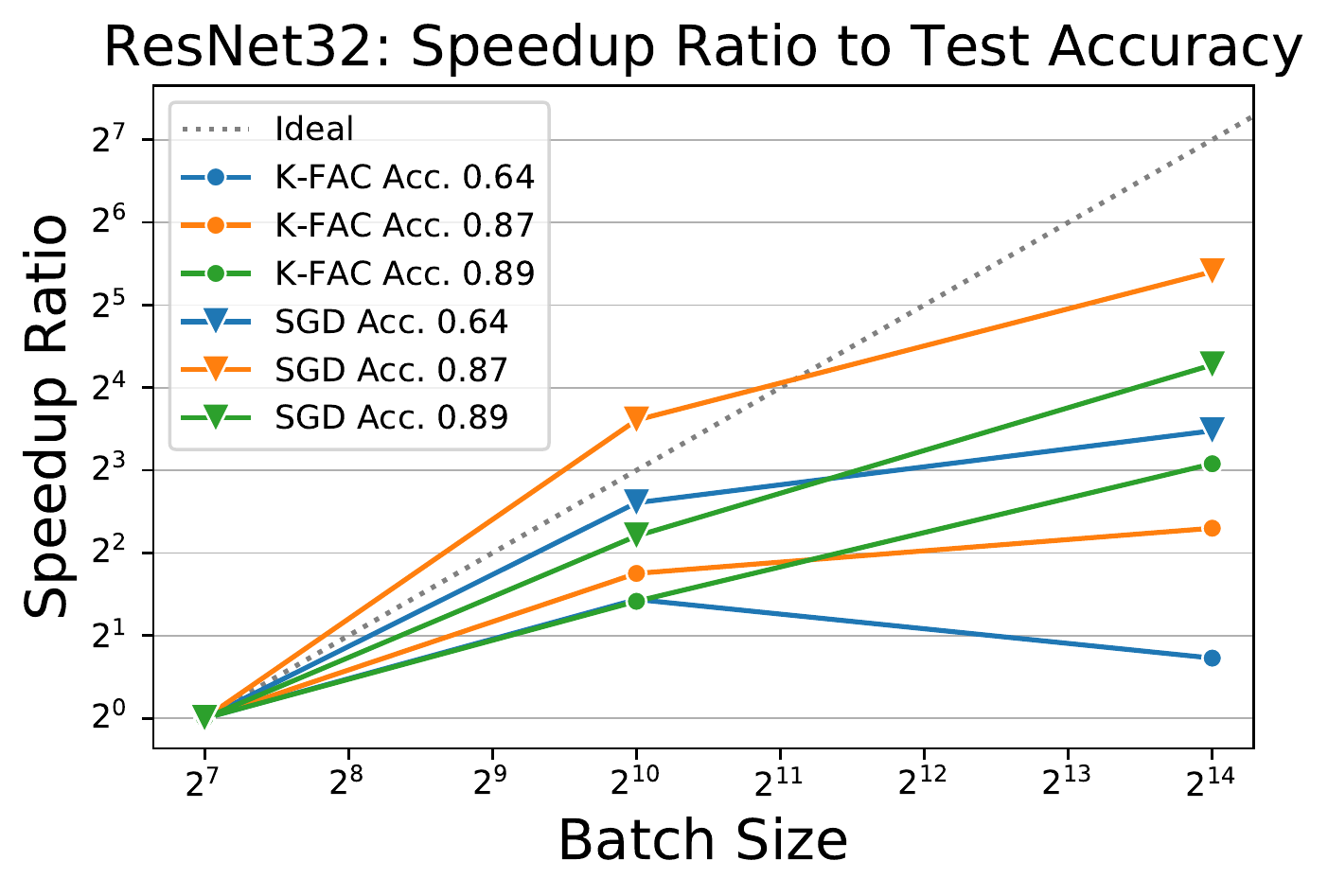}
  \includegraphics[width=.35\textwidth]{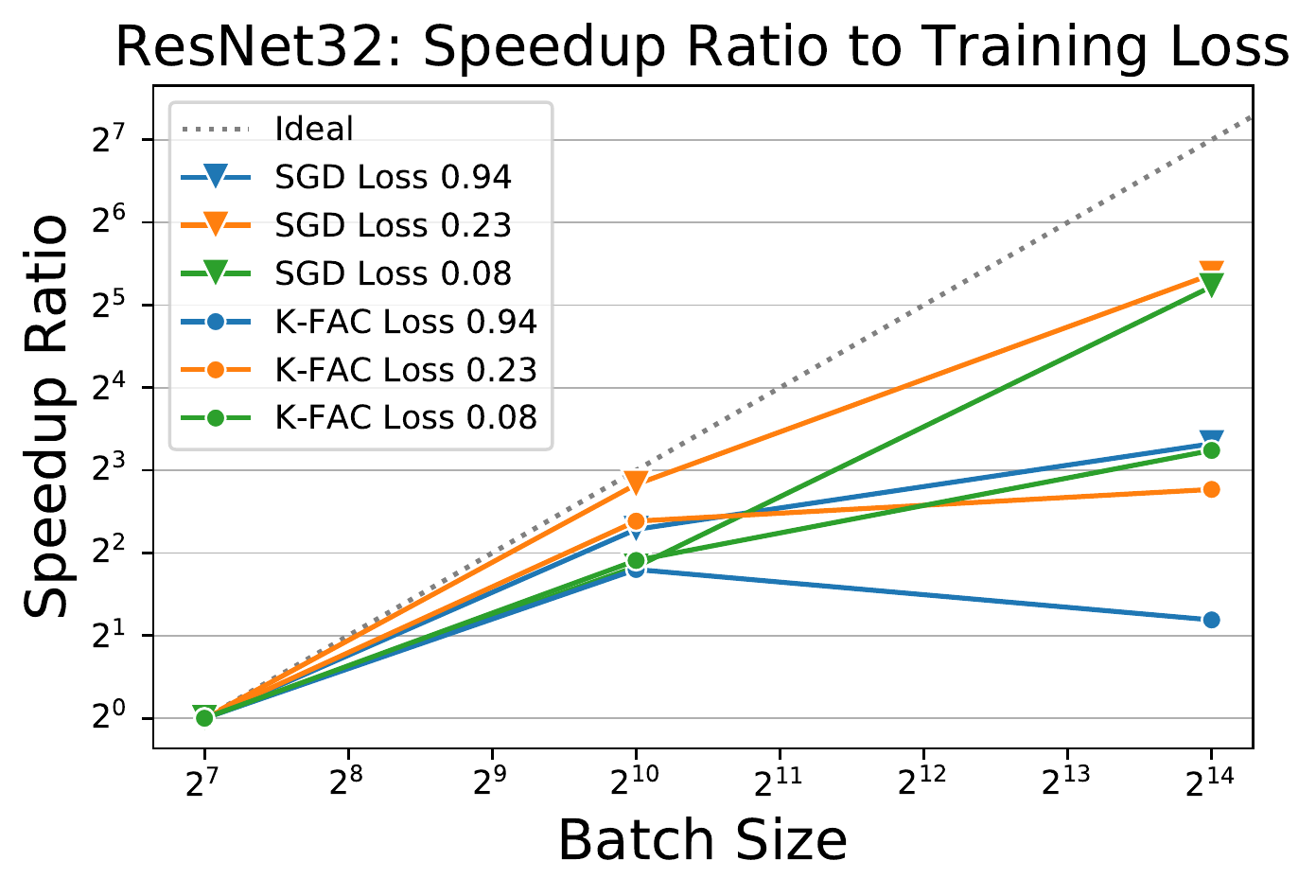}

\includegraphics[width=.35\textwidth]{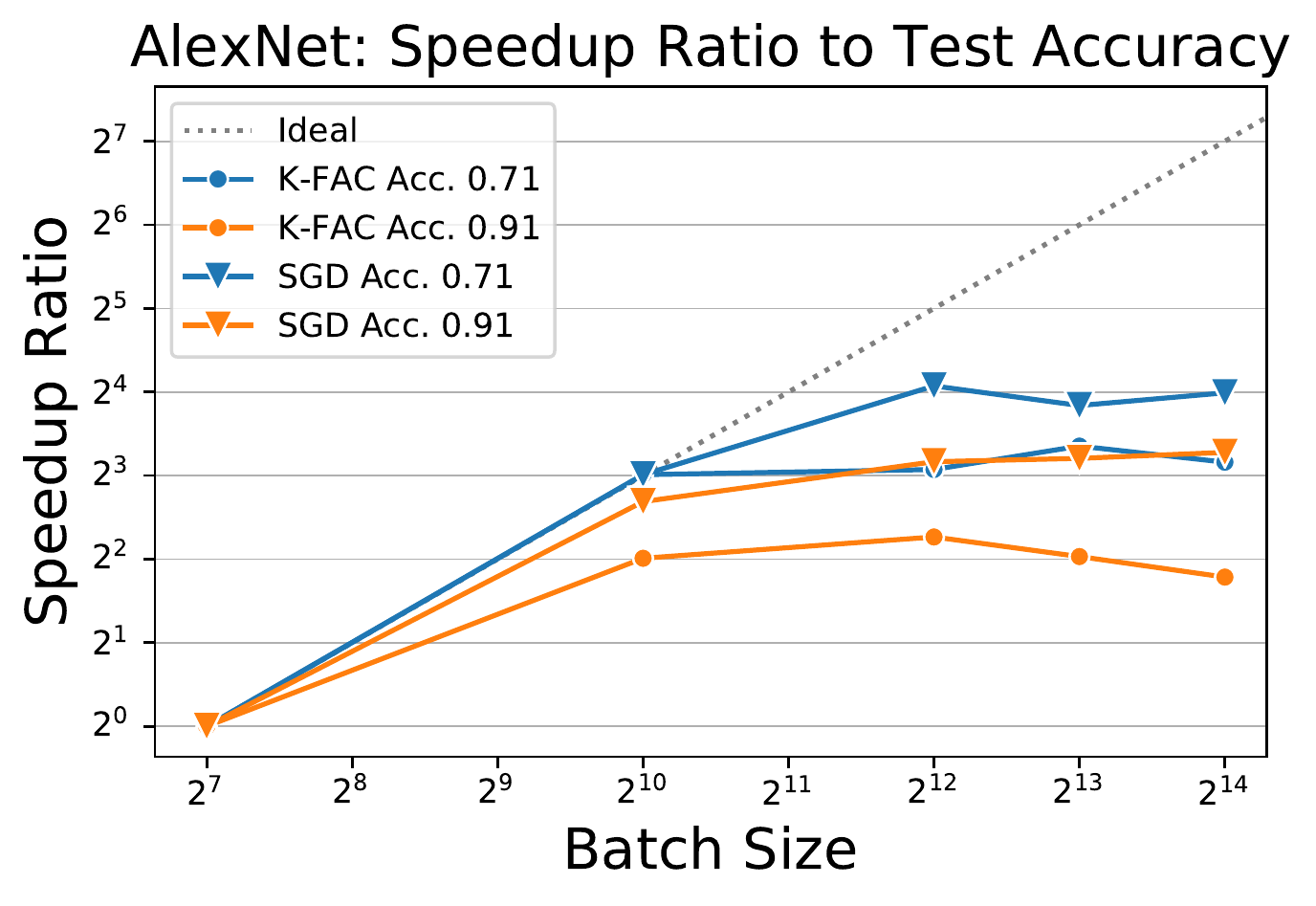}
  \includegraphics[width=.35\textwidth]{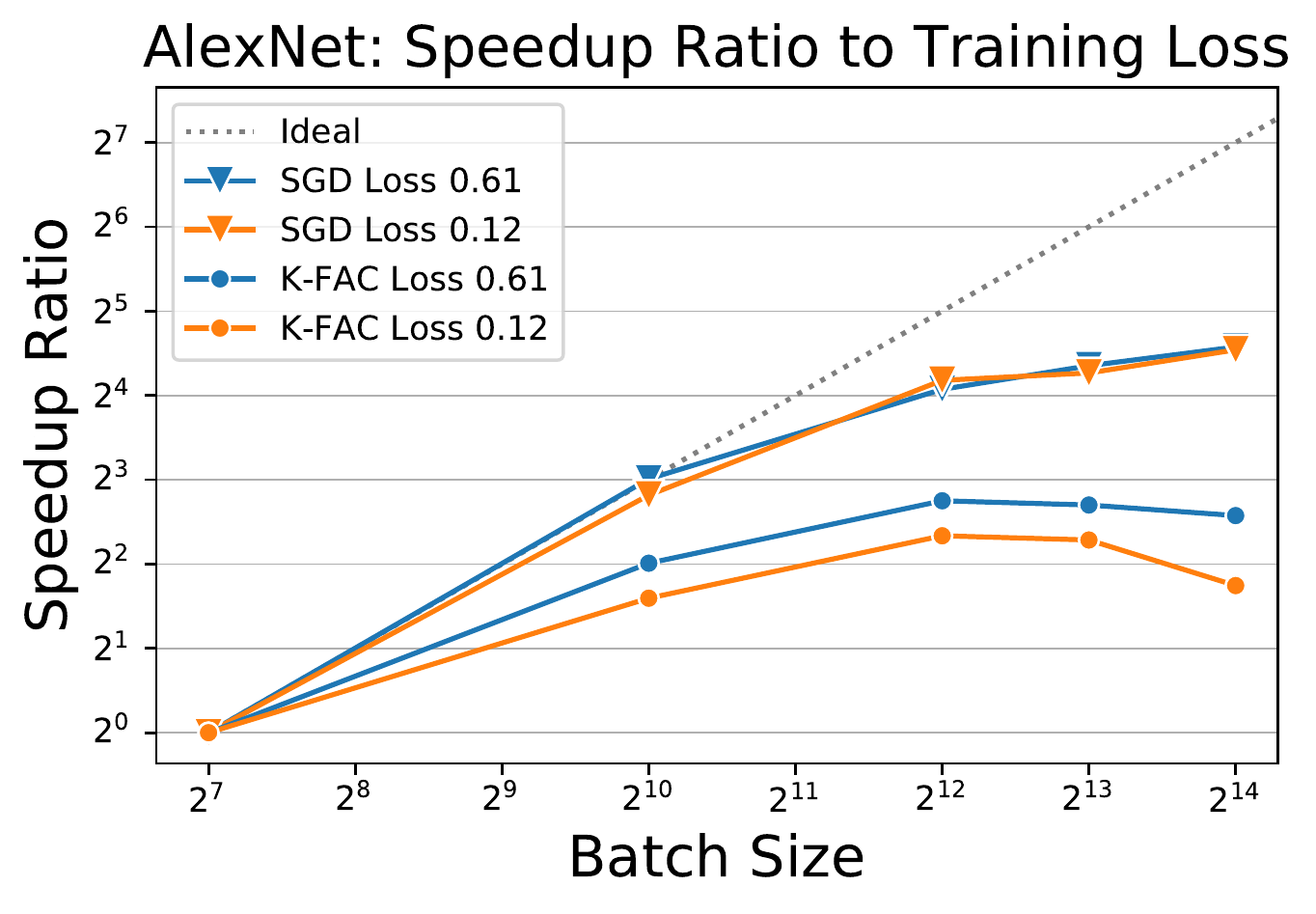}
\end{center}
\caption{From top to bottom: speed up to a target training loss / test accuracy versus batch size for both SGD and K-FAC with ResNet20 on CIFAR-10, ResNet32 on CIFAR-10 and AlexNet on SVHN, respectively.
The diminishing returns effect can be seen to be more prominent in K-FAC (circles) than in SGD (triangles).
}
\label{fig:combined_batch_loss_reduce}
\end{figure}

Training efficiency was measured for each batch size in terms of iterations to a target training loss or test accuracy ($k_{c}(m)$).
The speed up versus batch size relations are displayed in~\fref{fig:combined_batch_loss_reduce}.
Dotted lines denote the ideal scaling relationship between batch size and iterations.
We normalize each method-target line independently, dividing by the iterations at the smallest batch size $k(2^7)$ so that each of the dotted ideal lines is aligned in the plots, and we take the reciprocal to obtain the speedup function $s(m; 2^7) = k(2^7) / k(m)$, where $m$ is a given batch size.
We also present the iterations-batch size relations to same targets in Appendix~\ref{subsec:appendix-iterations-to-target}.
To ensure a fair comparison between batch sizes, similar to what is done in~\cite{Noah-EMP-CRIT-BS}, we select target loss values as follows:
\begin{itemize}[noitemsep,topsep=0.pt,parsep=0.pt,partopsep=0.pt,leftmargin=*]
    \item We wish to analyze how quickly using different batch sizes
    reaches a given threshold.  However, not all thresholds are feasible, since large batch sizes may never reach a low training loss, whereas small batches may reach it easily. 
    Thus, for speed up comparison purposes, we set thresholds such that all the
    batch sizes can reach it.  We choose a \textit{selected run} that belongs to the worst-performing batch size and method. This selection is made
    after loss-based hyperparameter tuning is finished. 
    \item Then, for each training 
stage of the selected run\footnote{
Learning rate decays separate the training process into stages as defined before.}, we generate a training loss/test accuracy target, using the value that is linearly interpolated at $80\%$ between the loss/accuracy values directly before and directly after the training stage.
\end{itemize}

We use the resulting target values in
\fref{fig:combined_batch_loss_reduce}.
For both K-FAC and SGD, diminishing return effects are present. 
In all examined cases, K-FAC deviates from ideal scaling (dotted lines) to a greater extent than SGD, as batch size increases. This difference explains why in~\fref{fig:combined_batch_v_acc} SGD is increasingly able to outperform K-FAC for large batches, given a fixed budget.
We note that for both SGD and K-FAC, the linear scaling regime is largely nonexistent, particularly for the highest-performance targets; diminishing returns begin immediately from the smallest batch size.  \fref{fig:combined_batch_loss_reduce} provides evidence contrary to the conjectures of~\cite{GOOG-LB-KFAC-HYPO}, which posit that K-FAC may exhibit a larger regime of perfect scaling than SGD.
The per-iteration training trajectories supporting
\fref{fig:combined_batch_loss_reduce} can be observed in Appendix~\ref{subsec:appendix-train-test-curve}.


\subsection{Hyperparameter Sensitivity of K-FAC}\label{sec:robustness_kfac}

\begin{figure*}[!t]
\begin{center}
\includegraphics[width=.32\textwidth]{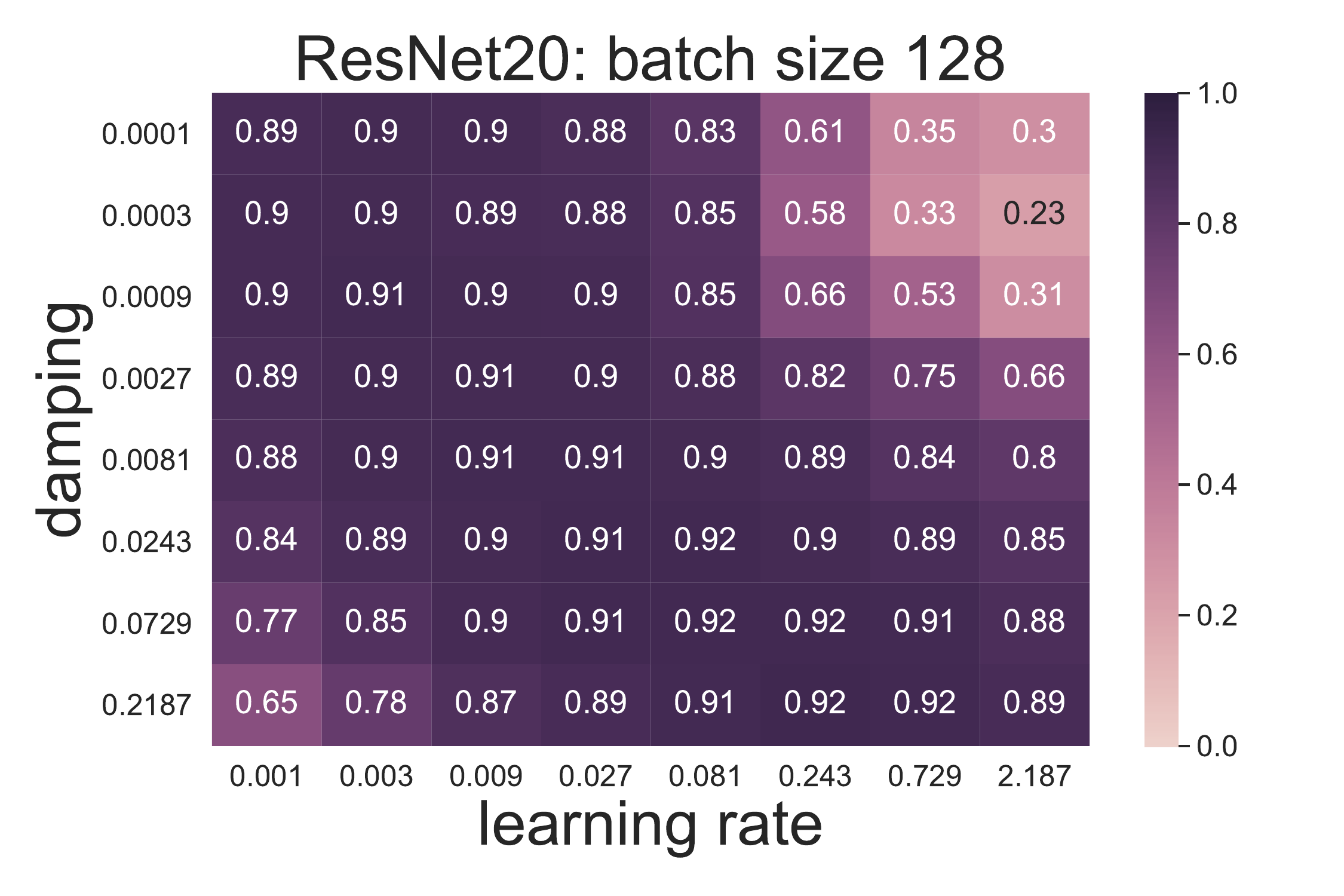}
\includegraphics[width=.32\textwidth]{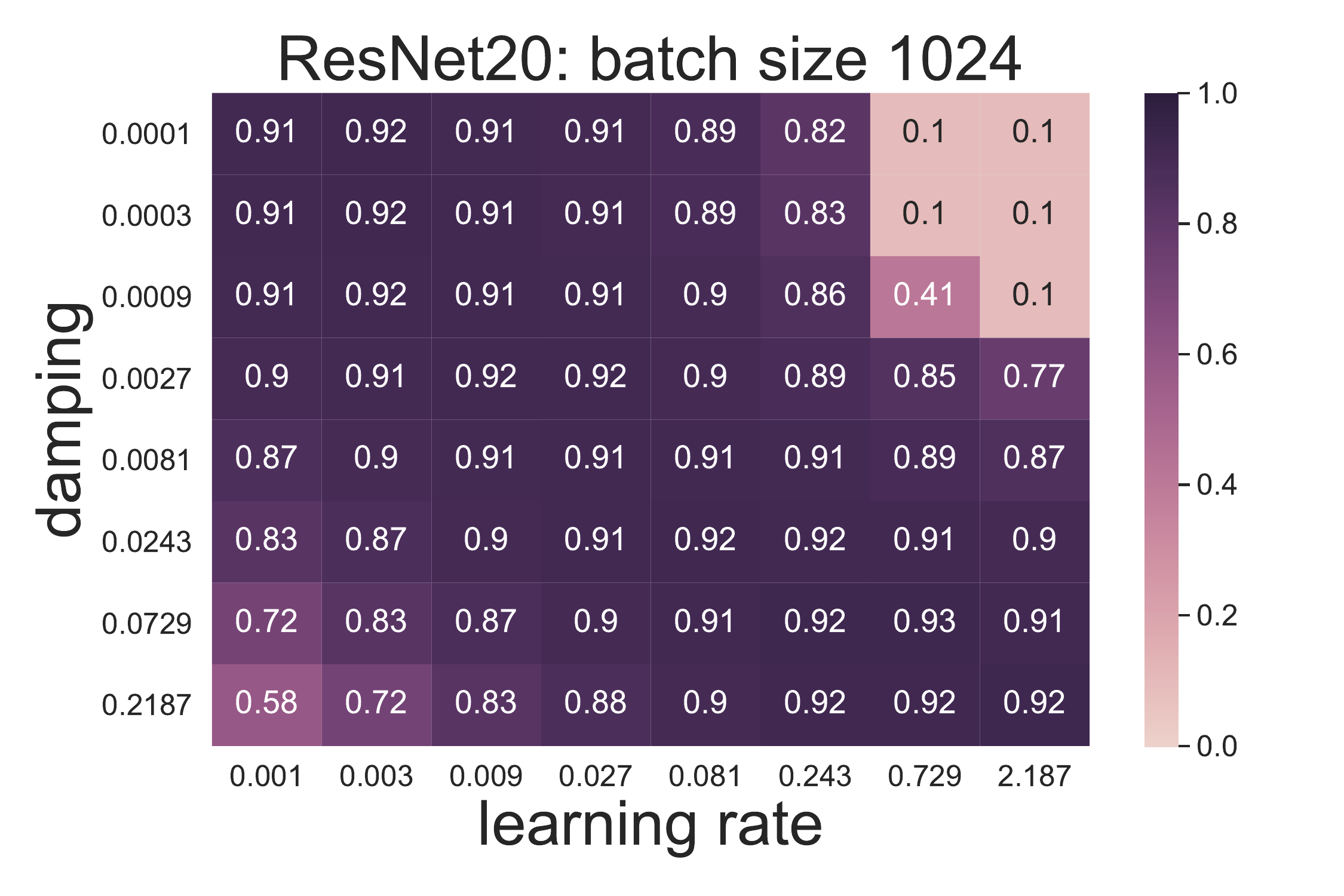}
\includegraphics[width=.32\textwidth]{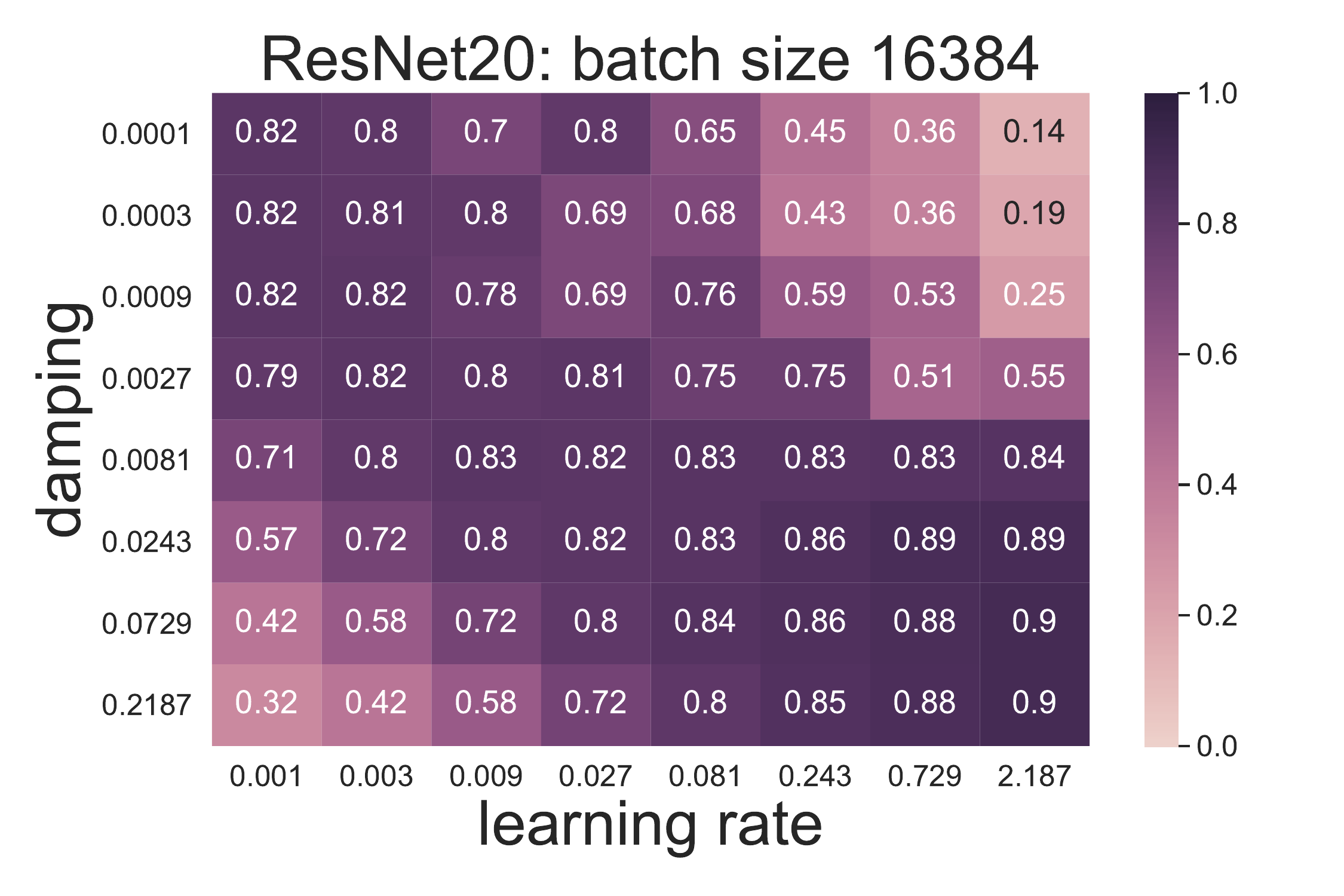}

\includegraphics[width=.32\textwidth]{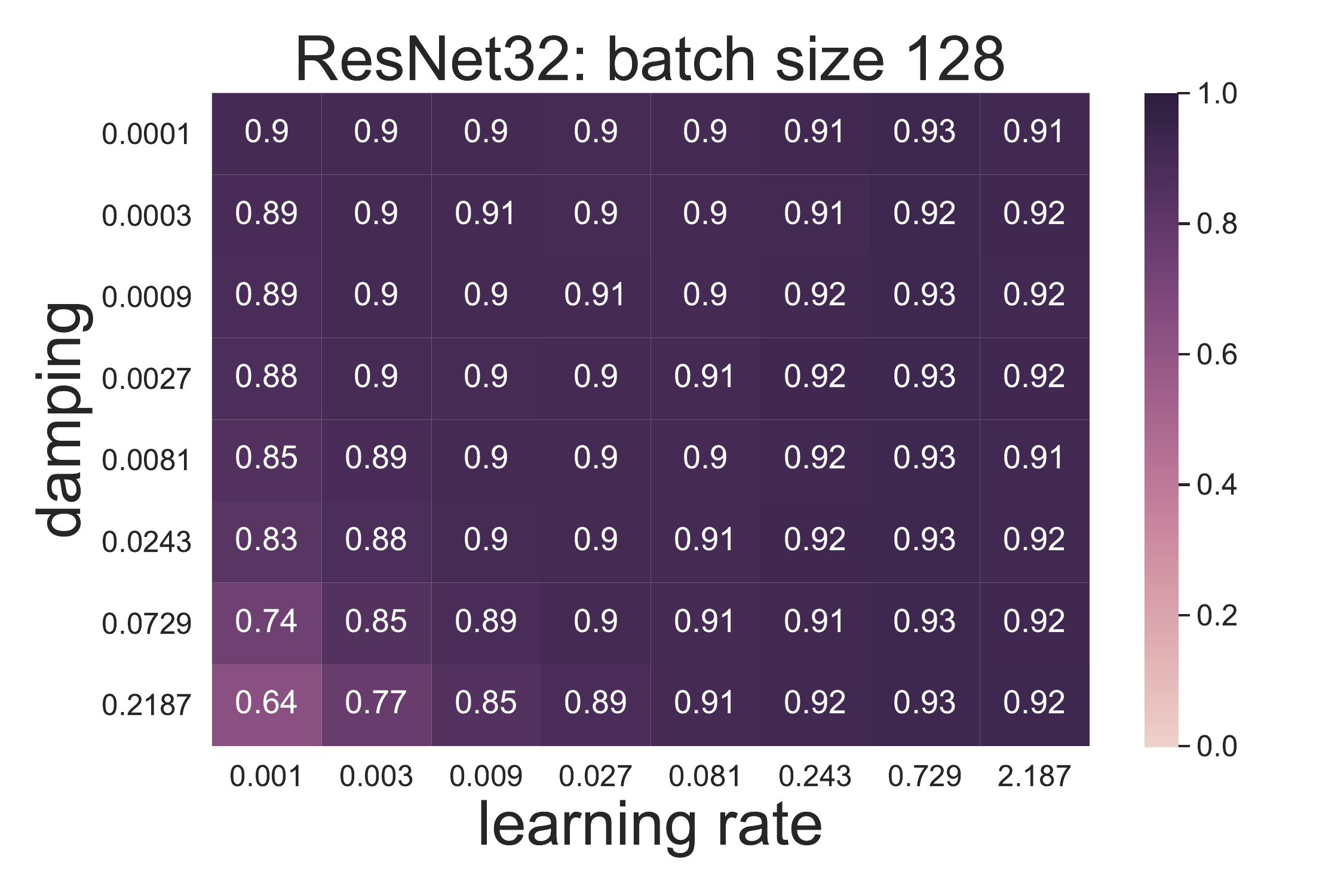}
\includegraphics[width=.32\textwidth]{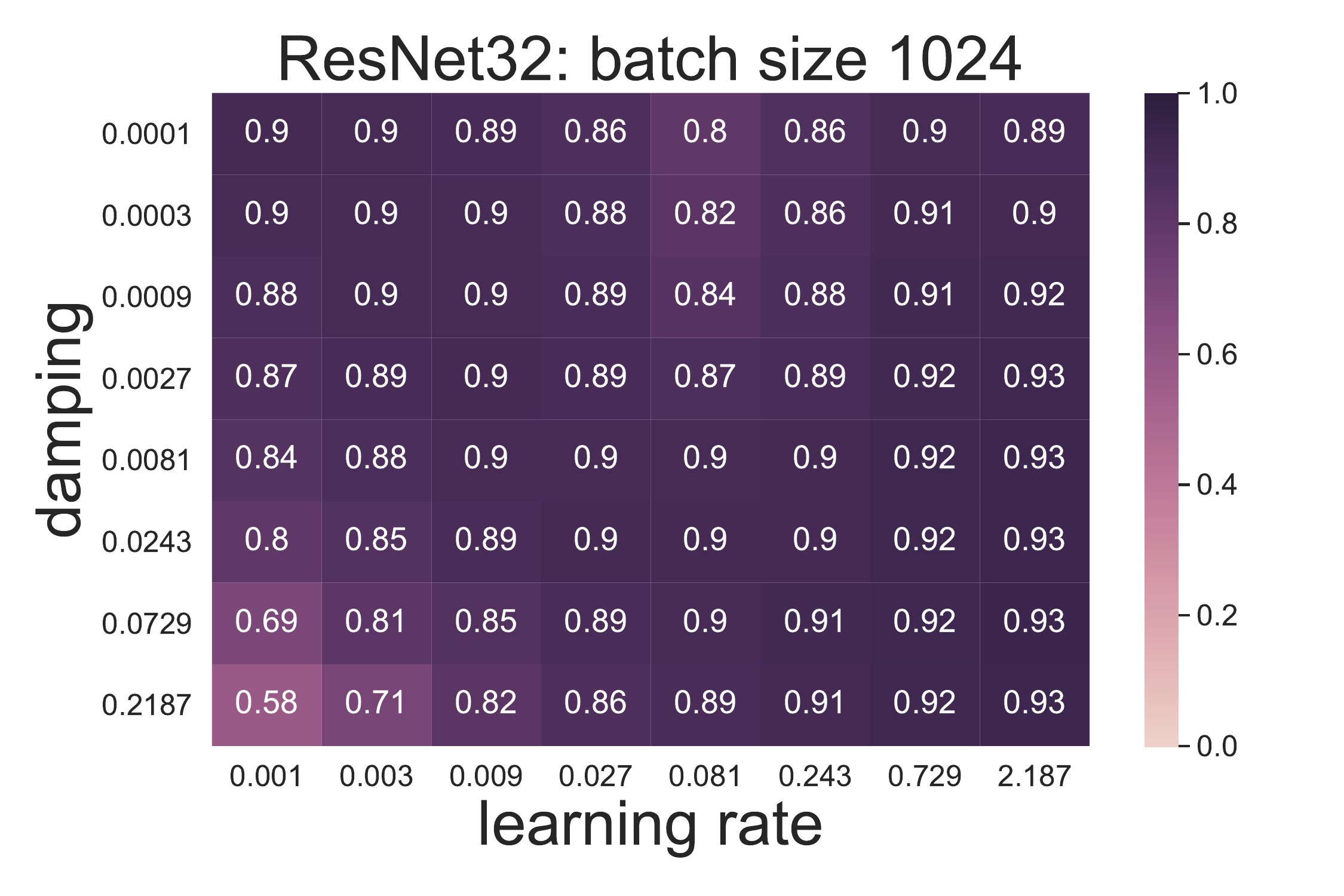}
\includegraphics[width=.32\textwidth]{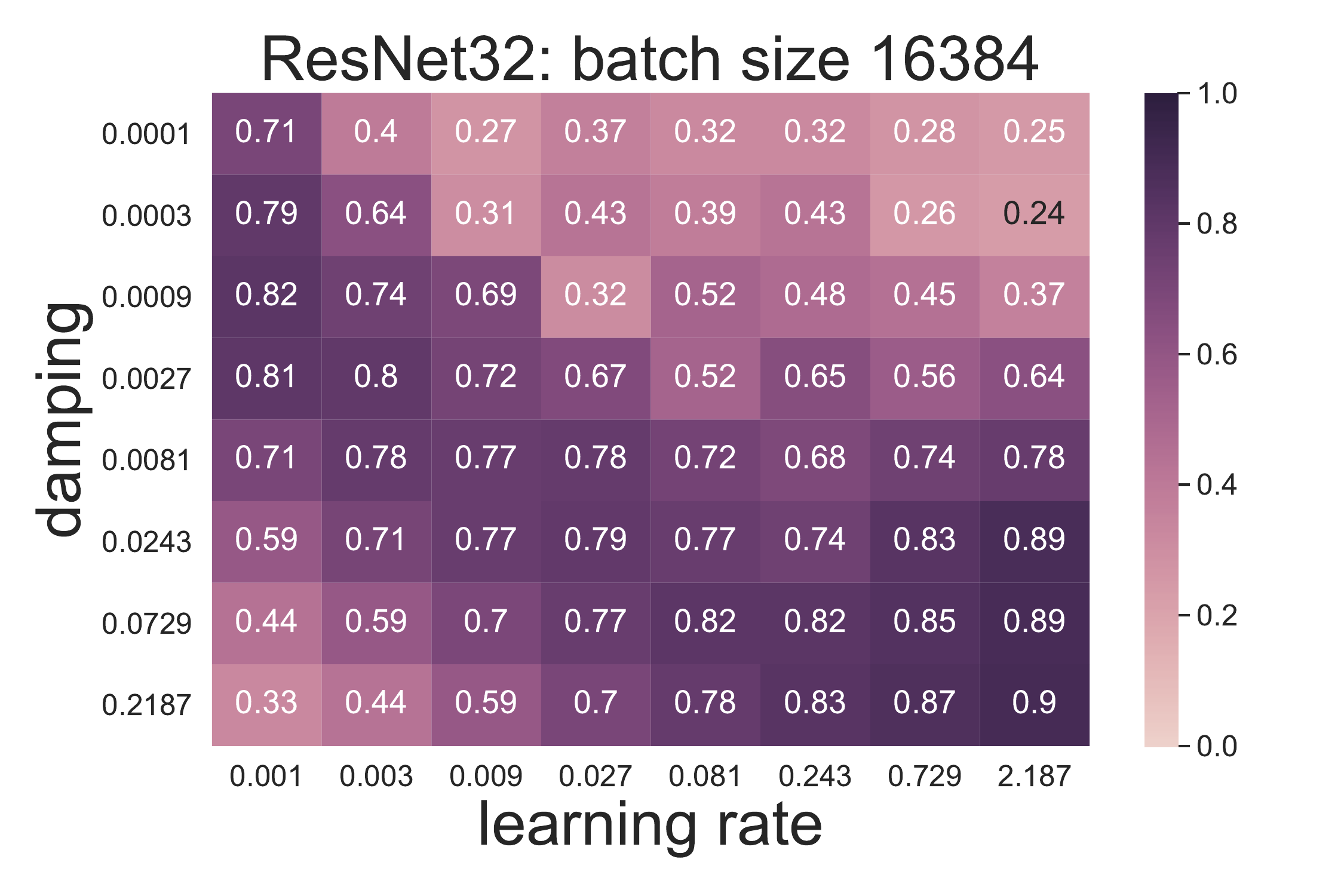}

\includegraphics[width=.32\textwidth]{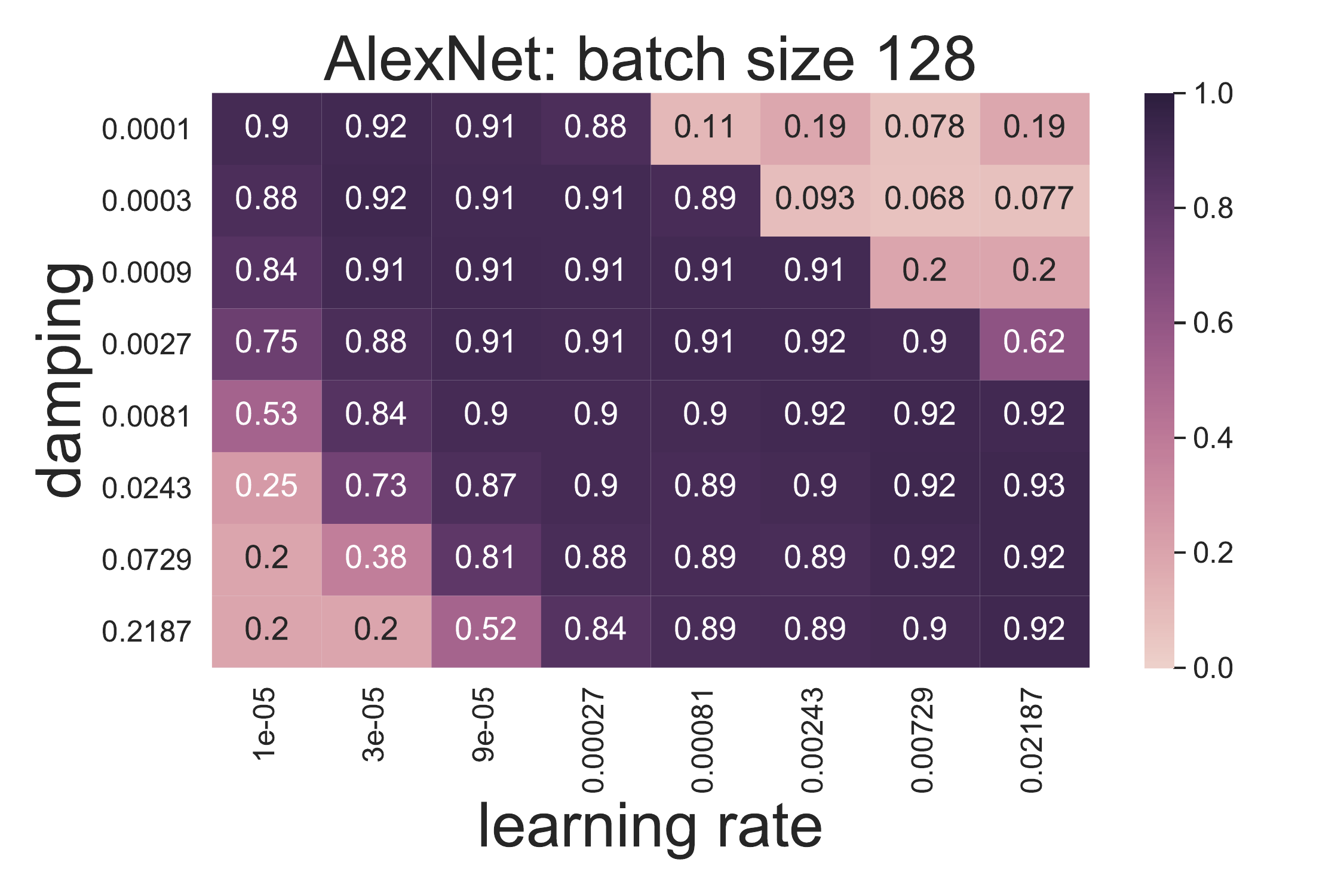}
\includegraphics[width=.32\textwidth]{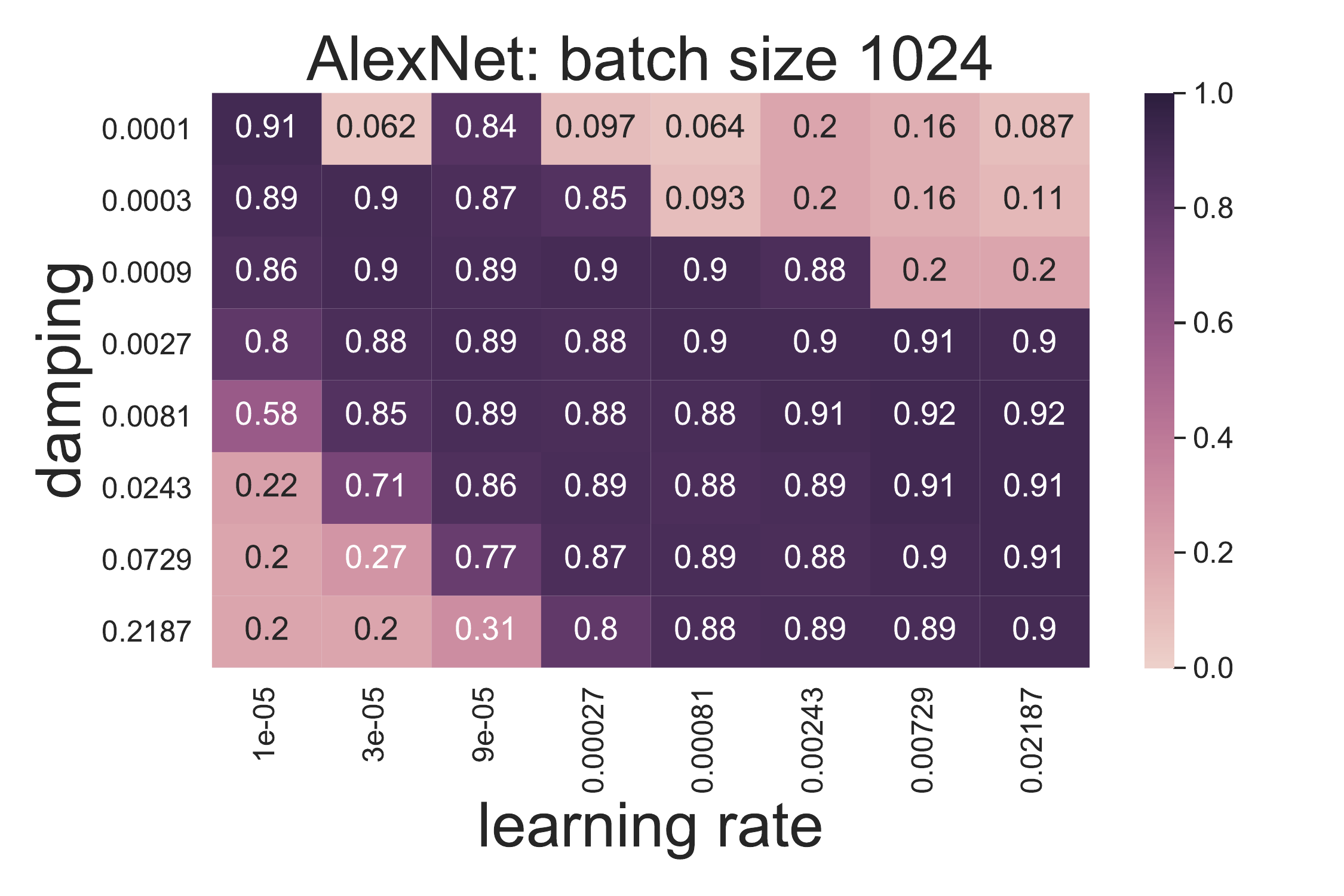}
\includegraphics[width=.32\textwidth]{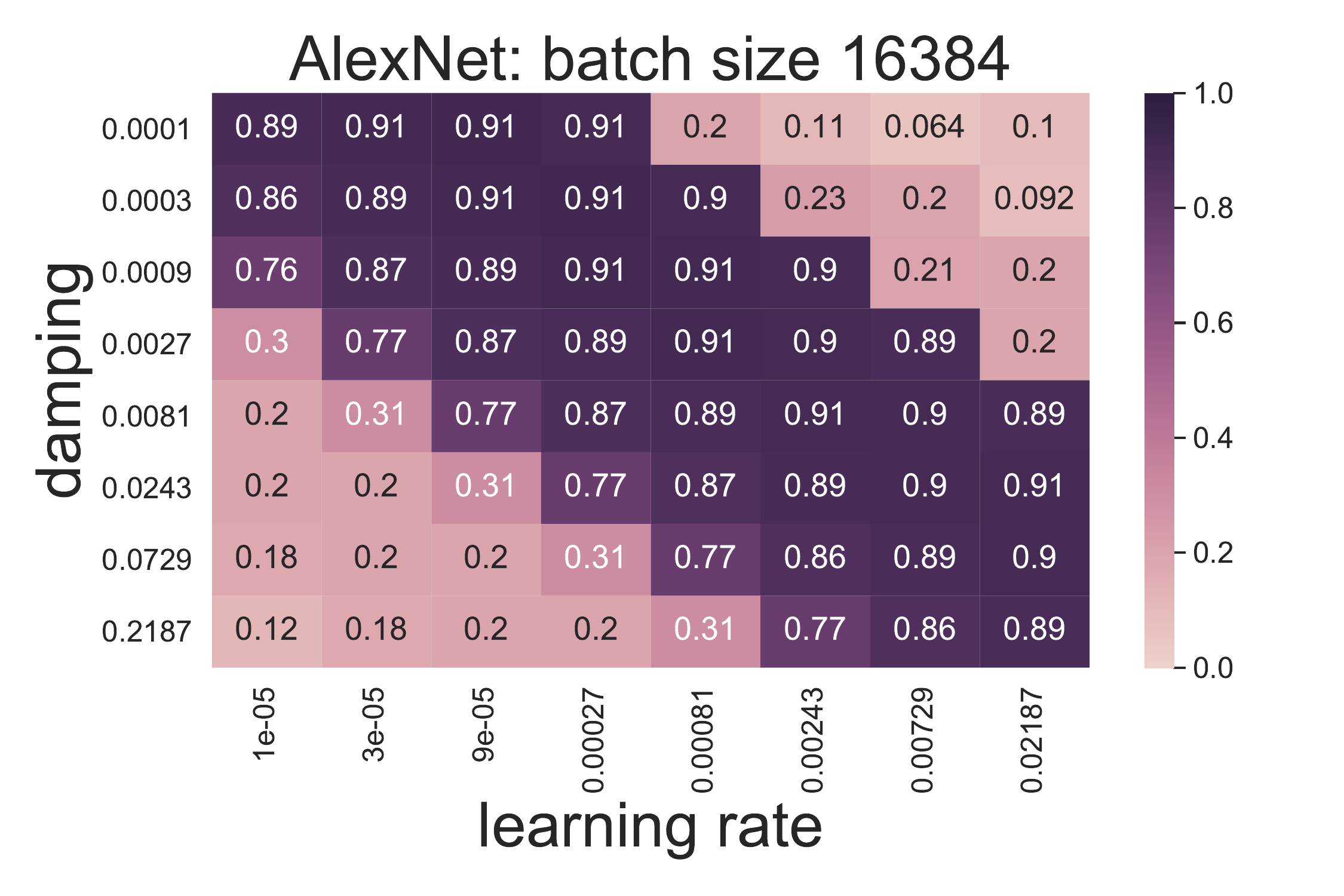}
\end{center}
\caption{From top to bottom: accuracy at end of training under adjusted-epoch budget versus damping and learning rate for batch sizes 
128, 1,024
and 16,384 for CIFAR-10 with ResNet20, CIFAR-10 with ResNet-32, SVHN with AlexNet, respectively.
A positive correlation between damping and learning rate is exhibited, as well as a shrinking of the high-accuracy region for large batch sizes. }
\label{fig:heatmap}
\end{figure*}

\begin{figure*}[!htp]
\begin{center}
  \subfloat[]{\label{subfig:box1}\includegraphics[width=.31\textwidth]{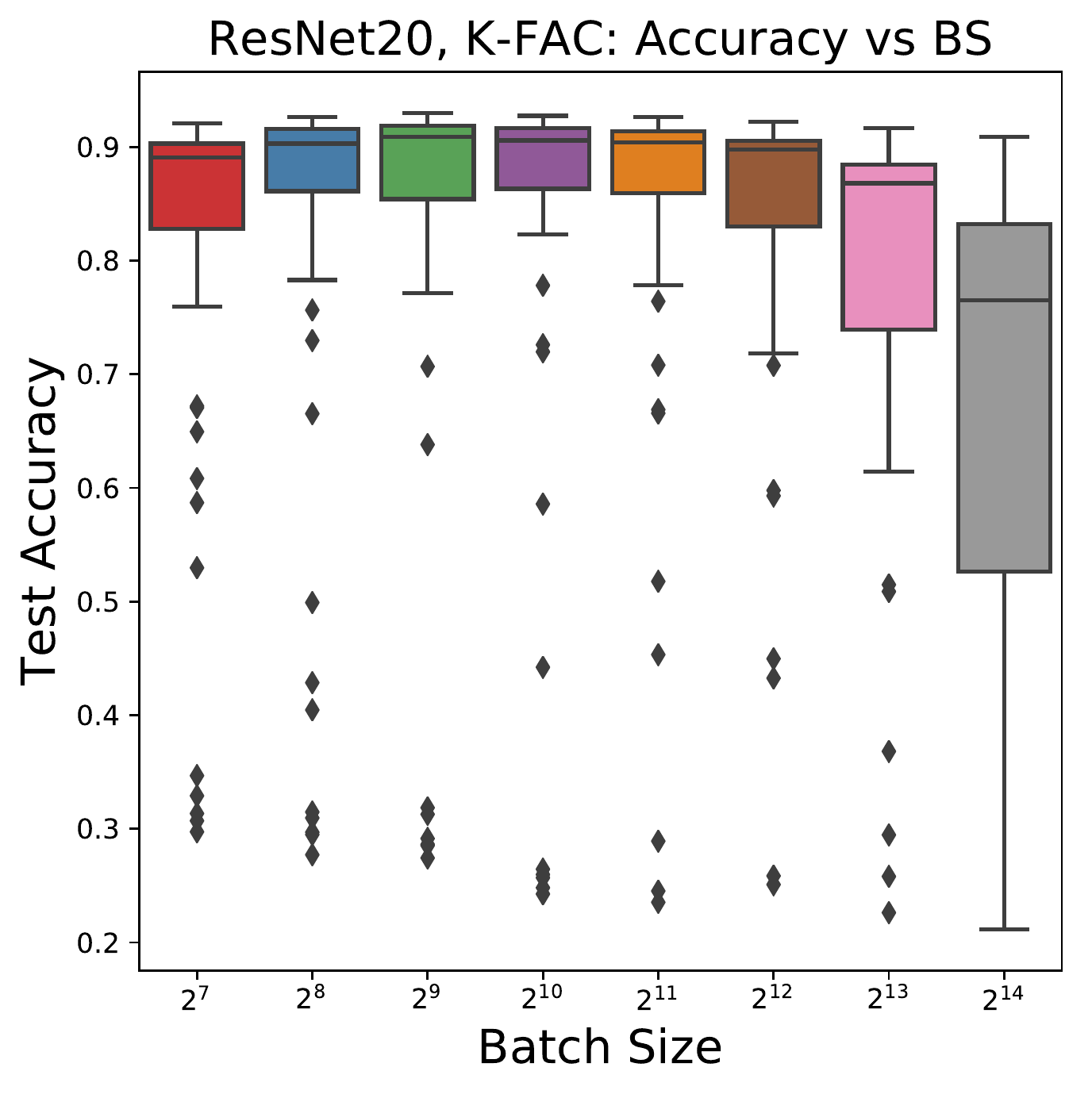}}
  \subfloat[]{\label{subfig:box2}\includegraphics[width=.31\textwidth]{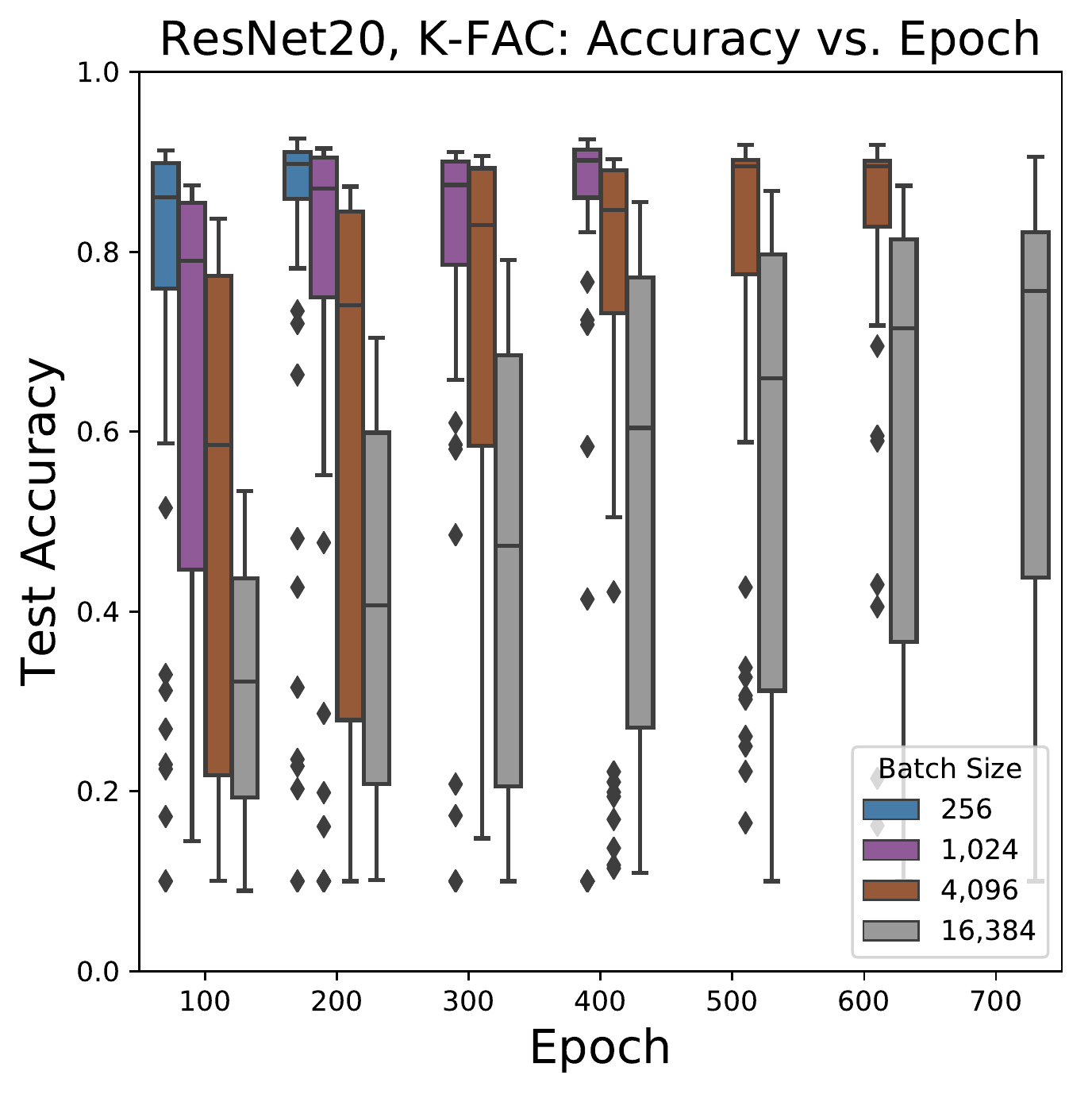}}
  \subfloat[]{\label{subfig:box3}\includegraphics[width=.31\textwidth]{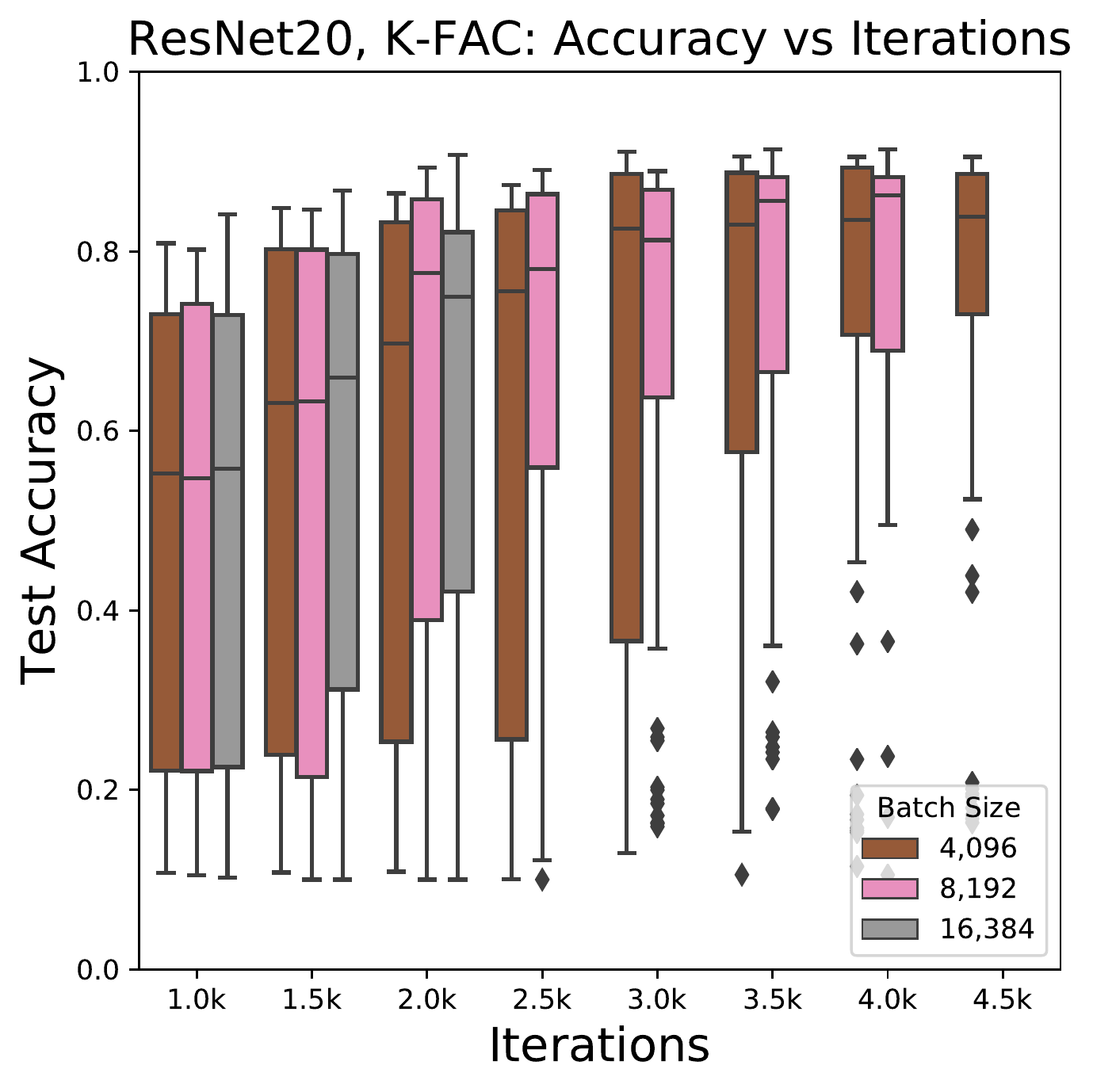}}

  \subfloat[]{\label{subfig:box4}\includegraphics[width=.31\textwidth]{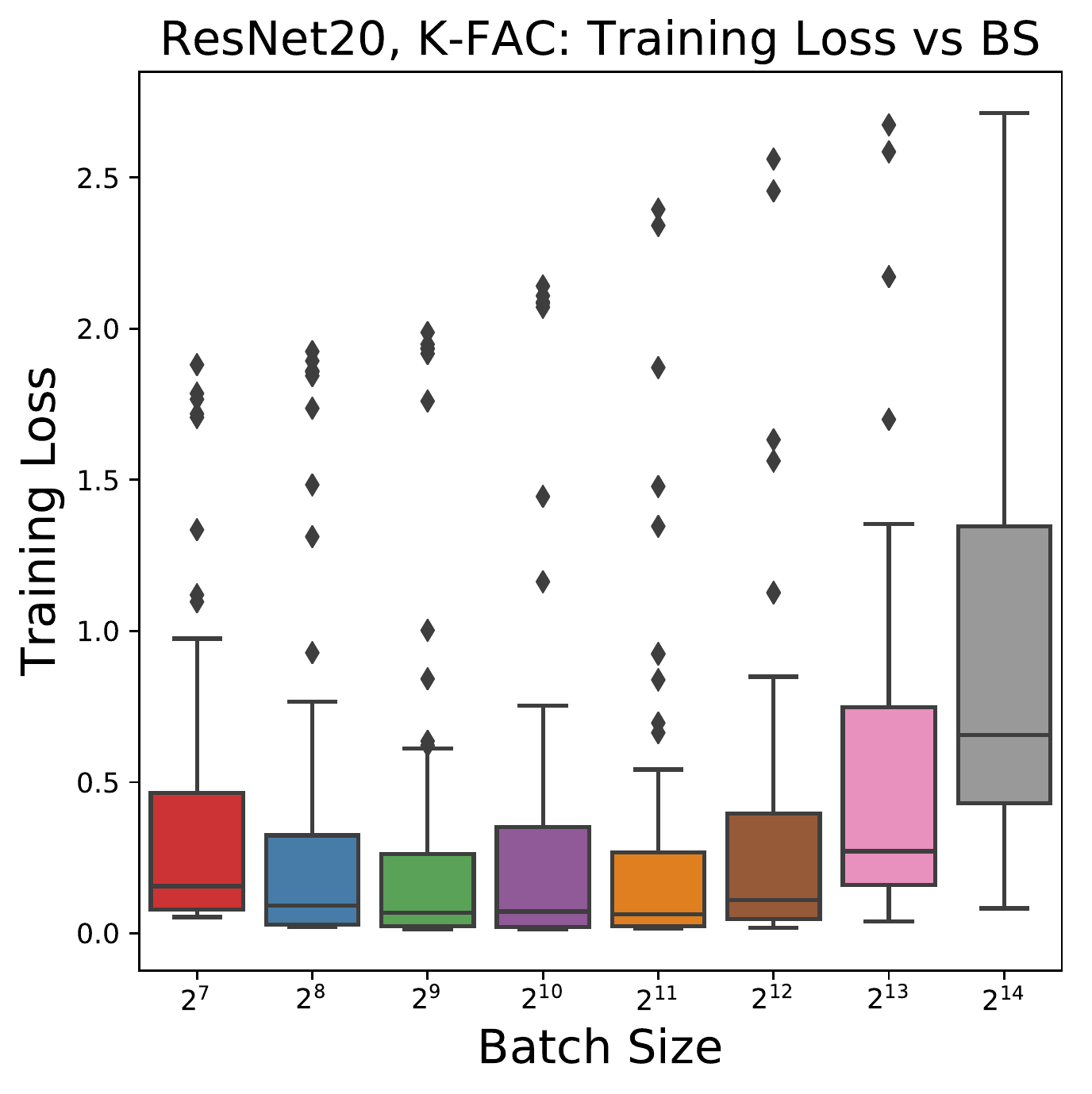}}
  \subfloat[]{\label{subfig:box5}\includegraphics[width=.31\textwidth]{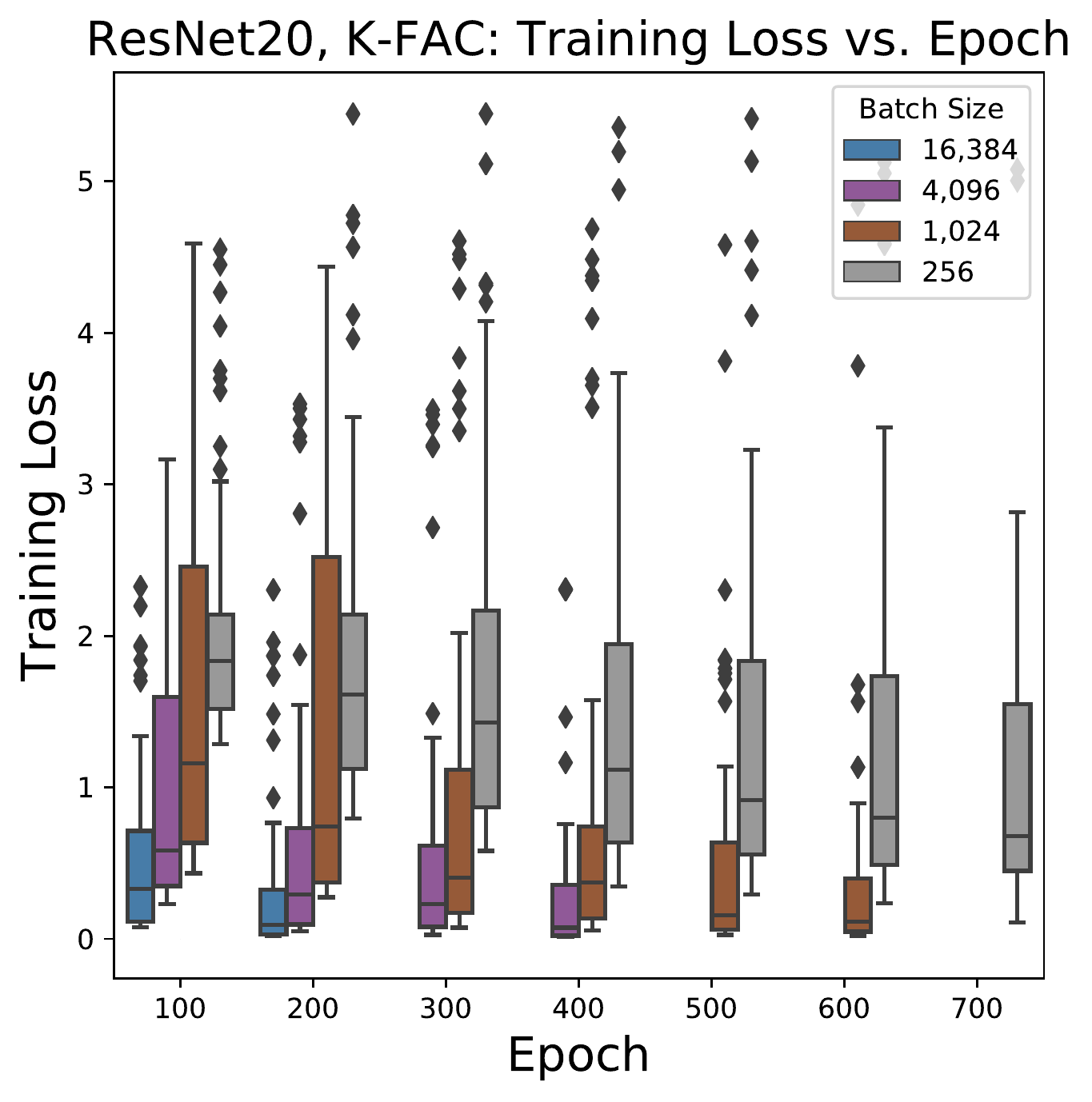}}
  \subfloat[]{\label{subfig:box6}\includegraphics[width=.31\textwidth]{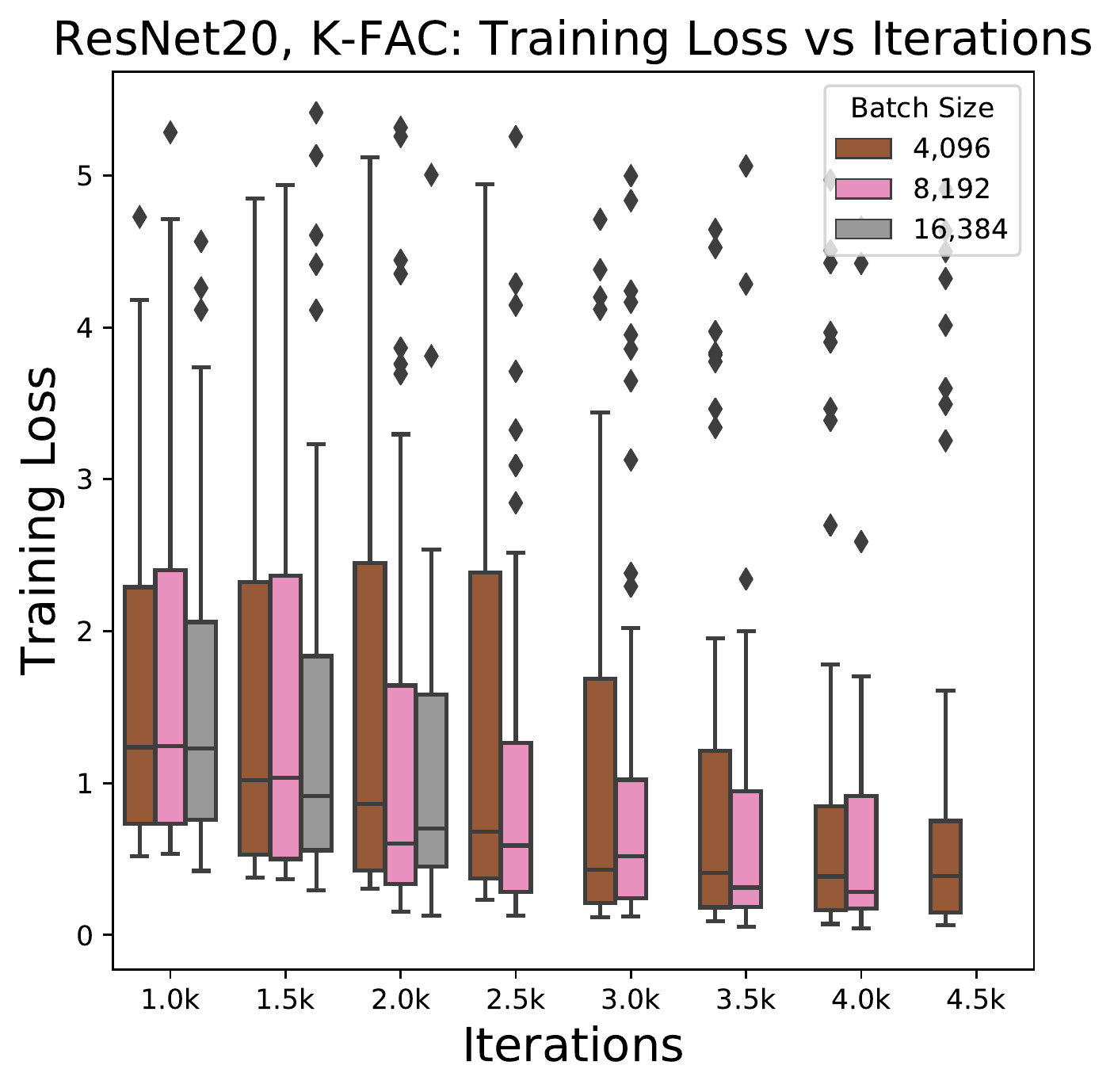}}
\end{center}
\caption{\textbf{(a)(d):} Test accuracy / training loss distribution vs. batch size for K-FAC at the end of training under an adjusted epoch budget. 
Larger batch sizes result in lower accuracies and higher training losses that are more sensitive to hyperparameter choice. 
\textbf{(b)(e):} Comparison of test accuracy / training loss distributions over various epochs.
Smaller batch sizes provide better solutions that are less sensitive to choice of hyperparameters. 
\textbf{(c)(f):} Comparison of test accuracy / training loss distributions over various iteration numbers.
Large batch sizes exhibit a trend of providing better solutions that are less sensitive to choice of hyperparameters. }
\label{fig:batch_v_acc_bp}
\end{figure*}

The hyperparameter tuning spaces on all three models for K-FAC are laid out in~\fref{fig:heatmap}, which relates the selected hyperparameters for damping and learning rate with test accuracy achieved under the adjusted epoch budget. The complete list of heatmaps for K-FAC hyperparameter tuning are provided in Appendix~\ref{subsec:appendix-heatmap}. All heatmaps we observe 
demonstrate a consistent trend in terms of:
\begin{itemize}[noitemsep,topsep=0.pt,parsep=0.pt,partopsep=0.pt,leftmargin=*]
    \item A positive correlation between damping and learning rate.
    \item A shrinking of the high-accuracy region with increasing batch size.
\end{itemize}
The second point suggests a relationship between batch size and hyperparameter sensitivity for K-FAC, which can be measured in terms of the volume of hyperparameter space corresponding to successful training. In evaluating hyperparameter sensitivity (or inversely robustness), we take the approach of~\cite{GOOG-LB-KFAC-HYPO}, distinguishing between two types of robustness, each corresponding to a different definition of ``successful training'':
\begin{itemize}[noitemsep,topsep=0.pt,parsep=0.pt,partopsep=0.pt,leftmargin=*]
    \item Epoch-based robustness, in which success is defined by training to a desired accuracy or loss within a fixed number of epochs.
    \item Iteration-based robustness, in which success is defined by training to a desired accuracy or loss within a fixed number of iterations.
\end{itemize}
It is important to note that a set of hyperparameters considered to be acceptable to a practitioner under an iteration budget may at the same time be considered unacceptable to a practitioner operating under an epoch budget. It is for this reason that we make this distinction.

Through this lens, the robustness behavior of K-FAC with ResNet20 on CIFAR-10 is exhibited in~\fref{fig:batch_v_acc_bp}. We describe the robustness behavior of the other two models in Appendix~\ref{subsec:appendix-robustness-behavior}. Distributions of training accuracy across hyperparameters are represented by box plots composed from the 64 hyperparameter configurations (8 damping parameters, 8 learning rate parameters). In~\fref{subfig:box1} and~\ref{subfig:box4},
we show the distributions of test accuracies and training losses 
for each batch size at the end of training under the adjusted epoch budget. The greater spread of accuracies observed for larger batch sizes indicates that given this budget we should consider batch sizes from $2^{12}$ to $2^{14}$ as more sensitive to hyperparameter tuning than batch sizes from $2^8$ to $2^{11}$. Informally, if we draw a horizontal line at a desired test accuracy of, e.g., 0.8, then the batch sizes with boxplots containing the majority of their hyperparameter distribution above the 0.8 line should be favored as being more robust.

We can simulate stopping of training in terms of epochs and iterations to extract insight about robustness for other types of budgets than our own. Regardless of the stopping criteria, we expect that longer training will yield greater robustness (although at the cost of significantly higher computational/budget overhead).
Figure~\ref{subfig:box2} and~\ref{subfig:box5} 
shows how the hyperparameter robustness of different batch sizes changes as a function of stopping epoch.
Each group along the X-axis corresponds to a hypothetical epoch budget. The relationship demonstrates that for K-FAC, \textit{robustness increases with amount of training, but more interestingly it decreases with batch size}. This can be observed by noting that for any fixed epoch, the distributions of accuracies corresponding to larger batch sizes fall lower  than their smaller-batch counterparts, meaning fewer hyperparameter configurations will fall above a desired accuracy threshold. A similar robustness trend is observed with the distributions of training losses.
We perform a similar analysis for iteration budgets in Fig.~\ref{subfig:box3} and~\ref{subfig:box6}, 
and we find as expected that robustness increases with training. Unlike the case of an epoch budget however, we find that for iteration budgets \textit{robustness increases with larger batch size}. This is observed by noting that the distributions corresponding to large batch are more concentrated towards higher accuracy and lower loss, although the effect is not as pronounced as in the fixed epoch case. 

Together, the results show that (i): epoch-based robustness is inversely related to batch size, and (ii): iteration-based robustness is \textit{directly} related to batch size.  This is analogous to the findings of~\cite{GOOG-LB-KFAC-HYPO} for SGD. Further work may seek to compare the nature of iteration-based or epoch-based robustness between the two methods in more detail.

%% file: s5_conclusion.tex
\section{Conclusions}\label{sec:conclusion}

Through extensive experimentation and training on both \mbox{CIFAR-10} and SVHN datasets, we find that K-FAC  exhibits similar diminishing returns with large batch training as SGD. 
In our results, K-FAC has comparable but not better training or testing performance than SGD given the same level of training and tuning. 
Comparing the scalability behaviors of the two methods, we find that K-FAC exhibits a smaller regime of ideal scaling than SGD, suggesting that K-FAC's scalability to large batch training is no better than SGD's. 
Finally, we find that K-FAC exhibits a similar relationship between budget and robustness as SGD, in which K-FAC is less robust to tuning under epoch budgets, but more robust to tuning under iteration budgets, mirroring the findings of similar work in literature for SGD~\cite{GOOG-LB-KFAC-HYPO}. 
Taken as a whole, our results suggest that, although K-FAC has been applied to large batch training scenarios, it encounters the same large-batch issues to an equal or greater extent as SGD.
It remains to be seen whether other variants of sub-sampled Newton methods~\cite{RM16a_TR,RM16b_TR} can lead to improved results or whether this is a more ubiquitous aspect of stochastic optimization algorithms applied to non-convex optimization problems of interest in machine~learning.

%% file: figure_dump.tex

\newpage

\appendix

\section{Appendix}

In this appendix, we provide additional details about our main results.

\subsection{Scaled Learning Rate Decay Schedule}\label{sec:comparison_fix_scale}

\begin{figure*}[!t]
\begin{center}
\subfloat[]{\label{subfig:schedule_1}\includegraphics[width=.45\textwidth]{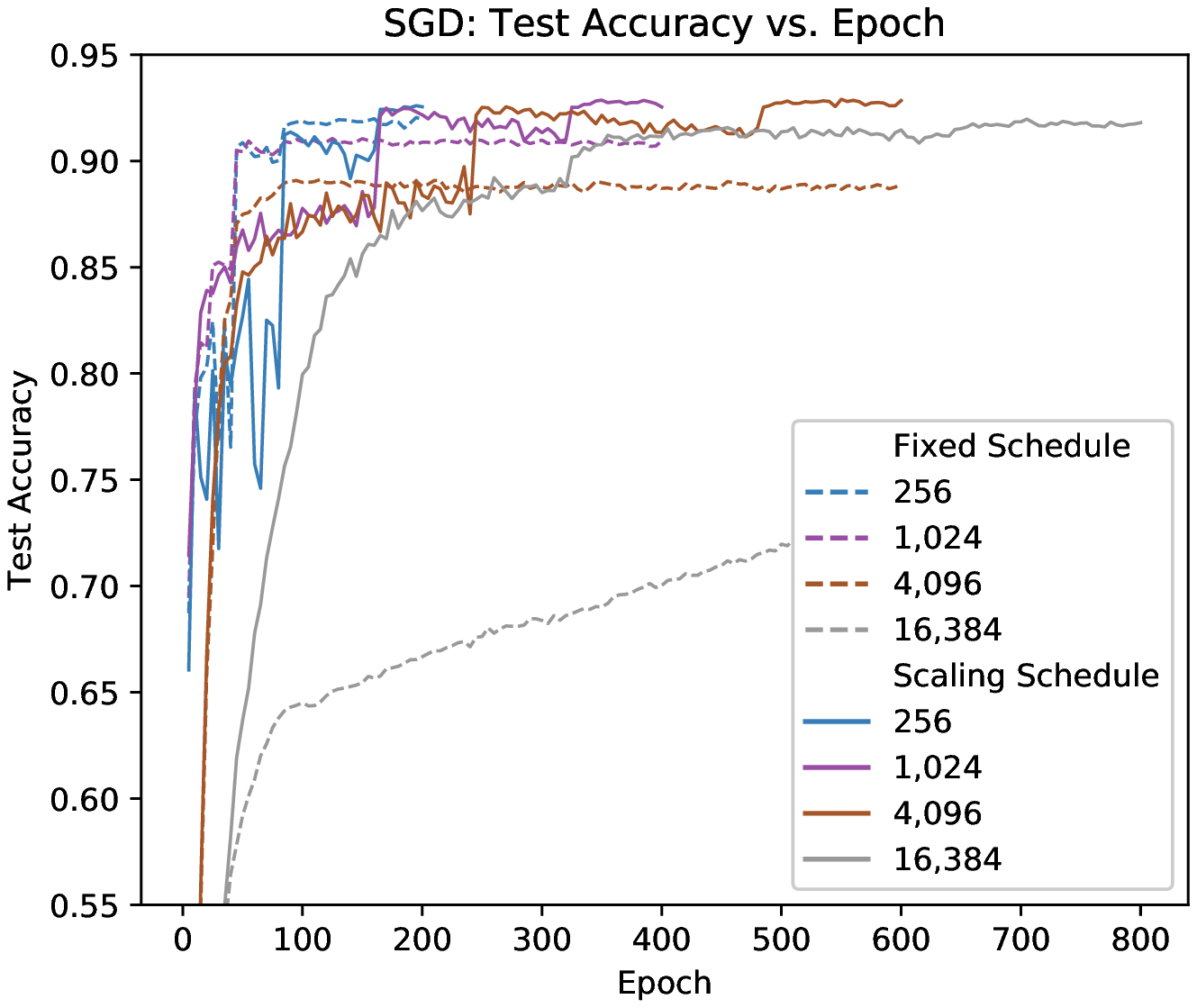}}
\subfloat[]{\label{subfig:schedule_2}\includegraphics[width=.45\textwidth]{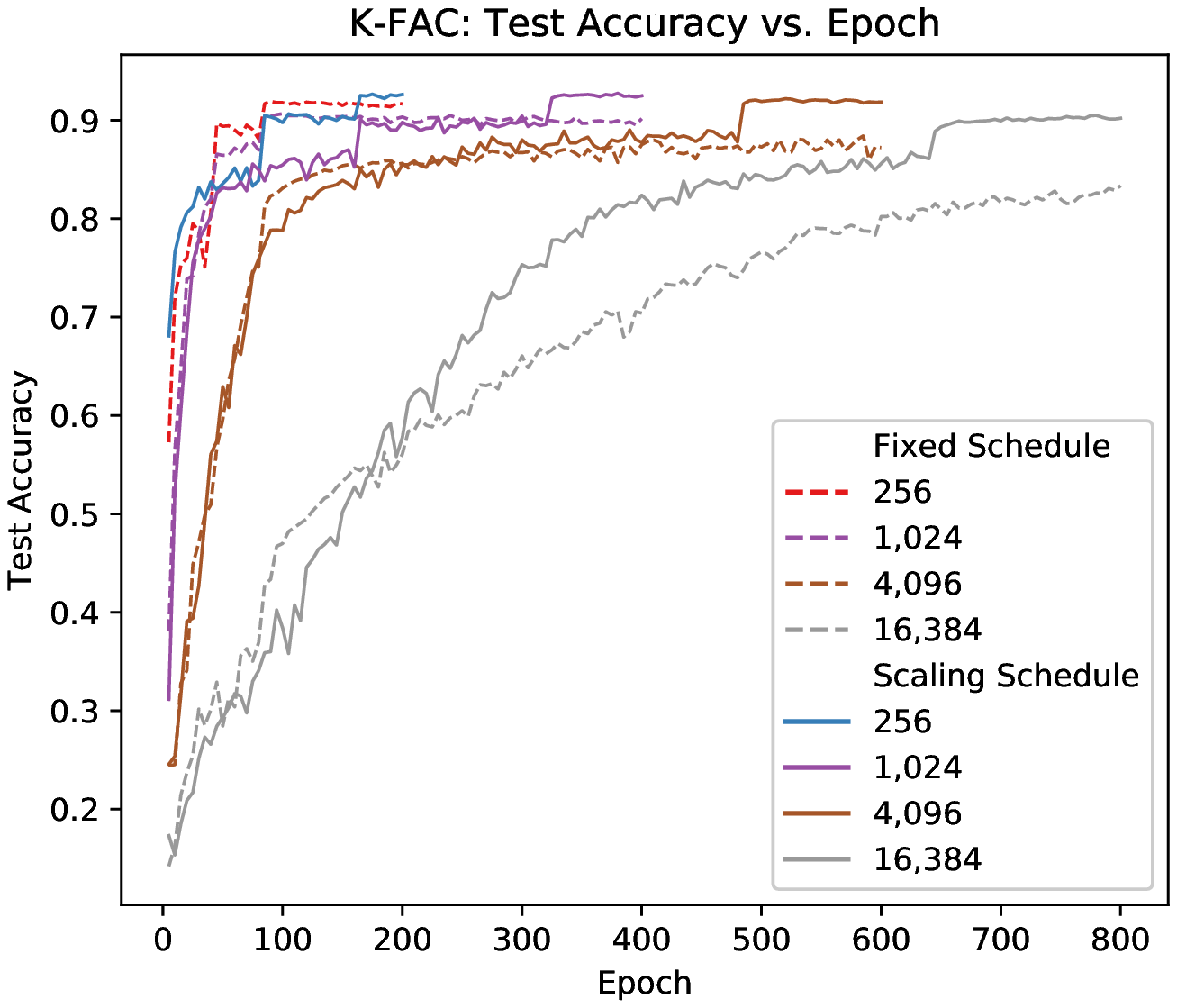}}

\subfloat[]{\label{subfig:schedule_3}\includegraphics[width=.45\textwidth]{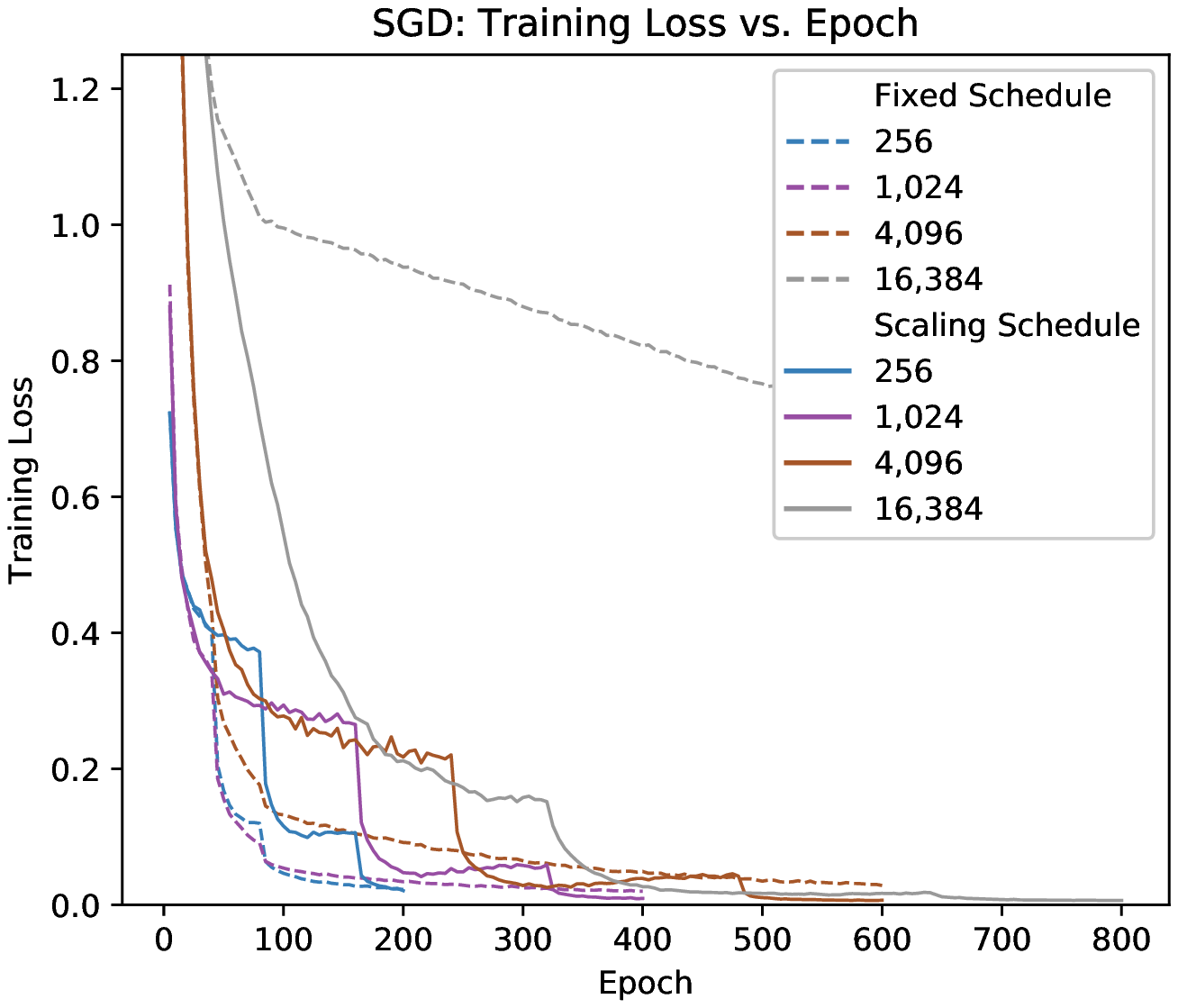}}
\subfloat[]{\label{subfig:schedule_4}\includegraphics[width=.45\textwidth]{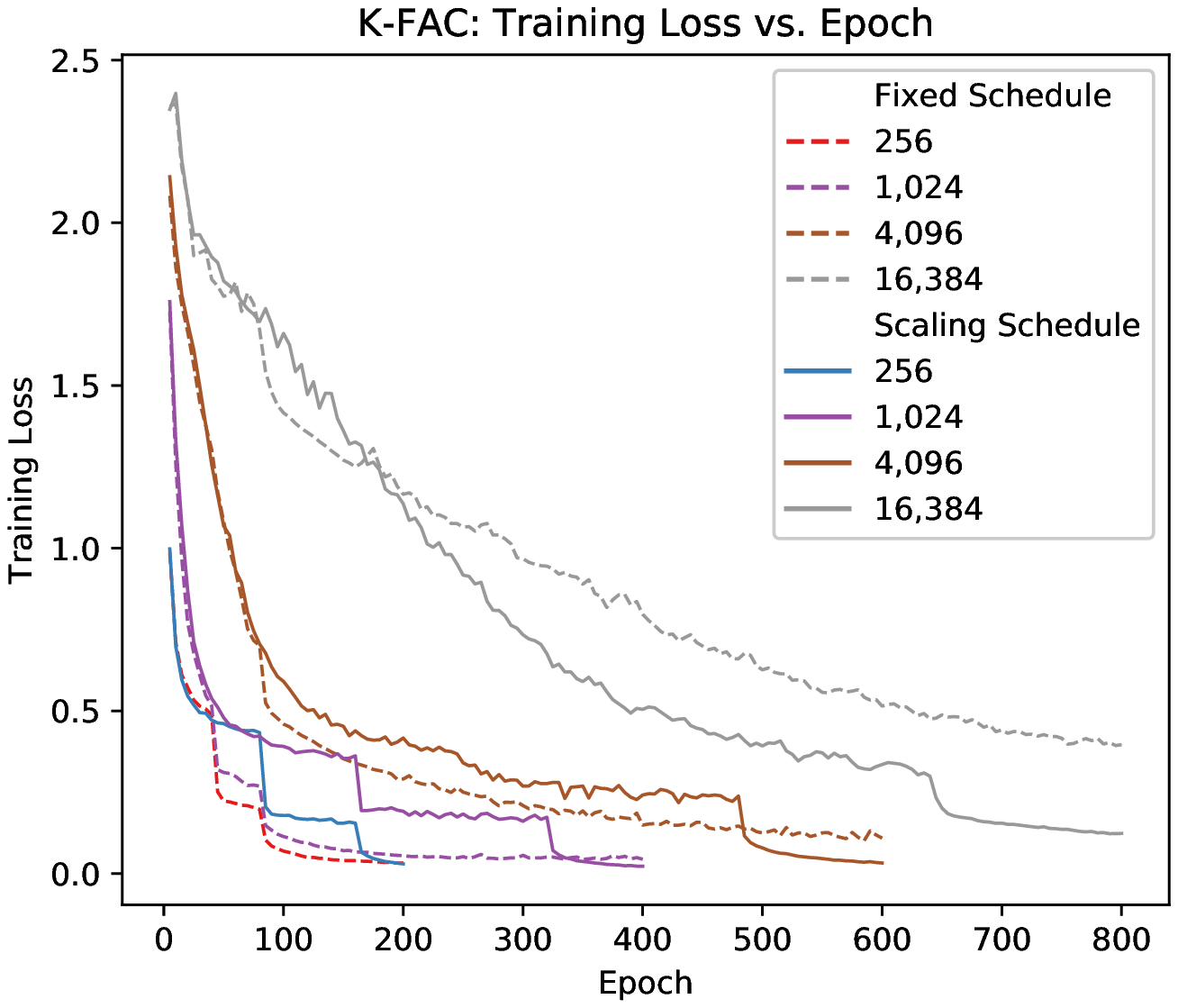}}
\end{center}
\caption{Learning rate schedule comparison for ResNet20 on CIFAR-10. (a)(b): Test accuracy versus epoch for SGD and K-FAC on a fixed learning rate schedule (dashed) and scaled learning rate schedule (solid).
(c)(d): Training loss versus epoch for SGD and K-FAC on a fixed learning rate schedule (dashed) and scaled learning rate schedule (solid).
For each batch size and schedule, other hyperparameters are chosen separately to maximize test accuracy or minimize loss. The scaled learning rate schedule significantly favors large batch sizes. }
\label{fig:sgd_fixed_scaling_epoch_acc}
\end{figure*}

Here, we compare training trajectories under fixed and scaled learning rate schedules to provide validation of our experimental setup.
We evaluate the scaled learning rate schedule discussed in Section~\ref{sec:setup} against a fixed learning rate schedule using ResNet20 on the \mbox{CIFAR-10} dataset. Extensive grid search is applied for both schedules, giving rise to best-performing runs which are illustrated in~\fref{fig:sgd_fixed_scaling_epoch_acc}. 

In Fig.~\ref{subfig:schedule_1} and~\ref{subfig:schedule_3}, we plot the highest-accuracy and lowest-loss training runs (chosen across the all hyperparameter configurations) for fixed learning rate schedule versus scaled learning rate schedule for SGD. Across all batch sizes, scaled learning rate schedule training gives rise to a higher final test accuracy and lower final training loss. For large batch sizes (e.g., 16K), the difference is more pronounced, with fixed learning rate schedule SGD making much slower progress than the scaled learning rate schedule counterpart. In particular, for the 16K batch size with fixed learning rate schedule, after 80 epochs it can be observed that the early decay severely slows the previously-rapid progress since the learning rate becomes too small. As a result the fixed-scaling large batch SGD slows its climb in accuracy, while the scaled learning rate schedule variant surges~ahead.

In Fig.~\ref{subfig:schedule_2} and~\ref{subfig:schedule_4}, we plot the analogous runs for K-FAC, with decays following a fixed versus scaled learning rate schedule. Similarly, scaled learning rate schedule K-FAC demonstrates higher end-of-training accuracy and lower loss across all batch sizes.
Based on these observations, we argue that our experimental setup is geared towards boosting large-batch scalability through its use of a scaled schedule for learning rate~decay.

\subsection{Normal versus Approximated Damping}
\label{subsec:damping}

Here, we explain the differences between two preconditioning mechanisms for K-FAC: \textit{normal damping}, as written in Eqn.~(\ref{equation:damping}); and \textit{approximated damping}, as recommended by~\cite{KFAC-G15}. We also provide an empirical evaluation of the two methods, and we observe that approximate damping exhibits hyperparameter sensitivity that is comparable to normal damping. First, we briefly discuss how to calculate directly the inverse of the preconditioned Fisher Information Matrix (preconditioned FIM), denoted $(F + \lambda I)^{-1}$. The preconditioned FIM block can be calculated as 
\begin{align*}
    (F_i + \lambda I)^{-1}  &= [(Q_A D_A Q_A^\top) \otimes (Q_G D_G Q_G^\top) + \lambda I  ]^{-1} \\
    &= [(Q_A \otimes Q_G) (D_A \otimes D_G + \lambda I)  (Q_A^\top \otimes Q_G^\top)]^{-1} \\
    &= (Q_A^\top  \otimes Q_G^\top ) (D_A \otimes D_G + \lambda I)^{-1}  (Q_A \otimes Q_G),
\end{align*}
where $Q_A D_A Q_A^\top$ is the eigendecomposition of $\mathbb{E}[A_{i-1}A_{i-1}^\top ]$ and $Q_G D_G Q_G^\top$ is the eigendecomposition for $\mathbb{E}[G_{i}G_{i}^\top ]$. A similar derivation can be found in Appendix A.2 in~\cite{grosse2016kronecker}. In our paper, we call the use of this damping \textit{normal damping}.

For efficiency reasons, \cite{KFAC-G15} proposes an alternative calculation to alleviate the burden of eigendecomposition by preconditioning on the Kronecker-factored Fisher blocks first:  
\begin{align*}
    (F_i + \lambda I)^{-1} \approx (\mathbb{E}[A_{i-1}A_{i-1}^\top ] + \sqrt{\lambda} I) ^{-1} \otimes (\mathbb{E}[G_iG_i^\top] + \sqrt{\lambda} I)^{-1}, 
\end{align*}
where $\lambda$ is a compound term that allows for complicated maneuvers of adaptive regularization (detailed in Section 6.2~\cite{KFAC-G15}). In our paper, we call this damping formulation \textit{approximated damping}.

\begin{figure*}[h]
\begin{center}
\subfloat[]{\label{subfig_damg_compare1}\includegraphics[width=.44\textwidth]{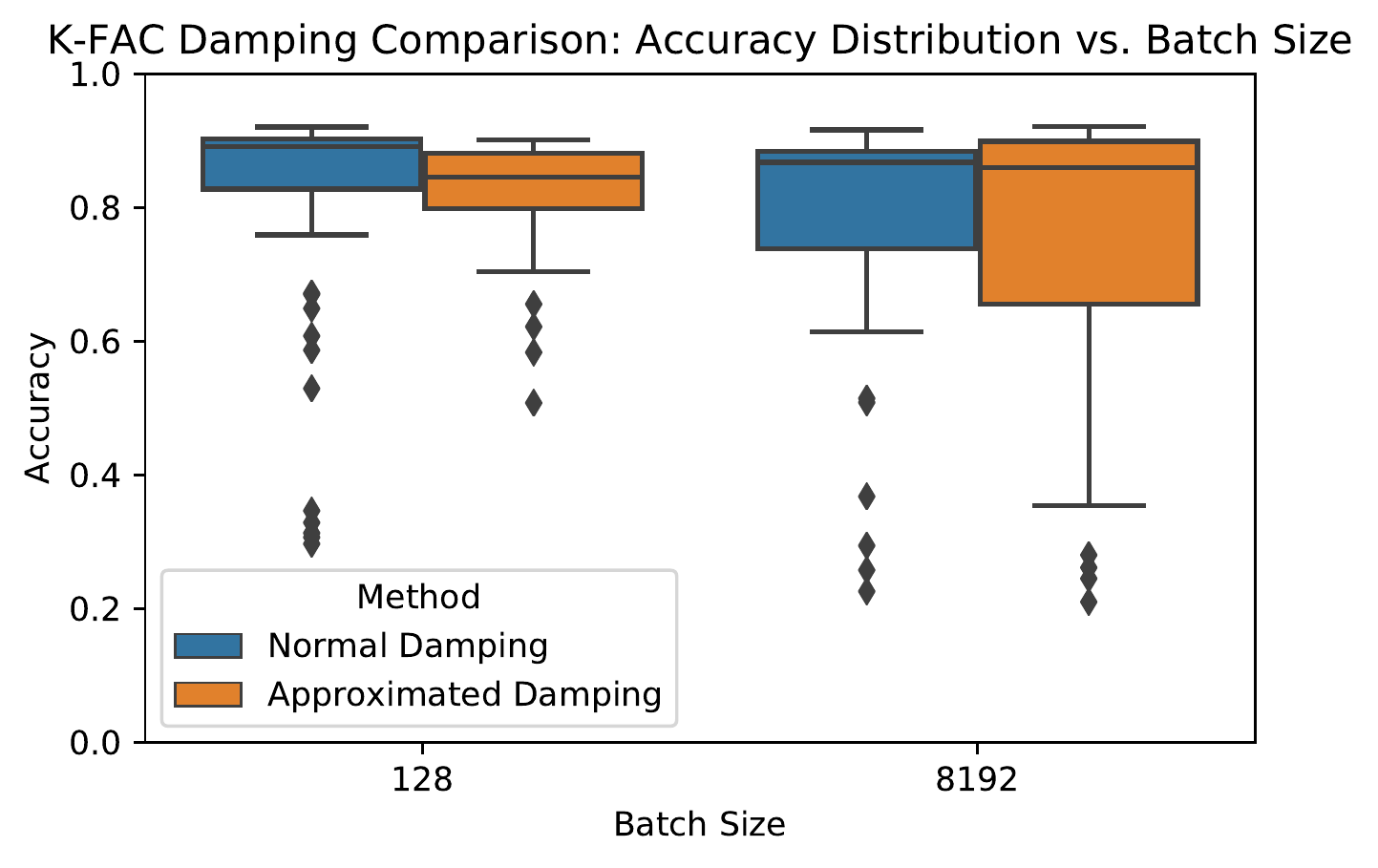}}
\subfloat[]{\label{subfig_damg_compare2}\includegraphics[width=.46\textwidth]{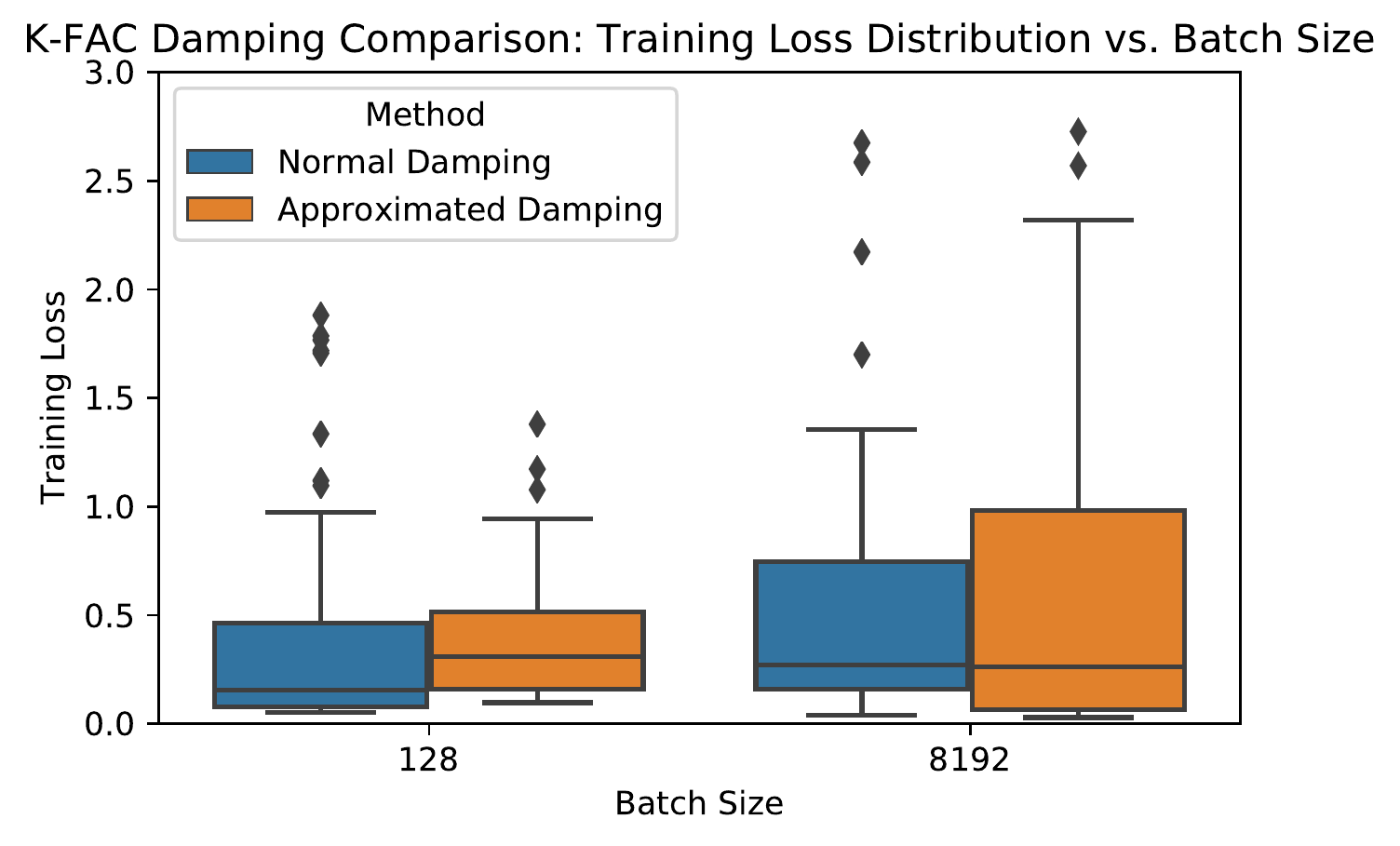}}
\end{center}
\caption{(a): Accuracy distribution versus batch size for both normal damping and approximated damping. (b): Training loss distribution versus batch size for both normal damping and approximated damping. All distributions are formed over the 64 runs of the hyperparameter grid search over damping and learning rate.
}
\label{fig:damping_compare}
\end{figure*}

We compare normal damping and approximated damping for K-FAC with ResNet20 on CIFAR-10, examining their effects for batch sizes 128 and 8,192.  In~\fref{fig:damping_compare} we compare the accuracy and training loss distributions of normal damping and approximated damping for batch sizes 128 and 8,192. For both damping methods, training appears to be more sensitive to hyperparameter tuning under large batch (batch size 8,192). More interestingly, we observe that the distributions generated with normal damping tend to be more robust. We can see, for instance, that for batch size 8,192, approximated damping has a high concentration of hyperparameter configurations yielding favorable values of loss and accuracy.

\subsection{Training Loss/Test Accuracy Trajectories}
\label{subsec:appendix-train-test-curve}

\begin{figure}[h]
\begin{center}
  \subfloat[]{\includegraphics[width=.44\textwidth]{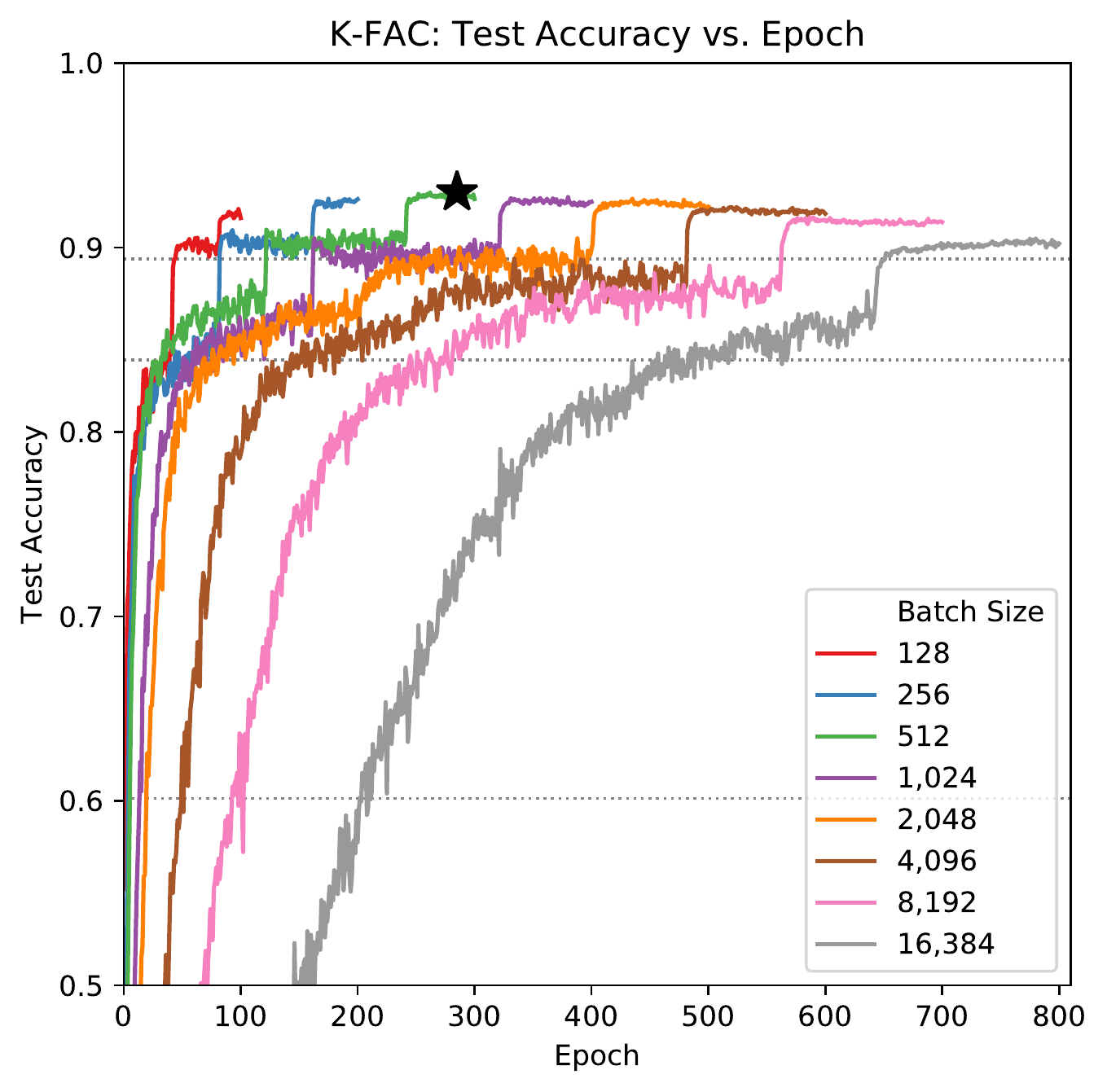}}
  \subfloat[]{\includegraphics[width=.44\textwidth]{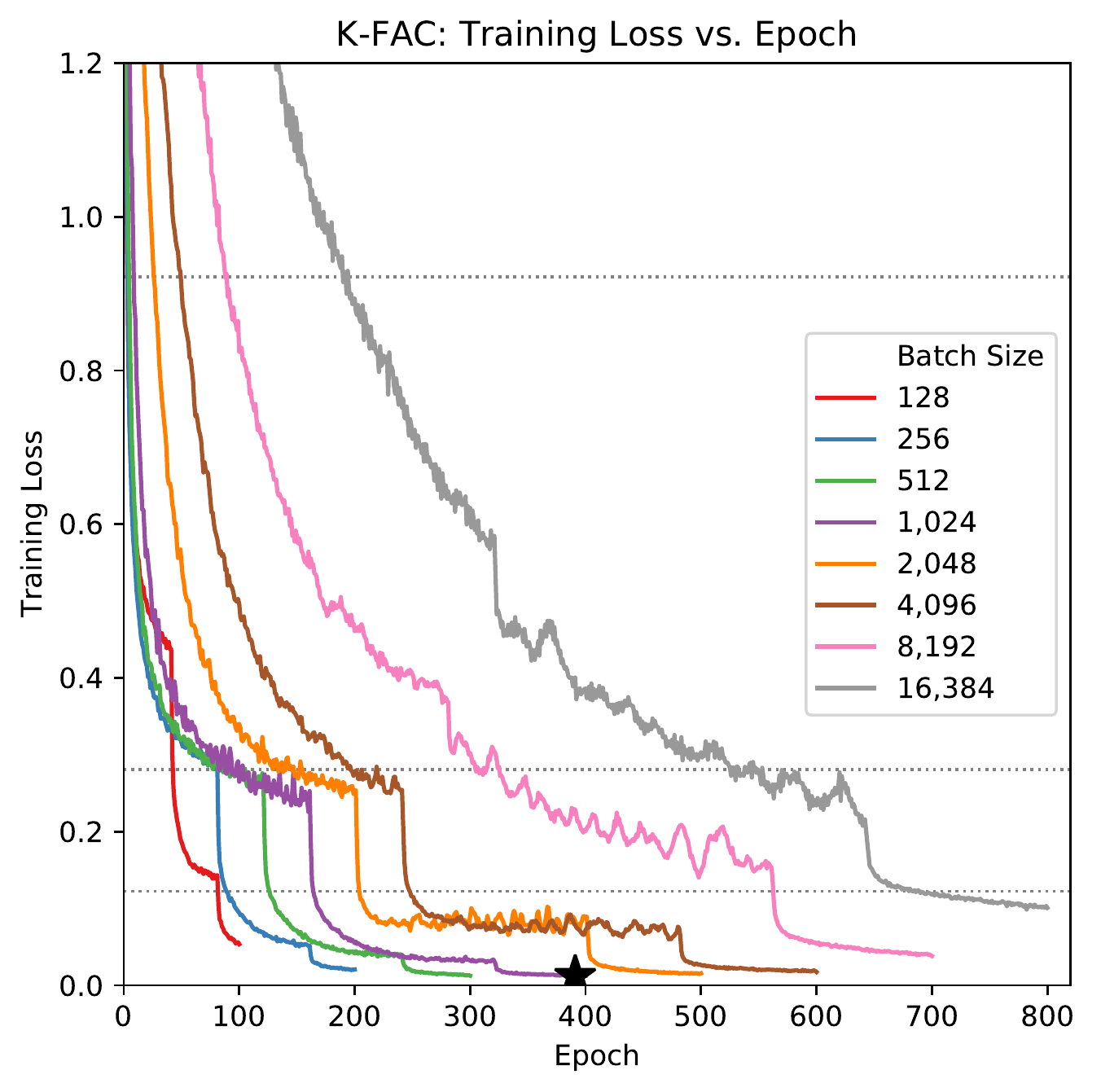}}
  
  \subfloat[]{\includegraphics[width=.44\textwidth]{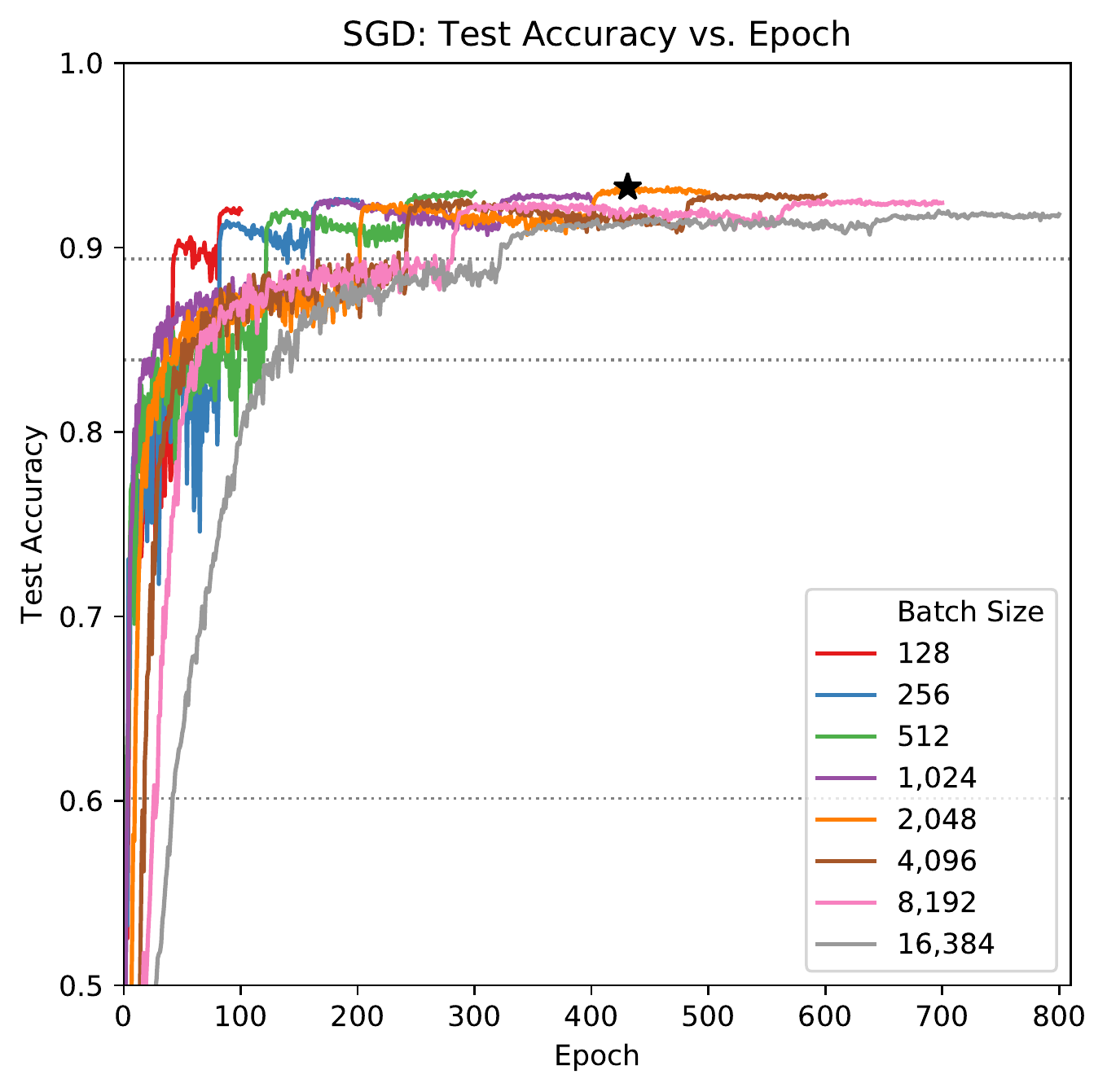}}
  \subfloat[]{\includegraphics[width=.44\textwidth]{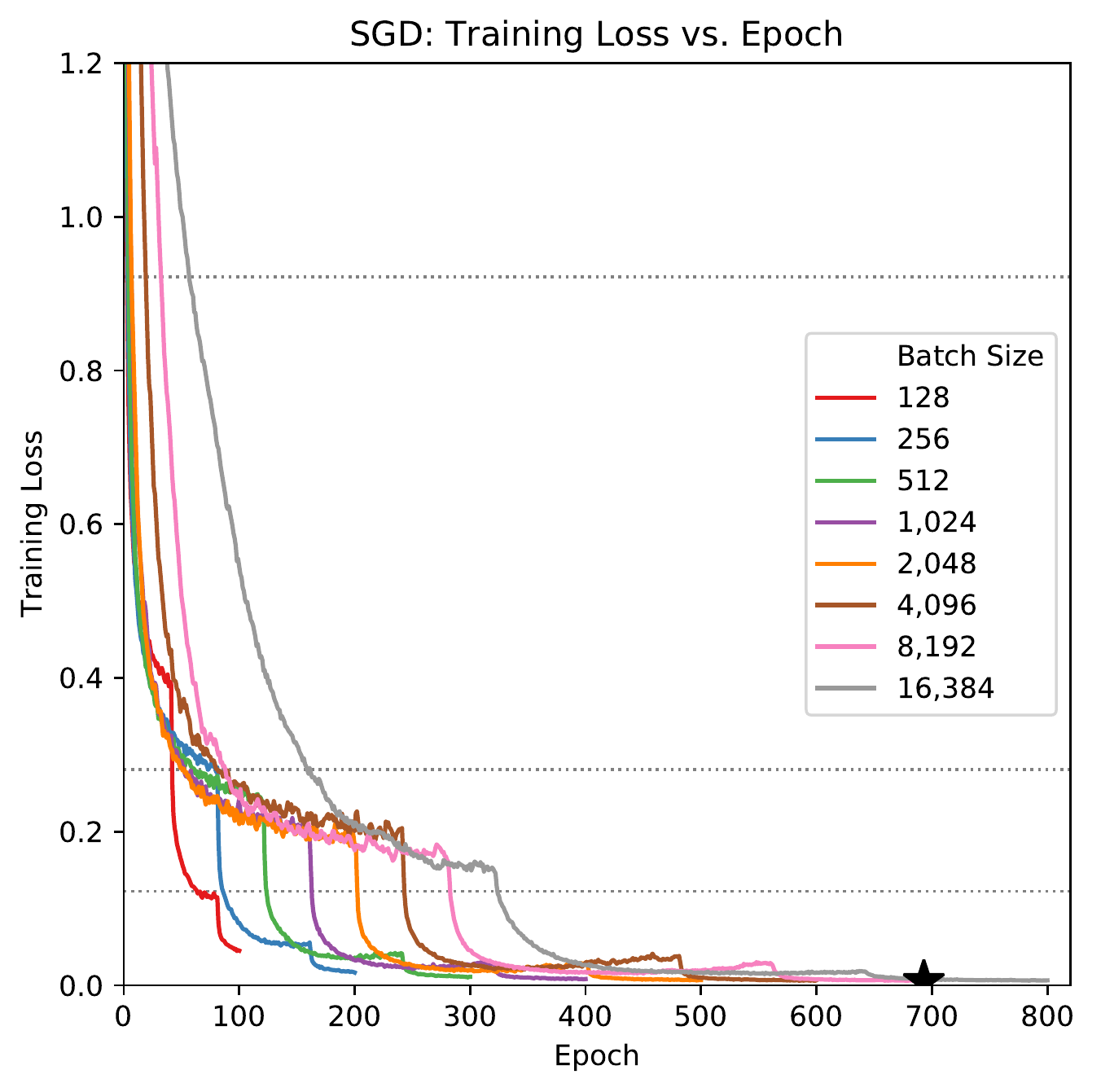}}
\end{center}
\caption{(a)(c): Test accuracy versus epoch for the accuracy-maximizing runs of each batch-size, with K-FAC above and SGD below.
(b)(d): Training loss versus epoch for the loss-minimizing runs of each batch-size, with K-FAC above and SGD below.
For each plot, the star denotes the best (maximal accuracy or minimal loss) performance achieved. Horizontal dotted lines show the target values for accuracy and loss used in~\fref{fig:combined_batch_loss_iters} and~\fref{fig:combined_batch_loss_reduce}.
}
\label{fig:epoch_acc}
\end{figure}
\begin{figure}[h]
\begin{center}
  \subfloat[]{\includegraphics[width=.44\textwidth]{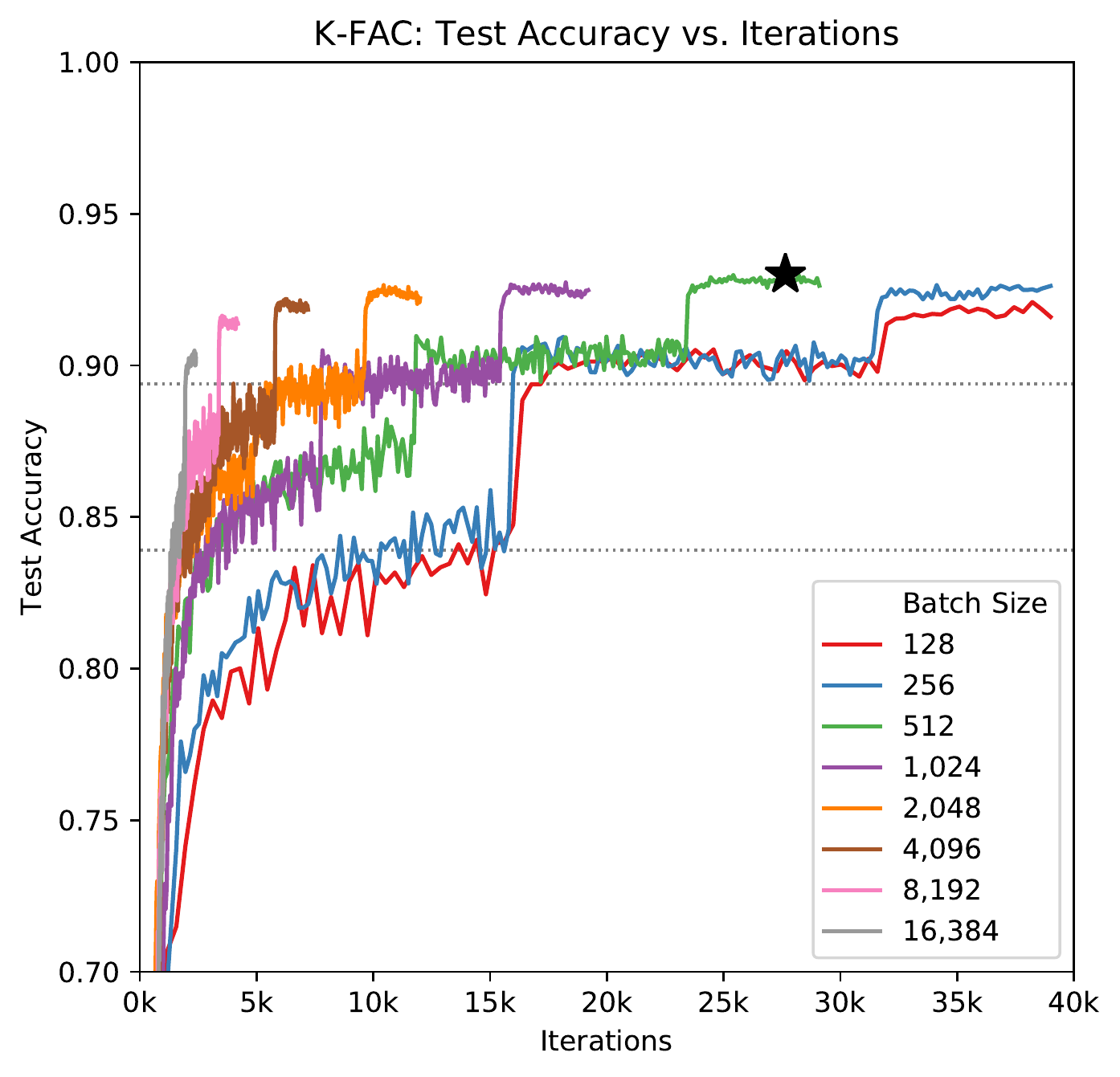}}
  \subfloat[]{\includegraphics[width=.44\textwidth]{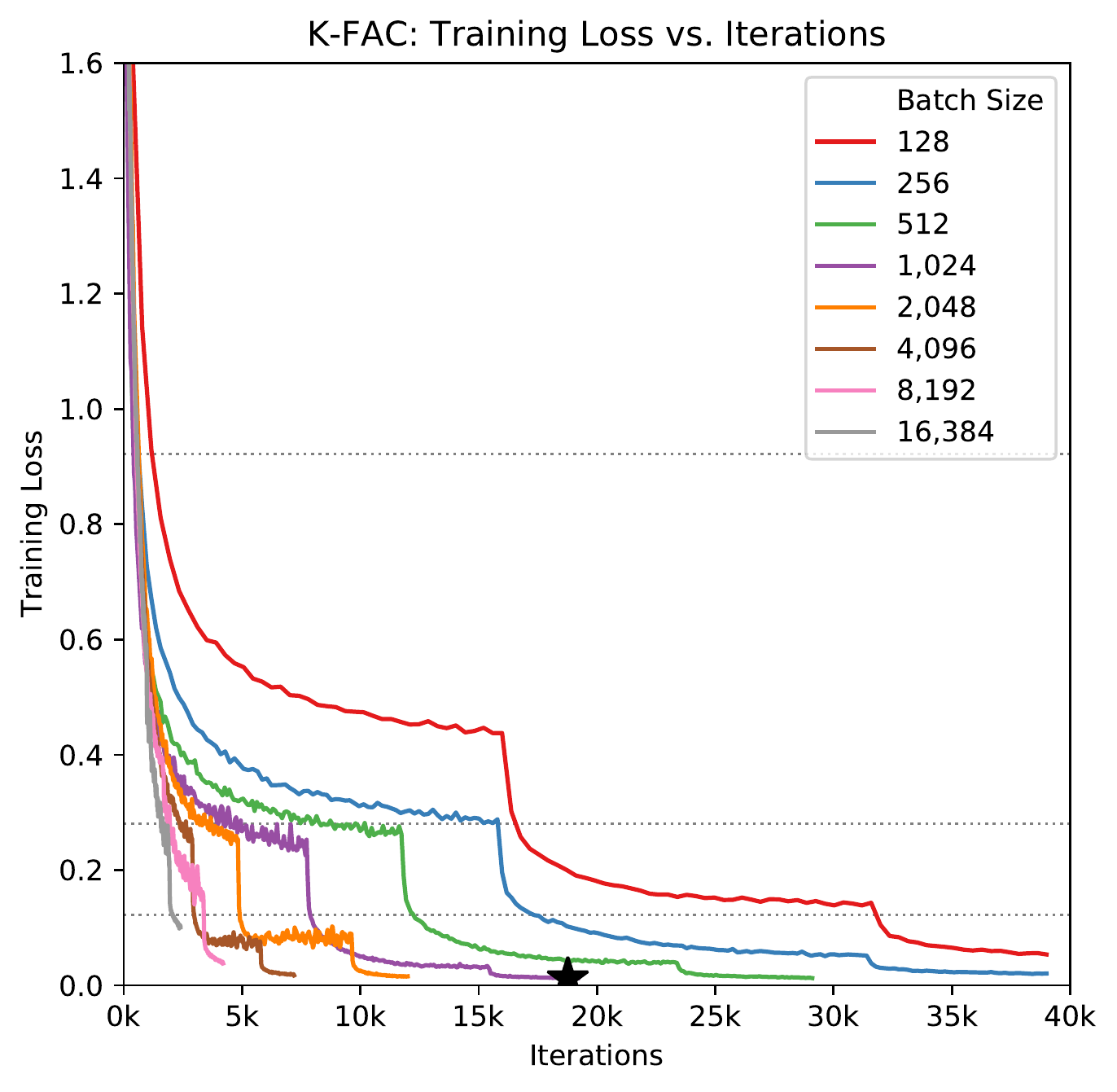}}
  
  \subfloat[]{\includegraphics[width=.44\textwidth]{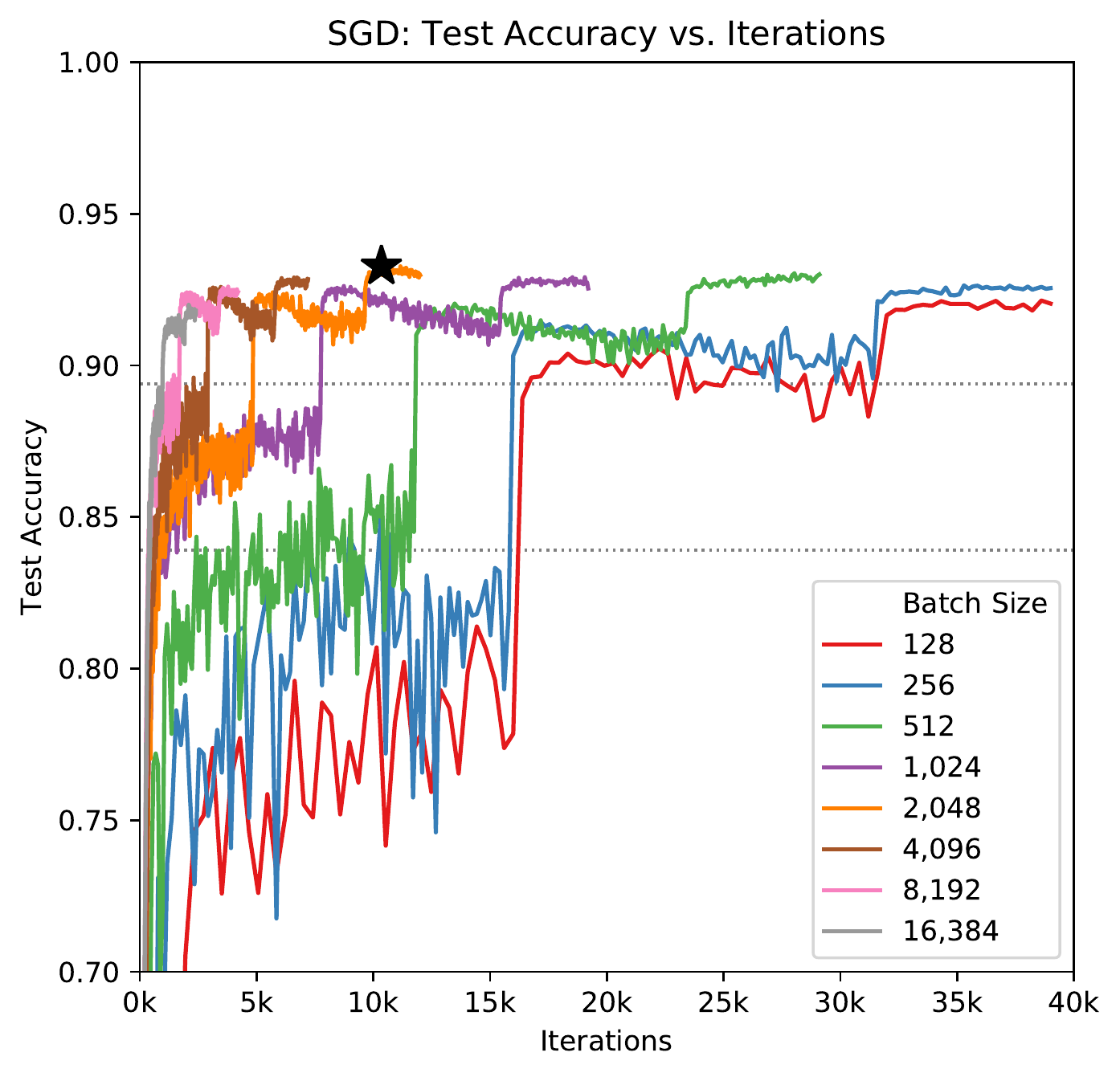}}
  \subfloat[]{\includegraphics[width=.44\textwidth]{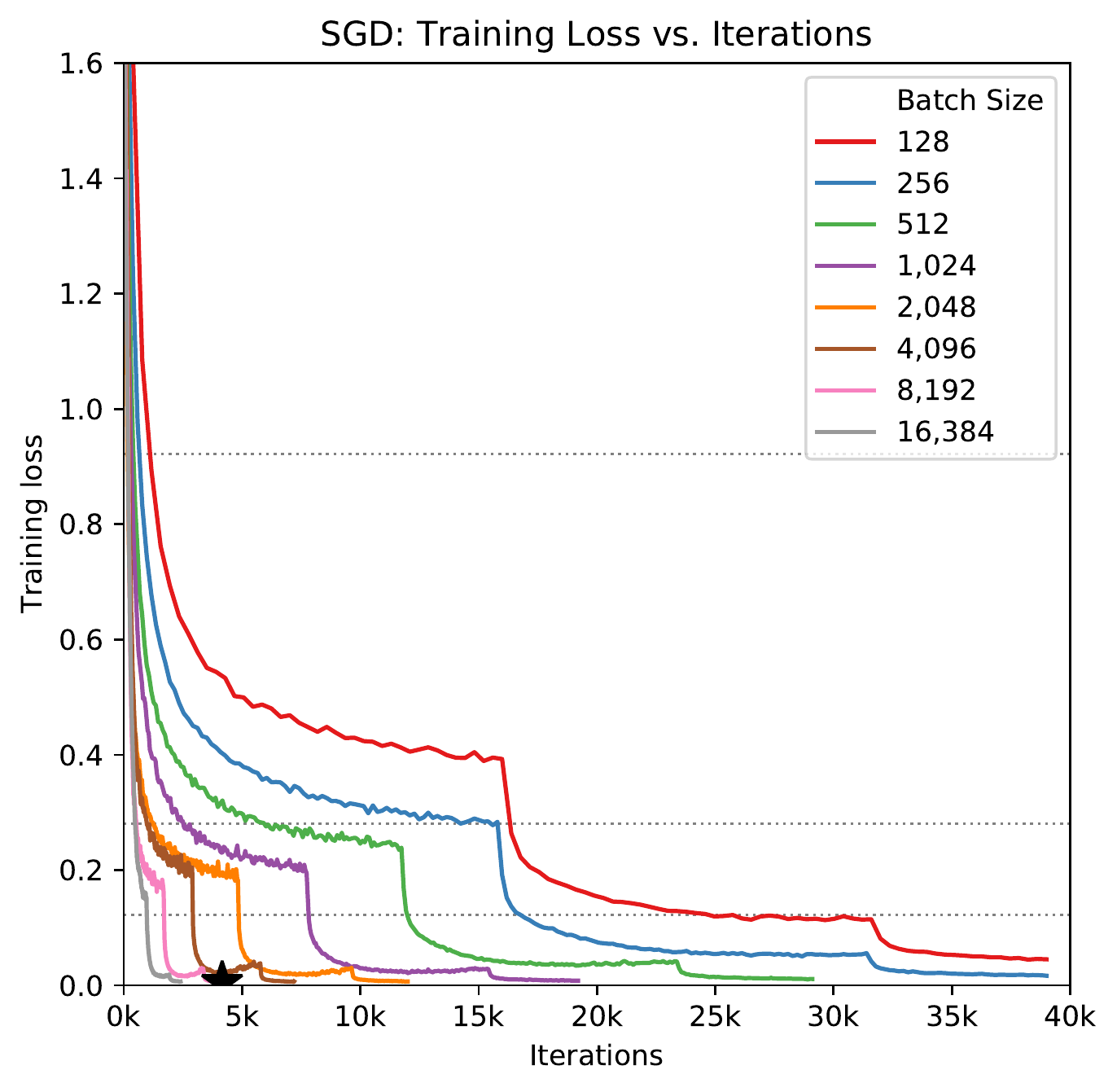}}
\end{center}
\caption{(a)(c):  Test accuracy versus iteration for the accuracy-maximizing runs of each batch-size, with K-FAC above and SGD below.
(b)(d): Training loss versus iteration for the accuracy-maximizing runs of each batch-size, with K-FAC above and SGD below. For each plot, the star denotes the best performance achieved. Horizontal dotted lines show the target values for accuracy and loss used in~\fref{fig:combined_batch_loss_iters} and~\fref{fig:combined_batch_loss_reduce}.
}
\label{fig:iter_acc}
\end{figure}

Here,
we show the detailed training curves for K-FAC and SGD on CIFAR-10 with ResNet20.
We plot the training loss and test accuracy over time for the optimal runs of each batch size for both SGD and K-FAC to support Fig.~\ref{fig:combined_batch_v_acc}
and \ref{fig:combined_batch_loss_reduce} in the main text. Time is measured in terms of both epochs (\fref{fig:epoch_acc}) and iterations (\fref{fig:iter_acc}). 

The best performance of each curve in Fig. \ref{fig:epoch_acc} helps to explain the shape of~\fref{fig:combined_batch_v_acc}, in which we plot the best test accuracy and training loss for each batch size for SGD and K-FAC. The best achieved accuracy can be seen to drop for both K-FAC and SGD beyond a certain batch size. The lowest loss is achieved at higher batch sizes for SGD than for K-FAC. 
\fref{fig:iter_acc} gives clues regarding the large-batch training speedup of both K-FAC and SGD. We can see from the figure that as batch size grows, fewer iterations are necessary to reach specific performance targets, represented by horizontal dotted lines. In the ideal scaling case, the intersections of training loss curves for each batch size would fall along the dotted target lines in the pattern of a geometric sequence with common ratio $1/2$. 

\subsection{Large Batch Scalabilities: Iterations to Target}
\label{subsec:appendix-iterations-to-target}

Here, we present the iterations-batch size relations to support our results in Section~\ref{sec:comparison_scalability}.
\fref{fig:iter_acc} gives clues regarding the large-batch training speedup of both K-FAC and SGD. We can see from the figure that as batch size grows, fewer iterations are necessary to reach specific performance targets, represented by horizontal dotted lines. In the ideal scaling case, the intersections of training loss curves for each batch size would fall along the dotted target lines in the pattern of a geometric sequence with common ratio $1/2$. 

\begin{figure}[h]
\begin{center}
  \includegraphics[width=.4\textwidth]{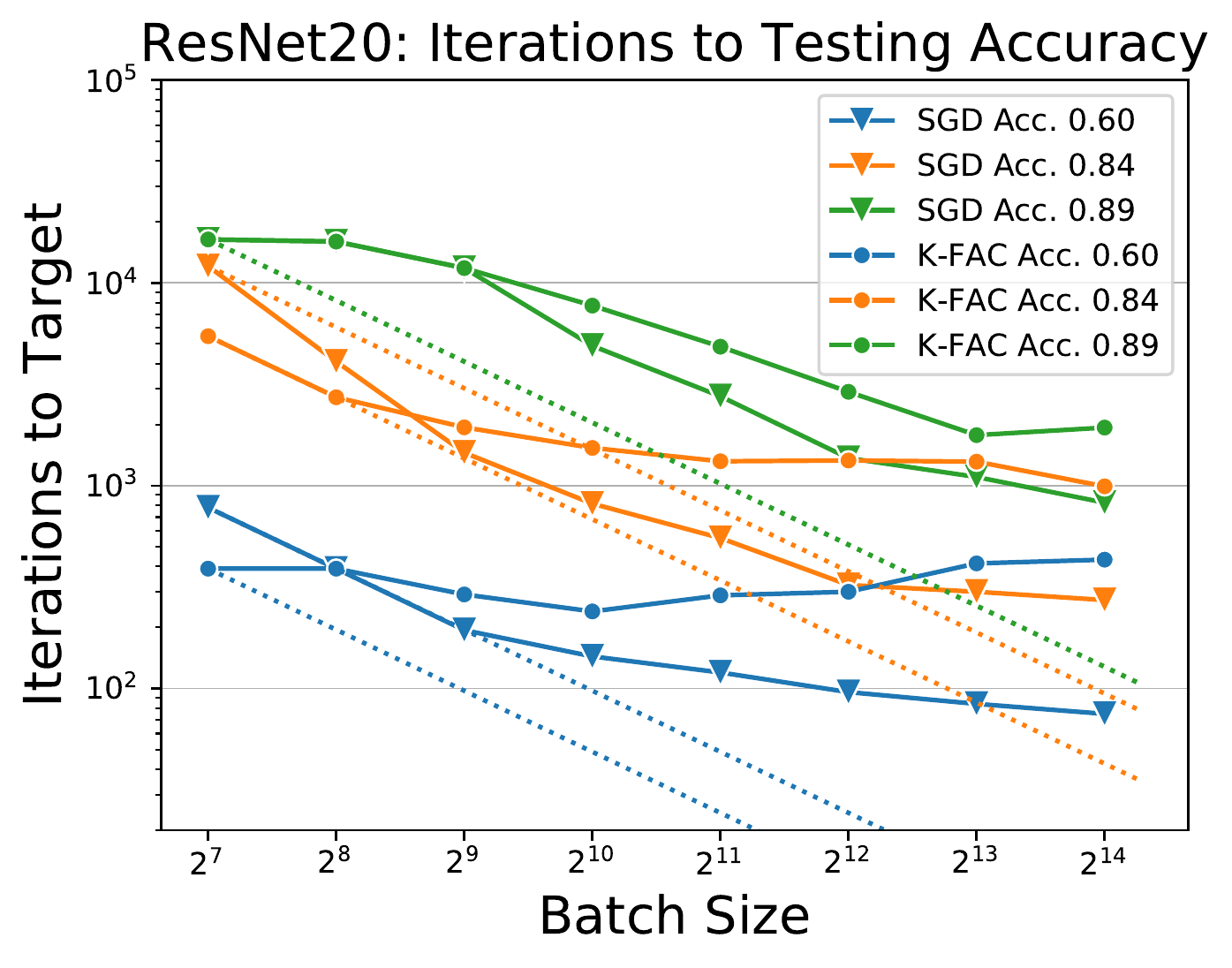}
  \includegraphics[width=.4\textwidth]{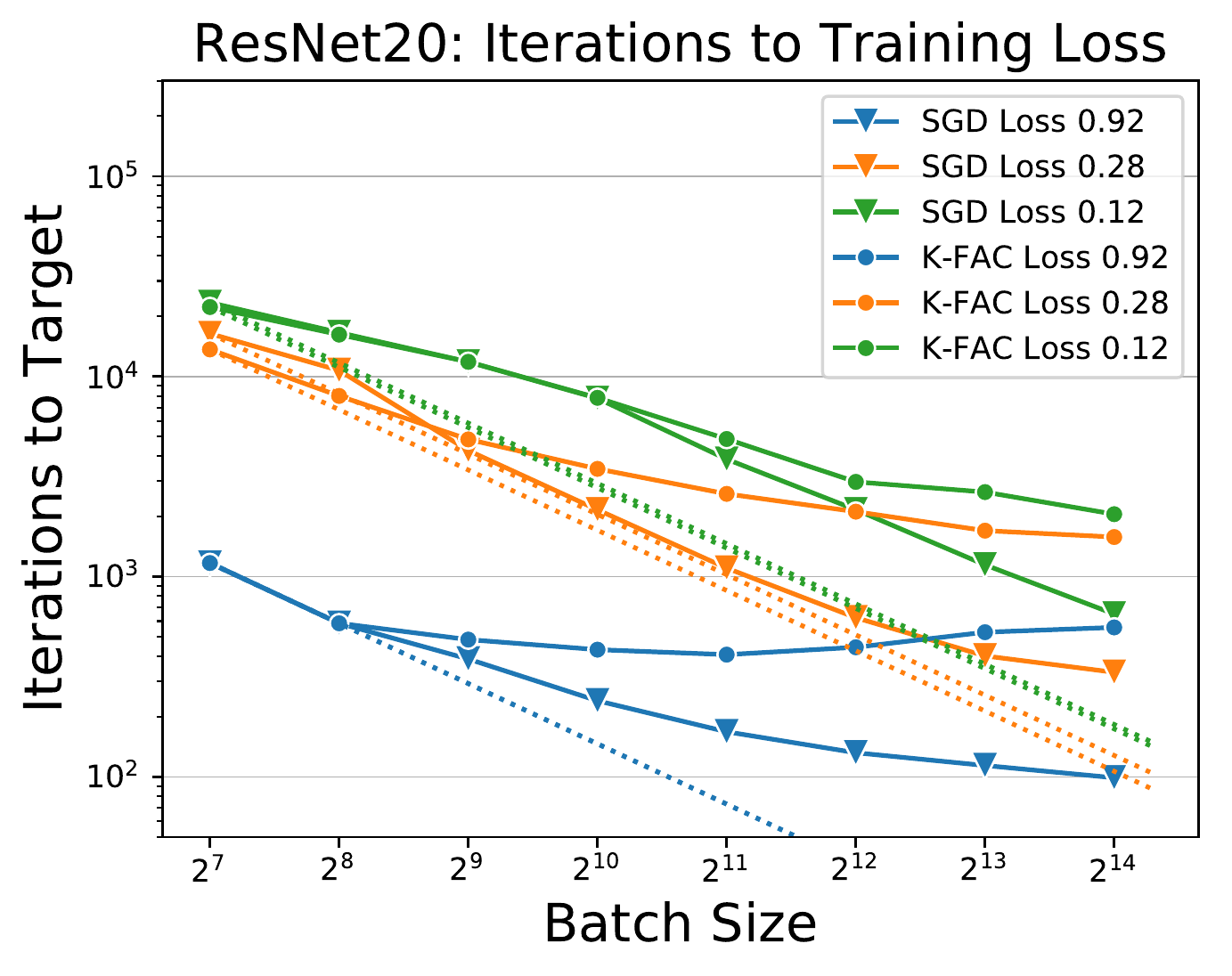}

  \includegraphics[width=.4\textwidth]{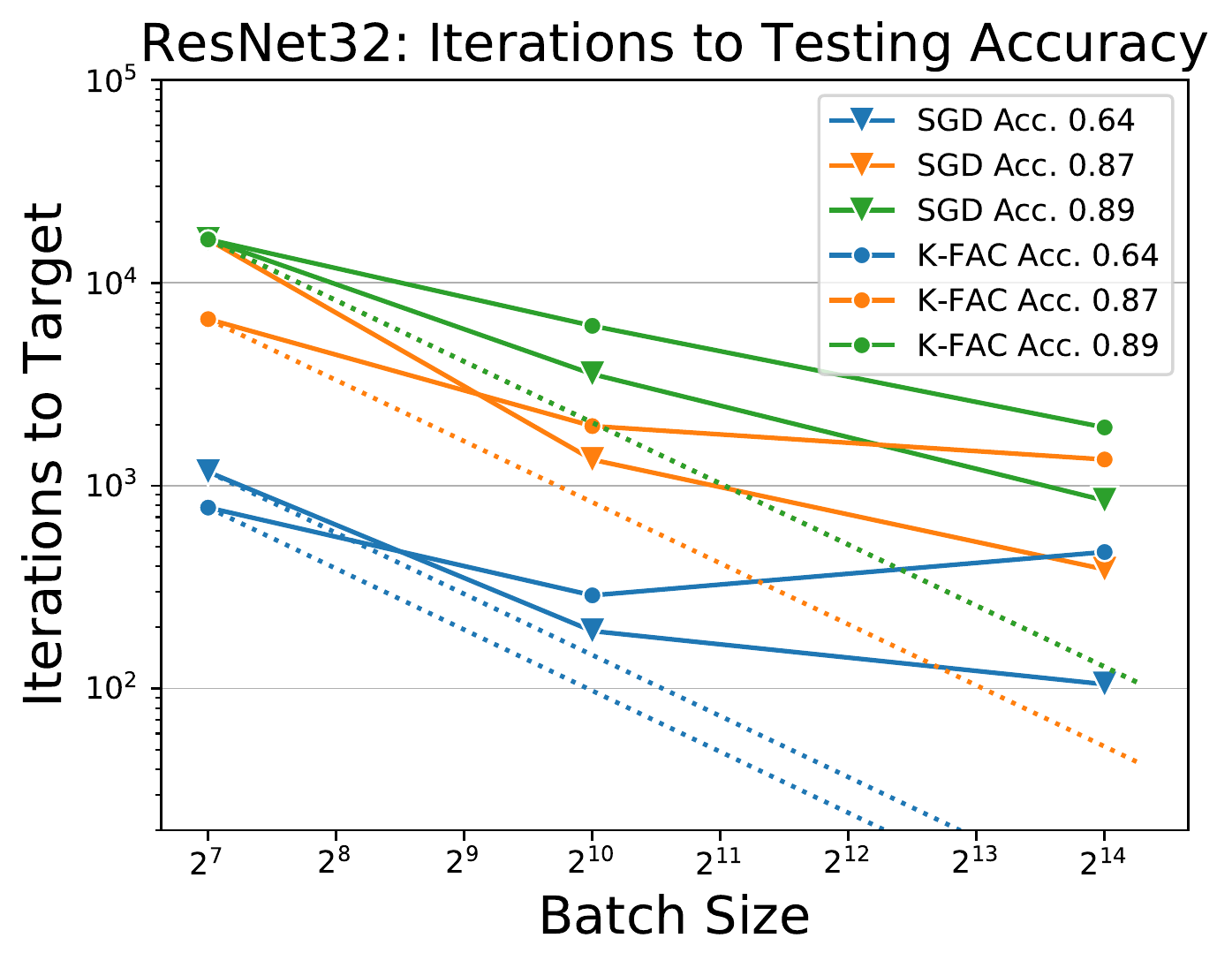}
  \includegraphics[width=.4\textwidth]{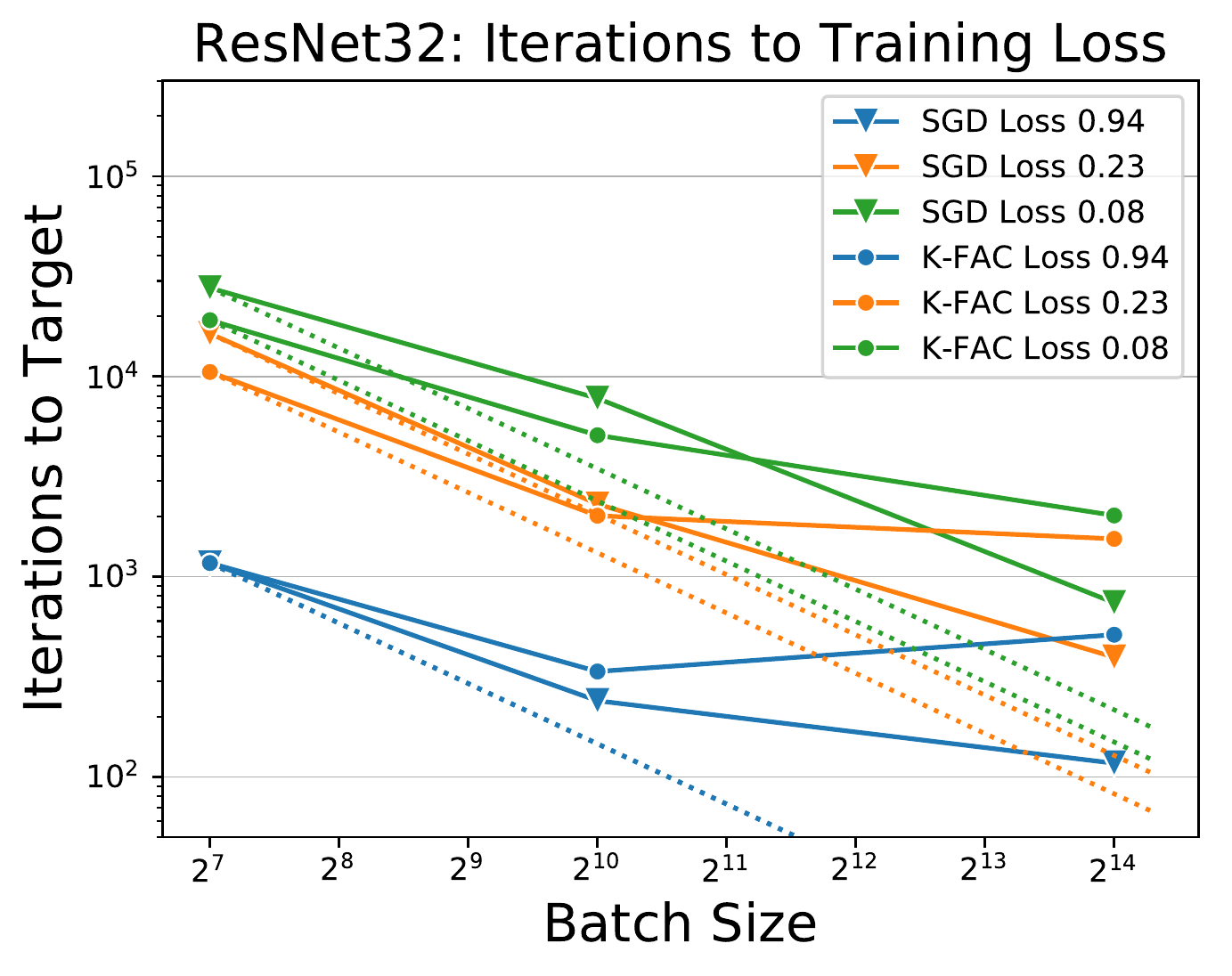}
  
  \includegraphics[width=.4\textwidth]{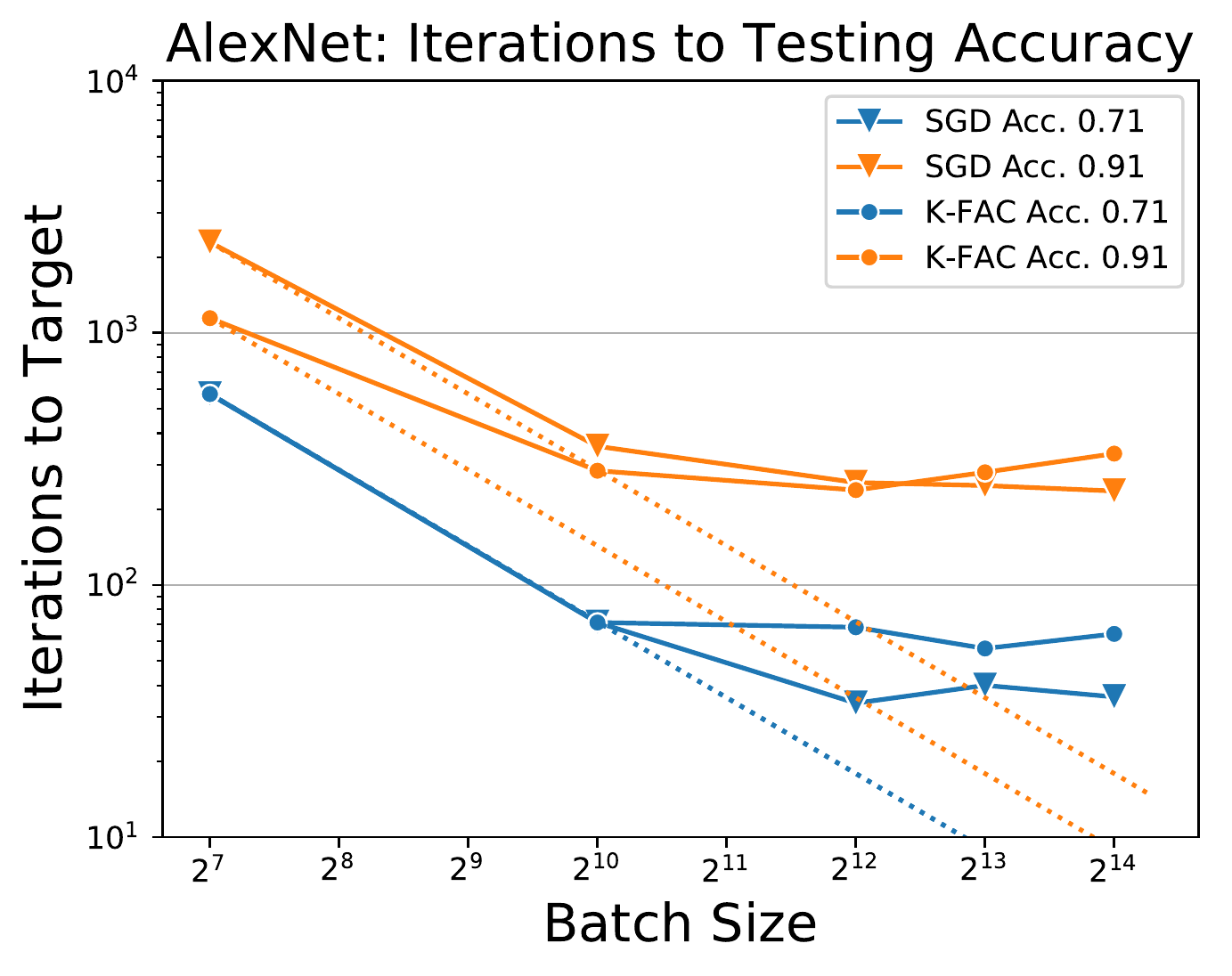}
  \includegraphics[width=.4\textwidth]{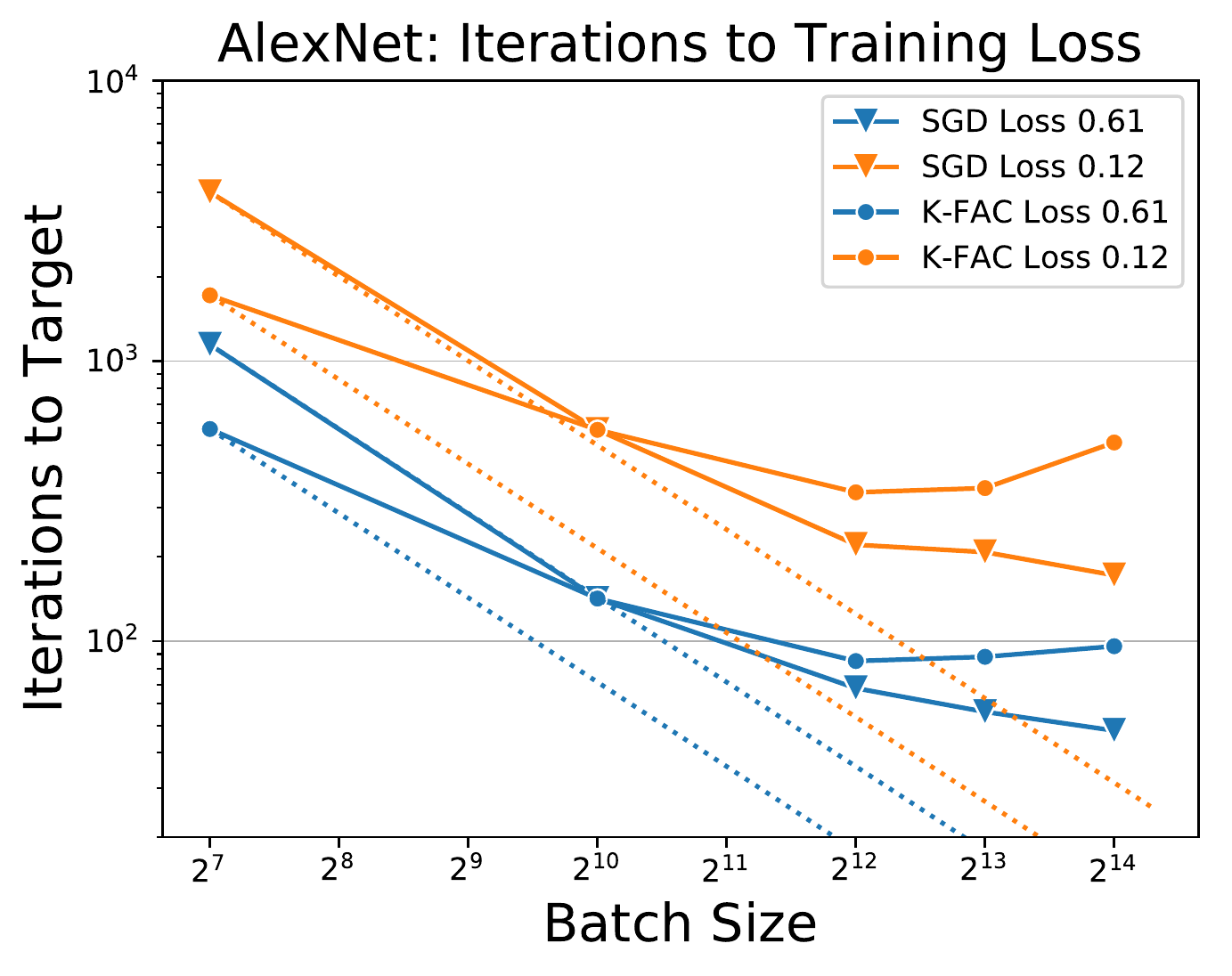}

\end{center}
\caption{
From top to bottom: Iterations to a target training loss / test accuracy versus batch size for both SGD and K-FAC with ResNet20 on CIFAR-10, ResNet32 on CIFAR-10 and AlexNet on SVHN, respectively.
The ideal scaling result that relates batch-size to convergence iterations is plotted for each method / target as a dotted line. Note that K-FAC has no better scalability than~SGD.
}
\label{fig:combined_batch_loss_iters}
\end{figure}

Iterations-to-target for each batch size are plotted directly in~\fref{fig:combined_batch_loss_iters}.  Dotted lines denote the ideal scaling relationship between batch size and iterations. 
The large-batch scalability behavior is captured in the slope of each line in~\fref{fig:combined_batch_loss_iters}.
For both K-FAC and SGD, diminishing return effects are present. 
In all examined cases, K-FAC deviates from ideal scaling (dotted lines) to a greater extent than SGD as batch size increases. 
This difference explains why, in~\fref{fig:combined_batch_v_acc}, SGD is increasingly able to outperform K-FAC for large batches given a fixed budget.

\subsection{K-FAC Heatmaps}
\label{subsec:appendix-heatmap}

\begin{figure*}[h]
\begin{center}
\includegraphics[width=.32\textwidth]{fig/kfac_batch_128.pdf}
\includegraphics[width=.32\textwidth]{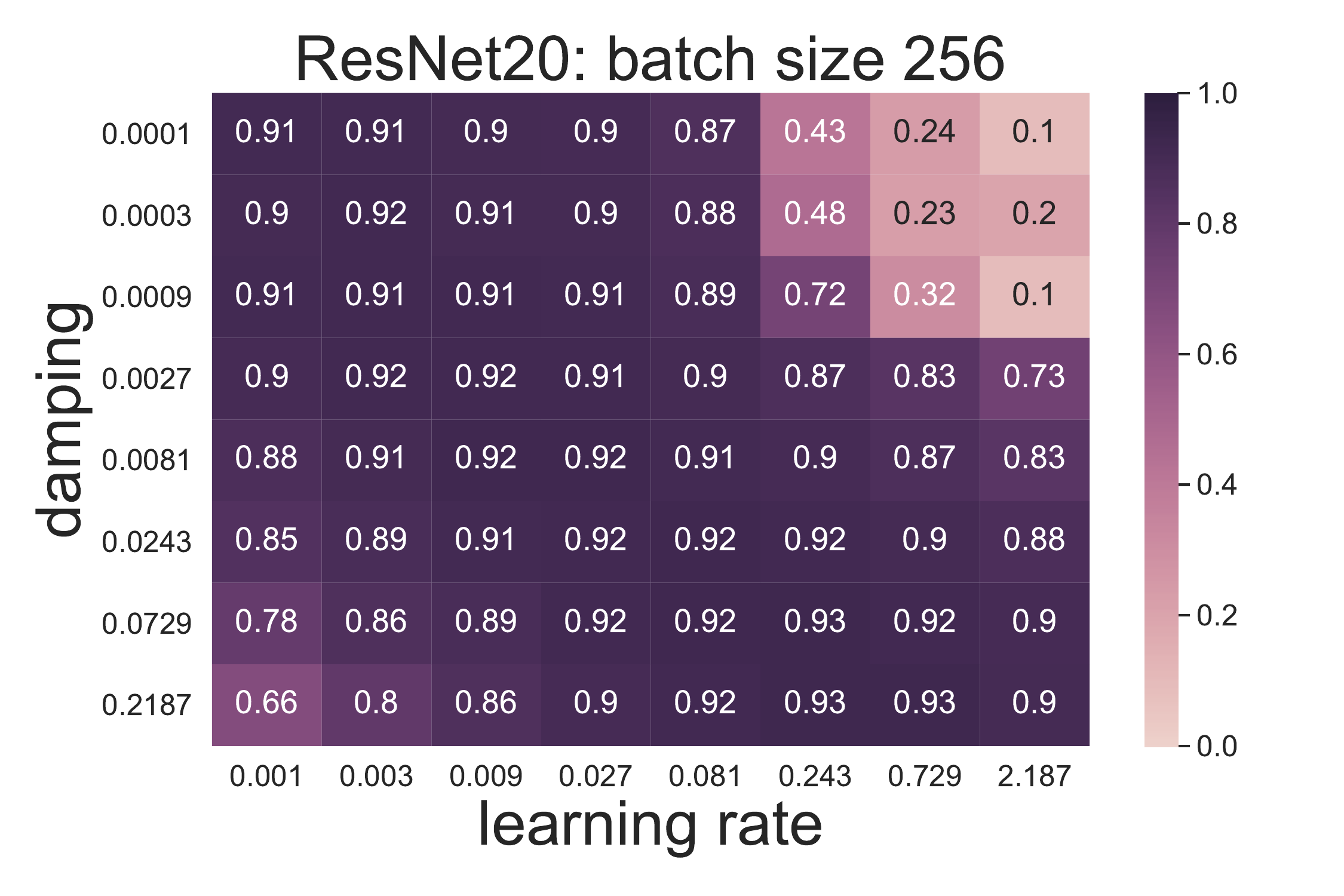}
\includegraphics[width=.32\textwidth]{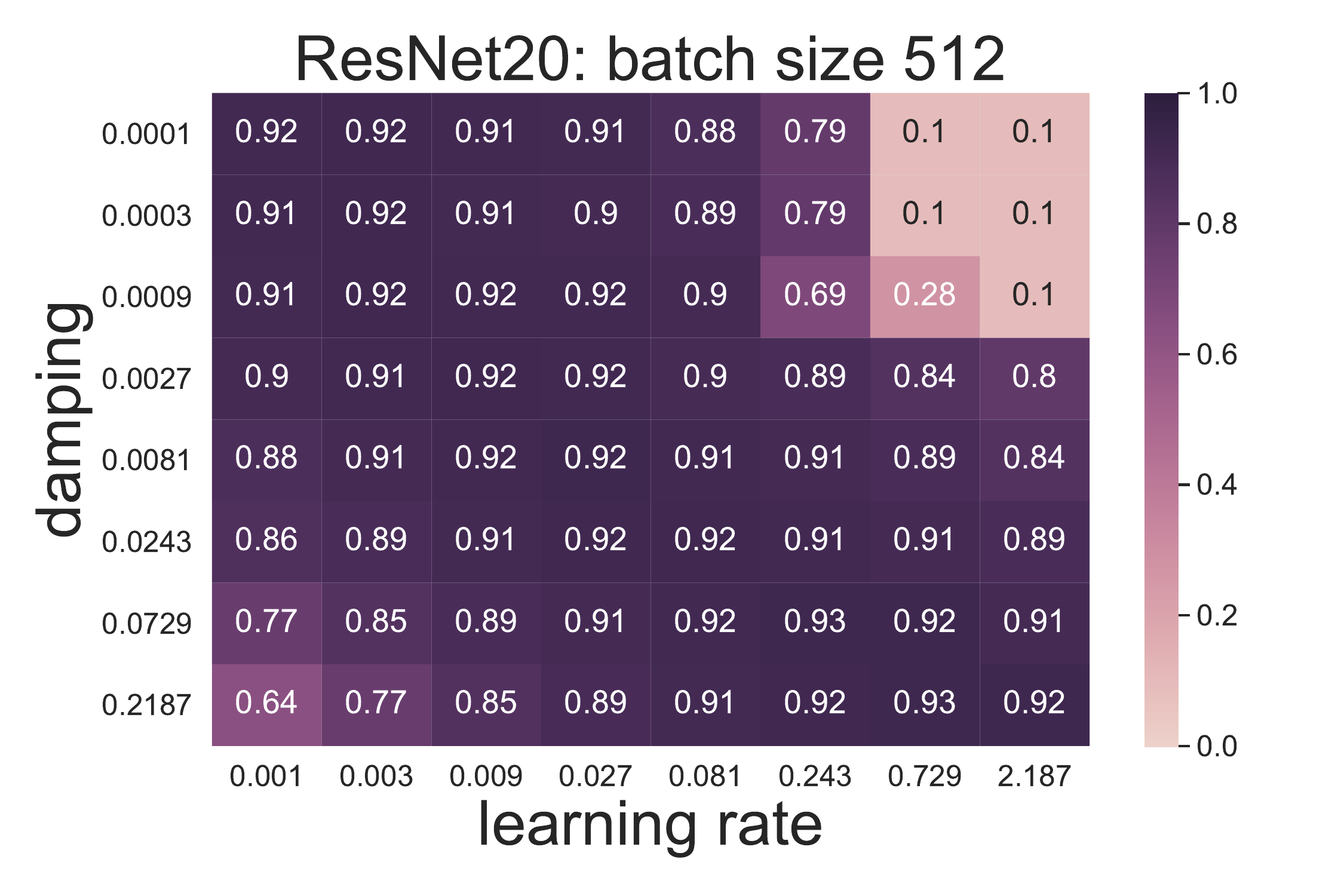}
\includegraphics[width=.32\textwidth]{fig/kfac_batch_1024.pdf}
\includegraphics[width=.32\textwidth]{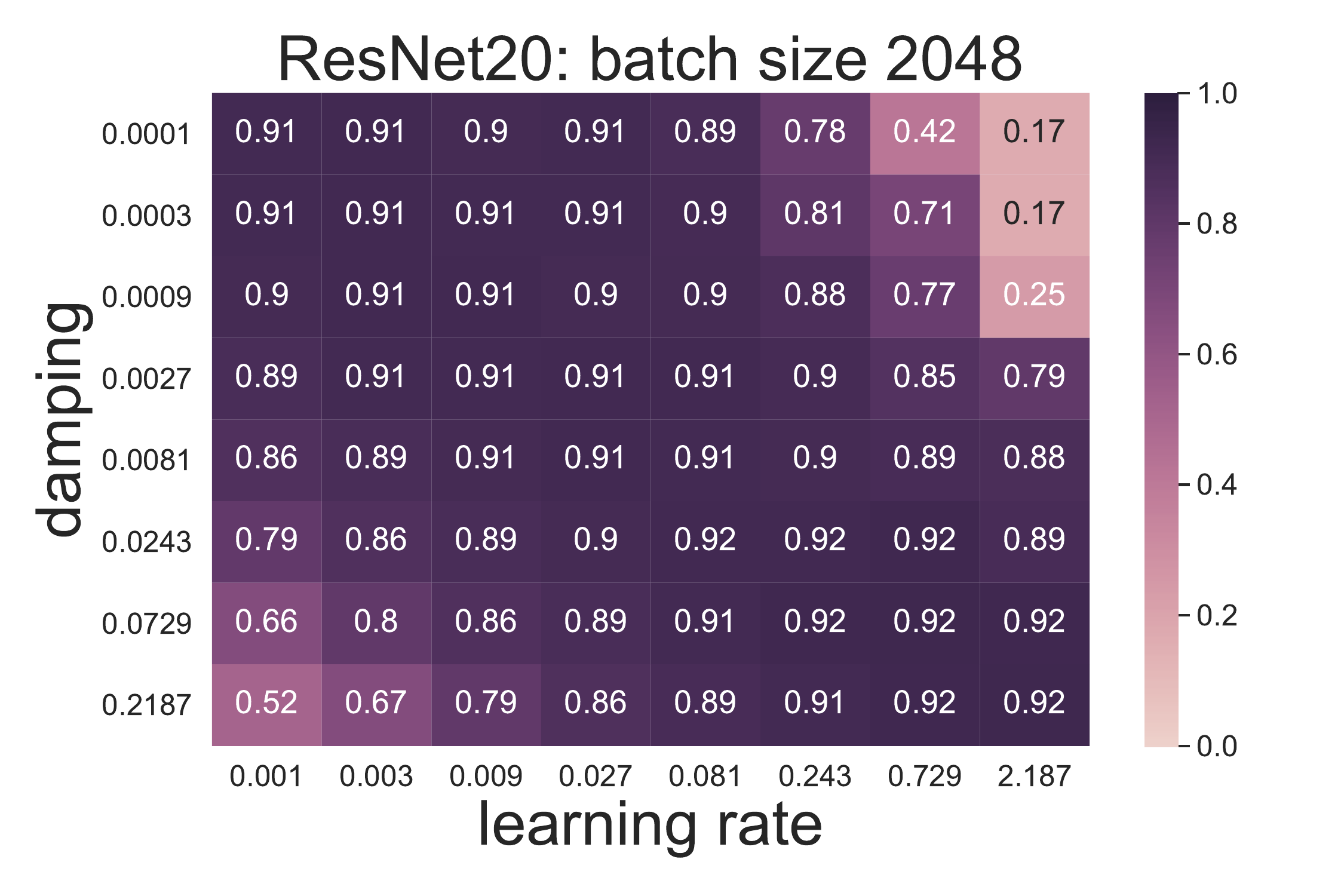}
\includegraphics[width=.32\textwidth]{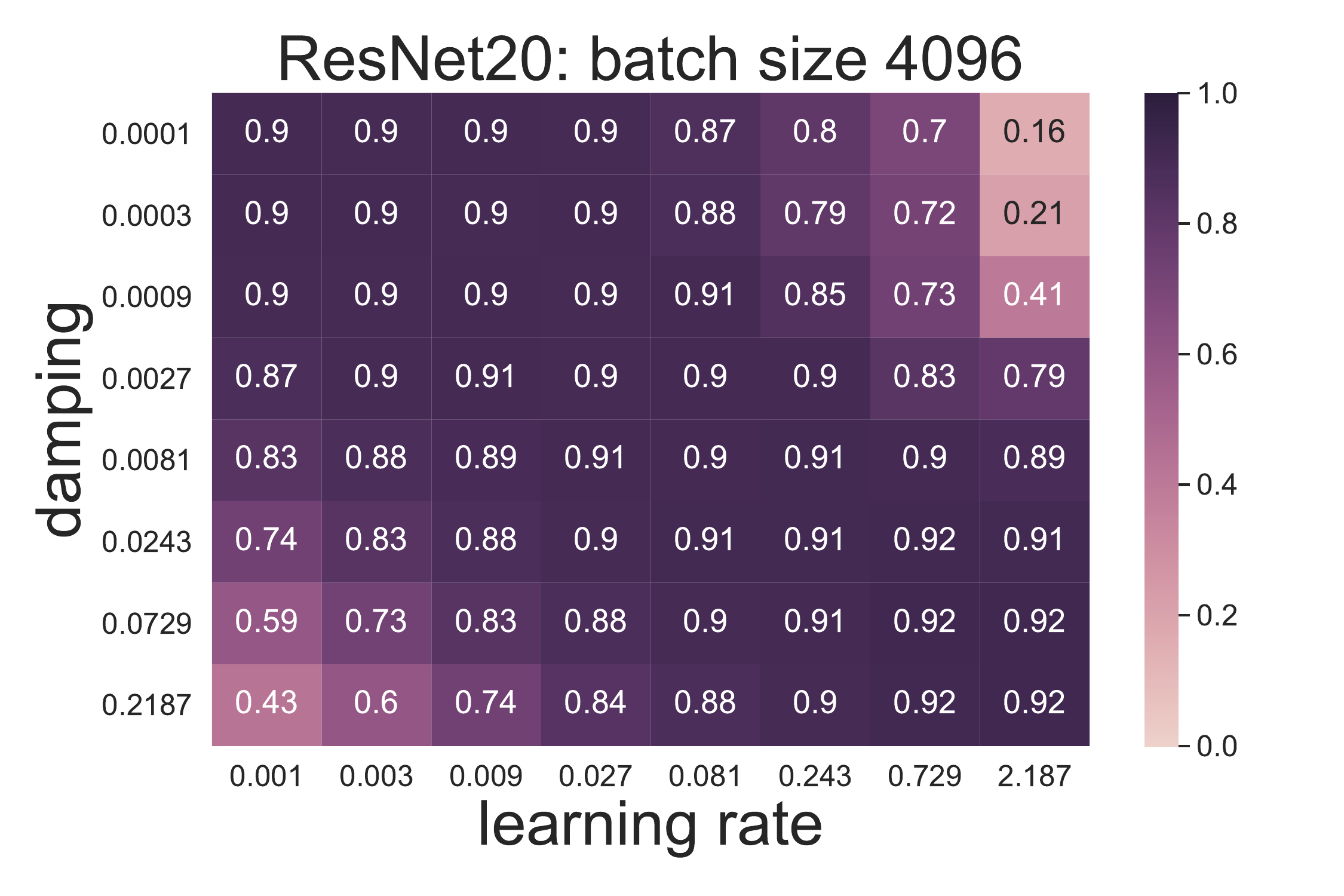}
\includegraphics[width=.32\textwidth]{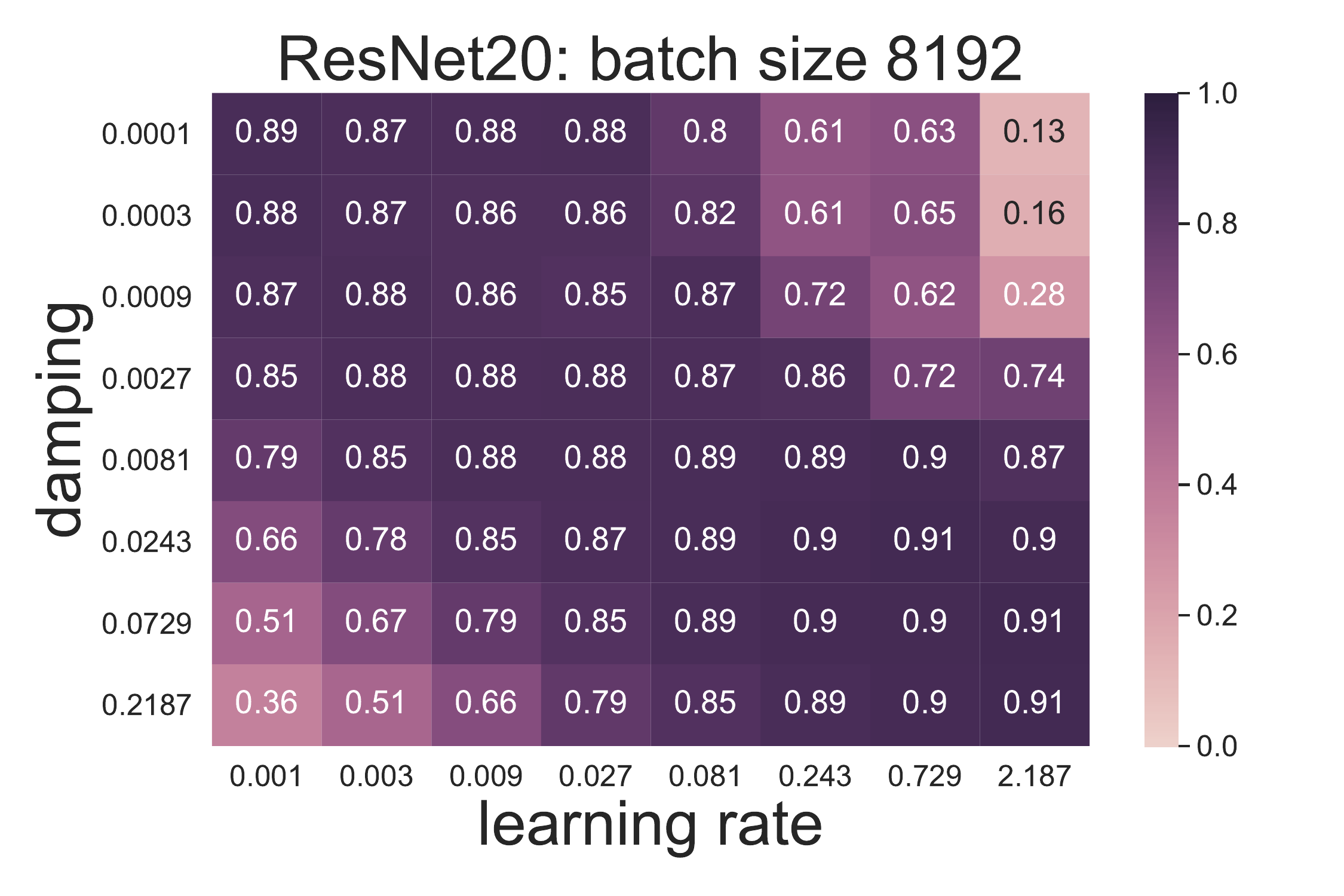}
\includegraphics[width=.32\textwidth]{fig/kfac_batch_16384.pdf}
\end{center}
\caption{Best accuracy achieved under adjusted epoch budget versus damping and learning rate for batch sizes from 128 to 16,384 with normal damping on CIFAR-10 with ResNet20.
A positive correlation between damping and learning rate is exhibited. When the batch size exceeds 4,096, we observe a shrinking of the high-accuracy region for large batch sizes. }
\label{fig:heatmap_complete}
\end{figure*}

\begin{figure}[h]
\begin{center}
  \subfloat[]{\includegraphics[width=.49\textwidth]{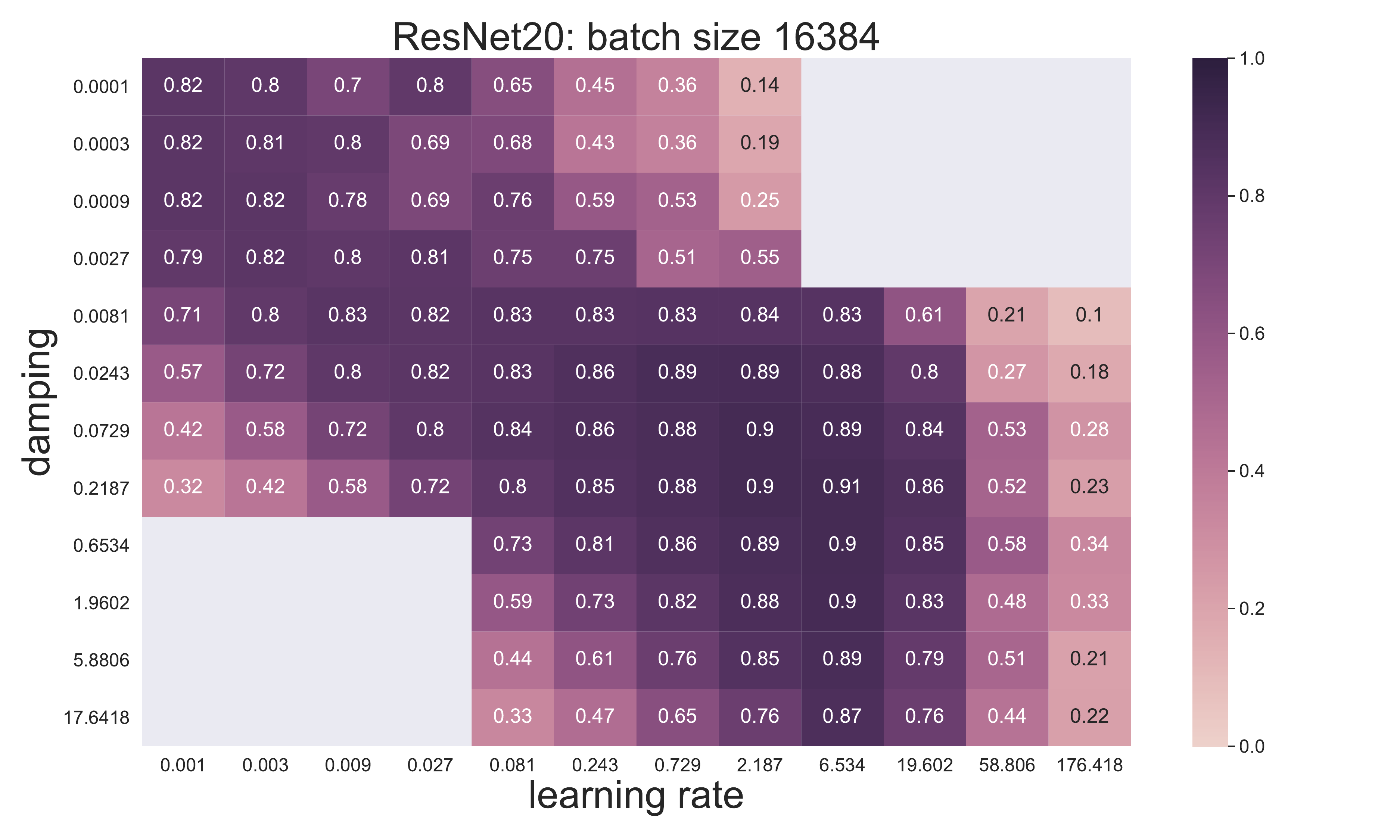}}
  \subfloat[]{\includegraphics[width=.49\textwidth]{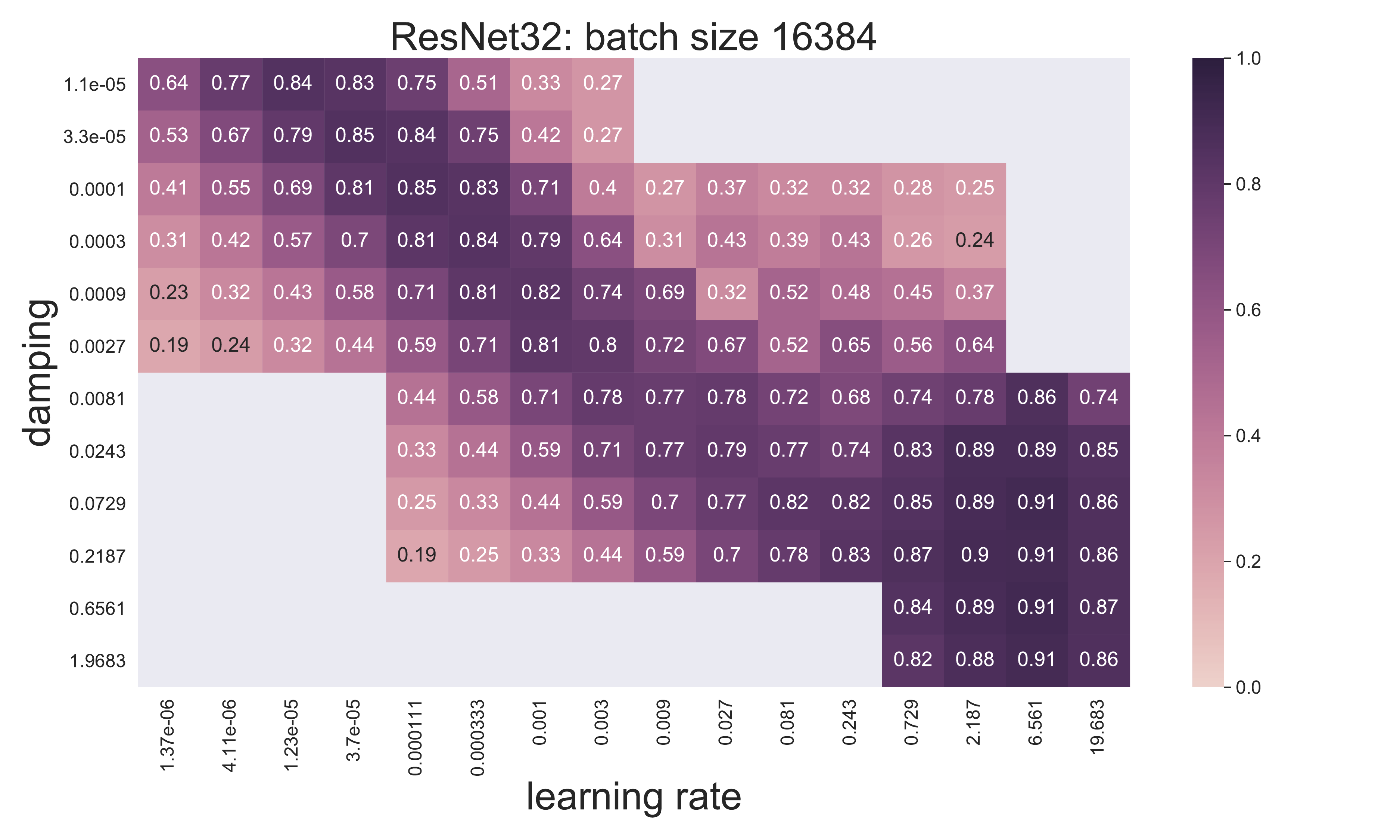}}
\end{center}
\caption{Best accuracy achieved versus damping and learning rate under a larger hyperparameter tuning space for batch size under 16,384 on CIFAR-10 with ResNet20 (left) and ResNet32 (right). For both cases, the areas of the high-accuracy region are still smaller than that of small batch training results.
}
\label{fig:full-heatmap-cifar10-Res20+32}
\end{figure}

Here, we show a complete list of heatmaps.
\fref{fig:heatmap_complete} displays the complete accuracy heatmaps for experiments on CIFAR-10 with ResNet20.
For each hyperparameter configuration, training was run until terminated according to the adjusted epoch budget. We can observe a shrinkage of the high-accuracy region when the batch size exceeds 4,096. In addition, we extend the hyperparameter tuning spaces for both ResNet20 and ResNet32 experiments under batch size 16,384, and the shrinkage of the high-accuracy region still exists, as is shown in~\fref{fig:full-heatmap-cifar10-Res20+32}. 
This strengthens the argument that K-FAC is more sensitive to hyperparameters under large batch sizes.

\subsection{Hyperparameter Sensitivity of K-FAC on Other Models and Datasets}
\label{subsec:appendix-robustness-behavior}

\begin{figure}[!htp]
\begin{center}
  \subfloat[]{\label{subplot-resnet32-cifar10}
  \includegraphics[width=.35\textwidth]{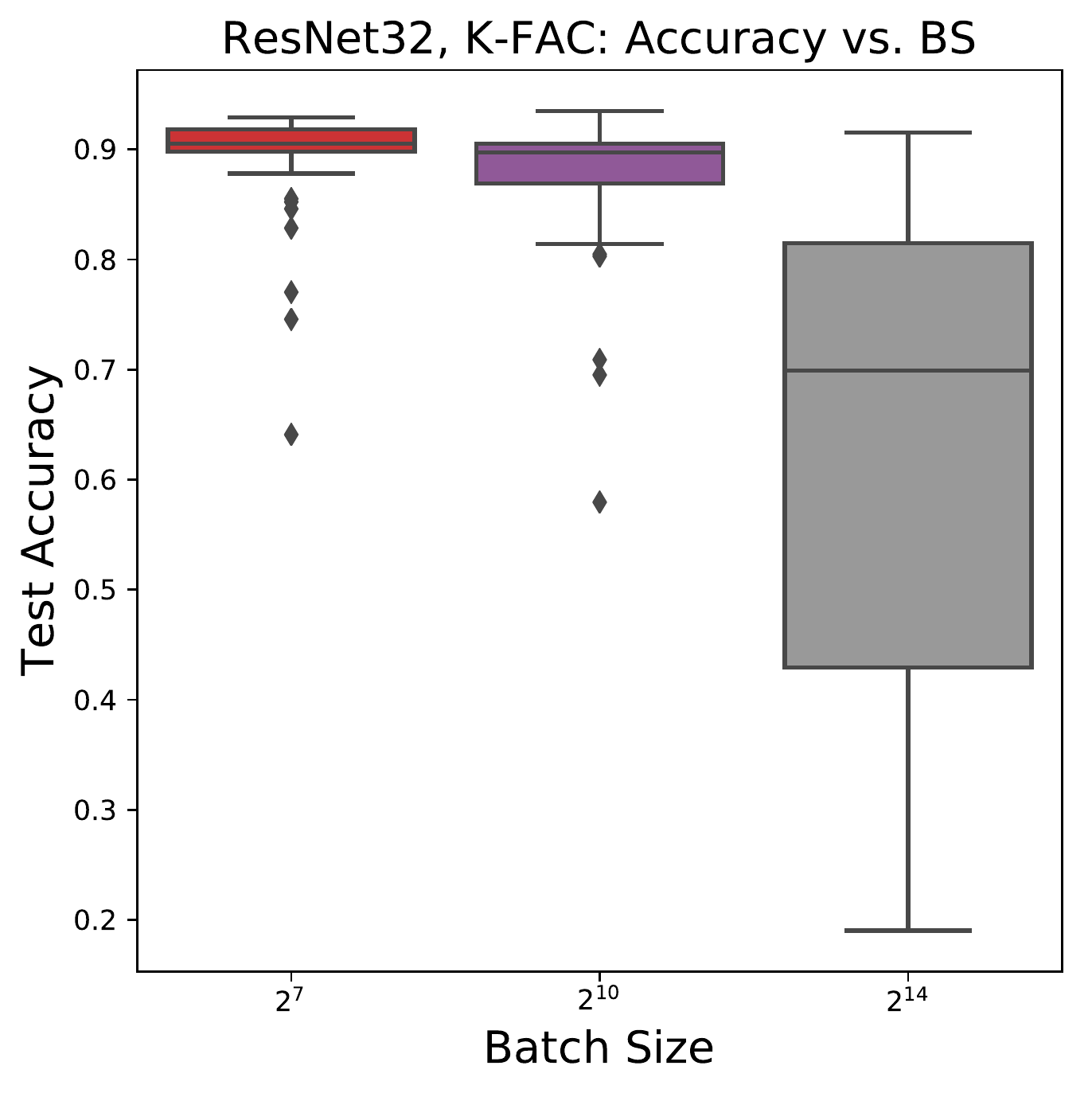}
  \includegraphics[width=.35\textwidth]{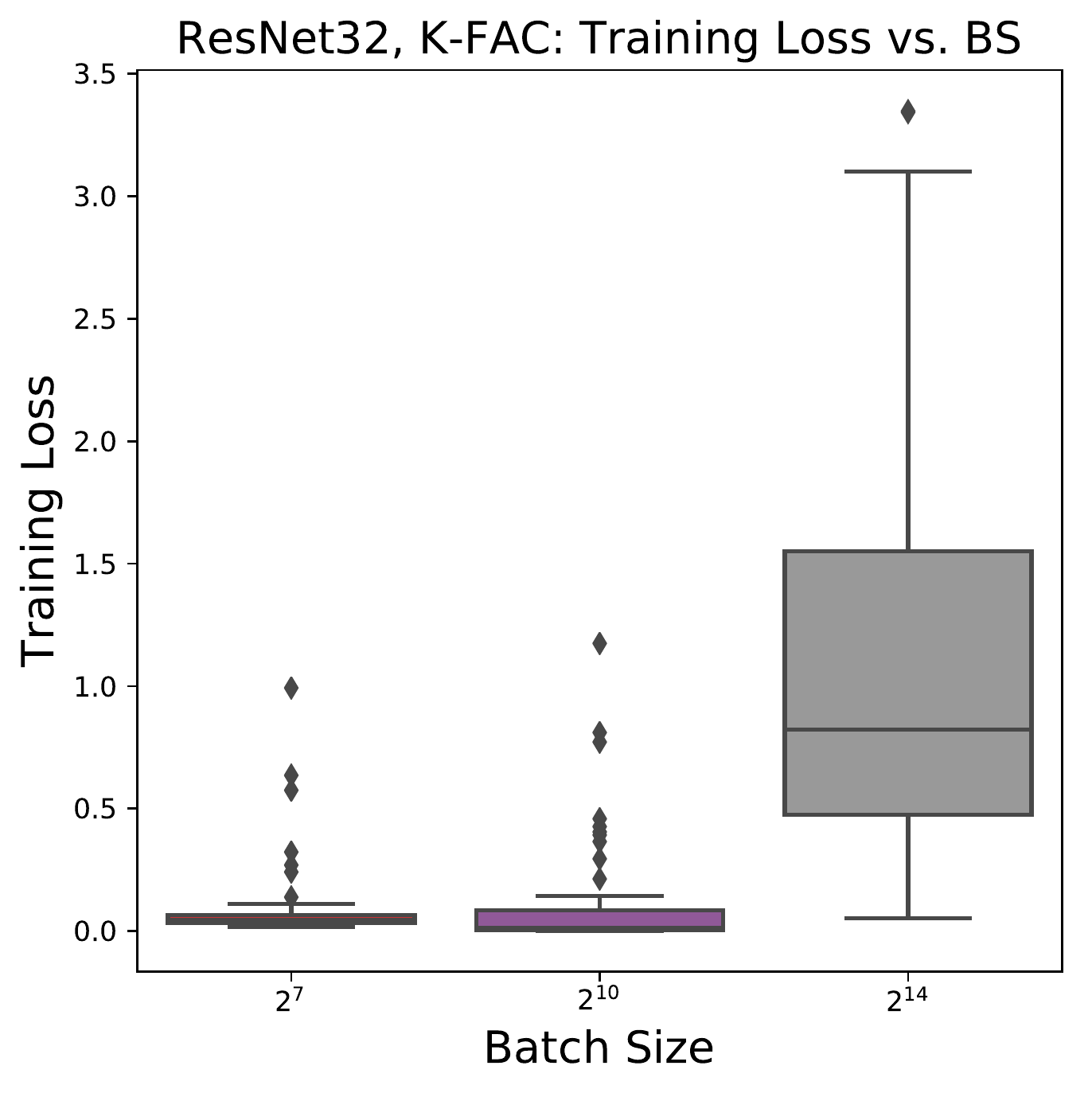}
  }
  
  \subfloat[]{\label{subplot-qalexnet-svhn}
  \includegraphics[width=.35\textwidth]{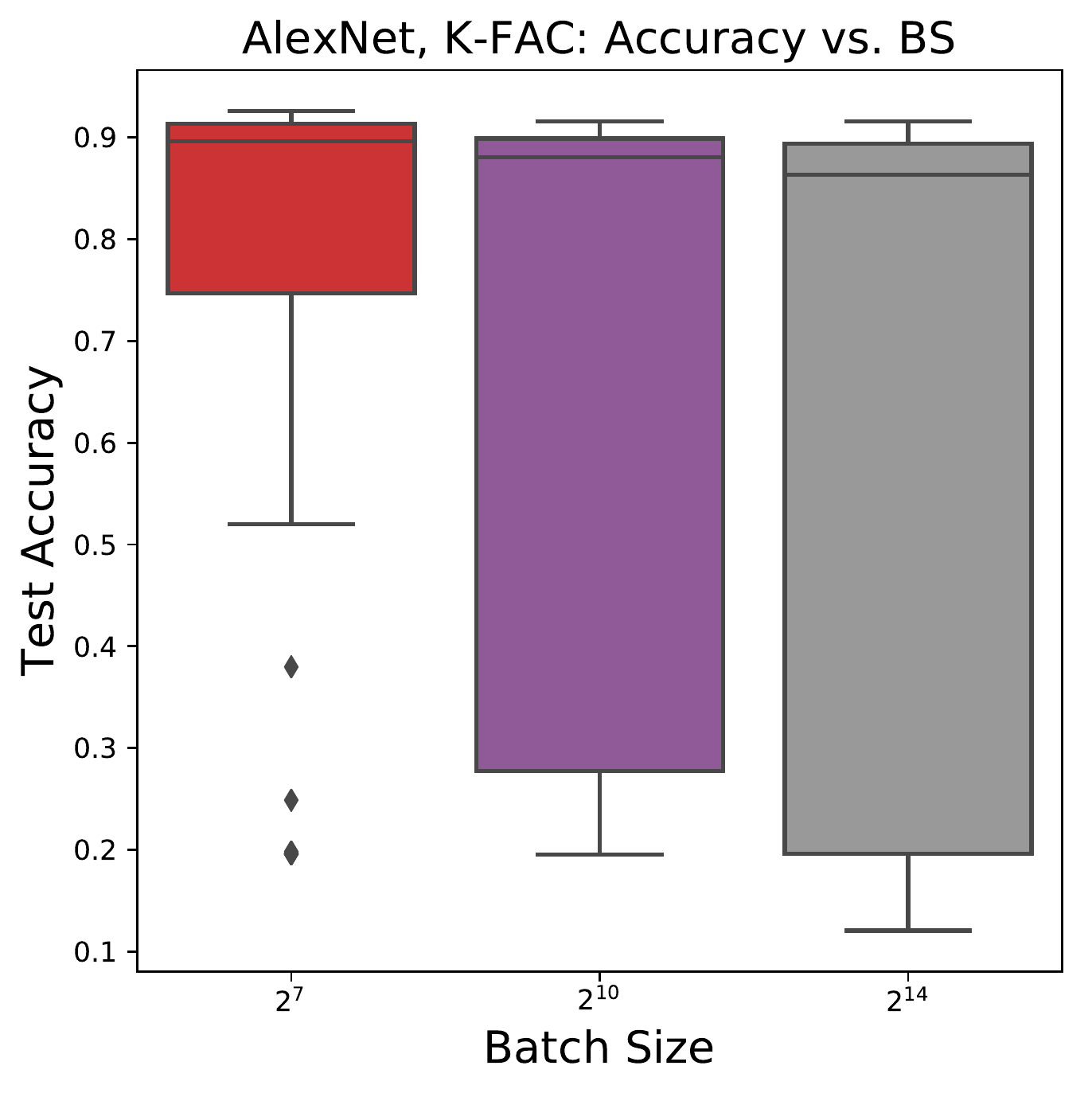}
  \includegraphics[width=.35\textwidth]{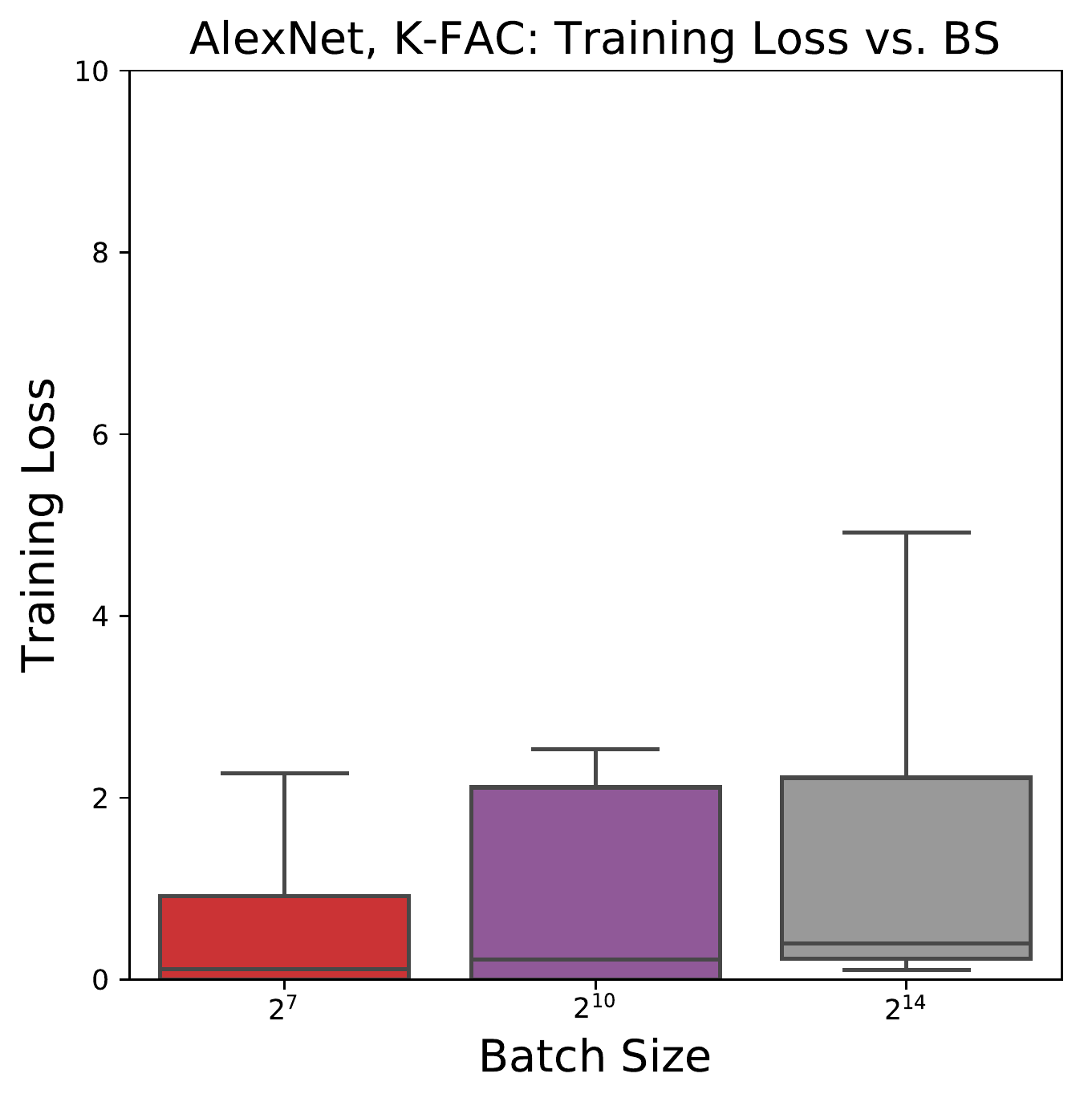}
  }
\end{center}
\caption{
(a): Test accuracy/training loss distribution versus batch size for K-FAC on CIFAR-10 with ResNet32 at the end of training under an adjusted epoch budget. (b): Test accuracy/training loss distribution versus batch size for K-FAC on SVHN with AlexNet at the end of training under an adjusted epoch budget. 
Larger batch sizes result in lower accuracy and higher training losses that are more sensitive to hyperparameter~choice. 
}
\label{fig:batch_v_acc_bp_add}
\end{figure}

Here, we show the hyperparameter sensitivity behavior of K-FAC on CIFAR-10 with ResNet32 and on SVHN with AlexNet to strengthen the argument that K-FAC is more sensitive to hyperparameter tuning under large batch sizes.
\fref{fig:batch_v_acc_bp_add} exhibits training loss and test accuracy distributions at the end of training by box plots, where each box contains 64 hyperparameter configurations. \fref{subplot-resnet32-cifar10} shows the results over 3 batch sizes ($2^7, 2^{10}, 2^{14}$) on CIFAR-10 with ResNet 32. \fref{subplot-qalexnet-svhn} shows the same range of batch sizes on SVHN with AlexNet. We can see an increase of hyperparameter sensitivity as the batch size increases. 

